\documentclass{article}

    \PassOptionsToPackage{numbers, compress}{natbib}

\usepackage[final]{neurips_2024}

\usepackage[utf8]{inputenc} 
\usepackage[T1]{fontenc}    
\usepackage{hyperref}       
\usepackage{url}            
\usepackage{booktabs}       
\usepackage{amsfonts}       
\usepackage{nicefrac}       
\usepackage{microtype}      
\usepackage{xcolor}         

\usepackage{amsmath}
\usepackage{amssymb}
\usepackage{mathtools}
\usepackage{amsthm}
\usepackage{enumitem}
\DeclareMathOperator{\diag}{diag}
\DeclareMathOperator{\E}{\mathbb{E}}

\title{Multivariate Probabilistic Time Series Forecasting with Correlated Errors}

\author{%
  Vincent Zhihao Zheng \\
  McGill University\\
  Montréal, QC, Canada \\
  \texttt{zhihao.zheng@mail.mcgill.ca} \\
  \And
  Lijun Sun\thanks{Corresponding author.} \\
  McGill University \\
  Montréal, QC, Canada \\
  \texttt{lijun.sun@mcgill.ca} \\
}

\begin{document}

\maketitle

\begin{abstract}

Accurately modeling the correlation structure of errors is critical for reliable uncertainty quantification in probabilistic time series forecasting. While recent deep learning models for multivariate time series have developed efficient parameterizations for time-varying contemporaneous covariance, but they often assume temporal independence of errors for simplicity. However, real-world data often exhibit significant error autocorrelation and cross-lag correlation due to factors such as missing covariates. In this paper, we introduce a plug-and-play method that learns the covariance structure of errors over multiple steps for autoregressive models with Gaussian-distributed errors. To ensure scalable inference and computational efficiency, we model the contemporaneous covariance using a low-rank-plus-diagonal parameterization and capture cross-covariance through a group of independent latent temporal processes. The learned covariance matrix is then used to calibrate predictions based on observed residuals. We evaluate our method on probabilistic models built on RNNs and Transformer architectures, and the results confirm the effectiveness of our approach in improving predictive accuracy and uncertainty quantification without significantly increasing the parameter size.

\end{abstract}

\section{Introduction}
Uncertainty quantification is crucial in time series forecasting, especially for applications that need more detailed insights than point forecasts. Probabilistic time series forecasting with deep learning (DL) has attracted attention for its ability to capture complex, nonlinear dependencies and provide the probability distribution of target variables \citep{gneiting2014probabilistic,benidis2022deep}. In multivariate time series, autoregressive models are widely used for probabilistic forecasting \citep{salinas2019high,rasul2020multivariate,rasul2021autoregressive}, modeling the joint one-step-ahead predictive distribution and generating multistep-ahead predictions in a rolling manner. To enable scalable learning, these models often assume that errors are independent over time. Typically, time series variables follow a Gaussian distribution \(\mathbf{z}_t = m(\mathbf{h}_t) + \boldsymbol{\eta}_t\), where \(m(\cdot)\) is the mean function and \(\boldsymbol{\eta}_t \sim \mathcal{N}(\mathbf{0}, \boldsymbol{\Sigma}_t)\) is a stochastic error process with contemporaneous covariance matrix \(\boldsymbol{\Sigma}_t\). The assumption of time-independence implies \(\operatorname{Cov}(\boldsymbol{\eta}_s, \boldsymbol{\eta}_t) = \mathbf{0}\), \(\forall s \ne t\). This holds when the model can account for all correlations between successive time steps through hidden states determined by previous values. However, real-world data often violate this assumption, as residuals exhibit substantial cross-correlation due to omission of important covariates and model misspecification.

Modeling error autocorrelation (or cross-correlation) is a key area of research in statistical time series models. A common approach is to assume that the error series follows a dependent temporal process, such as an autoregressive integrated moving average (ARIMA) model \citep{hyndman2018forecasting}. Deep learning models face similar challenges. Previous studies have attempted to incorporate temporally-correlated errors into the training process by modifying the loss function \citep{sun2021adjusting, zheng2023enhancing}. However, these methods, based on deterministic output, are not easily applicable to probabilistic forecasting models, particularly in a multivariate setting. A notable innovation is the batch training method introduced by \citet{zheng2024better}, which trains a univariate probabilistic forecasting model using generalized least squares (GLS) loss over batched errors. This approach parameterizes a dynamic covariance matrix to capture error autocorrelation, which is then used to calibrate the predictive distribution of time series variables. While this method consistently improves probabilistic forecasting performance compared to naive training (i.e., without considering autocorrelated errors); however, it is only applicable to univariate models, such as DeepAR \citep{salinas2020deepar}.

In this paper, we introduce an efficient method for learning error cross-correlation in multivariate probabilistic forecasting models. Our focus is on deep learning models that are autoregressive with Gaussian-distributed errors. Modeling cross-correlation in multivariate models presents challenges due to increased dimensionality, as the covariance matrix scales with the number of time series \(N\). To address this computational challenge, we propose characterizing error cross-correlation through a set of independent latent temporal processes using a low-rank parameterization of the covariance matrix. This approach prevents the computational cost from growing with the number of time series. Our method offers a general-purpose approach to multivariate probabilistic forecasting models, offering significantly improved predictive accuracy.

\textbf{Contributions:}
\begin{enumerate}[nosep, noitemsep, leftmargin=*]
    \item We introduce a plug-and-play method for training autoregressive multivariate probabilistic forecasting models using a redesigned GLS loss. (\S \ref{sec:methods}) 
    \item We propose an efficient parameterization of the error covariance matrix across multiple steps, enabling efficient computation of its inverse and determinant through matrix inversion and determinant lemmas. (\S \ref{sec:training})
    \item The learned covariance matrix is used to fine-tune the predictive distribution based on observed residuals. (\S \ref{sec:prediction})
    \item We demonstrate that the proposed method effectively captures error cross-correlation and improves prediction quality. Notably, these improvements are achieved through a statistical formulation without significantly increasing the size of model parameters. (\S \ref{sec:results})
\end{enumerate}

\section{Probabilistic Time Series Forecasting}

Denote \(\mathbf{z}_{t}=\left[z_{1,t},\dots,z_{N,t}\right]^\top \in \mathbb{R}^{N}\) as the vector of time series variables at time step $t$, where $N$ is the number of time series. Probabilistic time series forecasting can be formulated as estimating the joint conditional distribution \(p\left(\mathbf{z}_{{T+1}:{T+Q}} \mid \mathbf{z}_{{T-P+1}:{T}}; \mathbf{x}_{{T-P+1}:{T+Q}}\right)\) given the observed history $\{\mathbf{z}_{t}\}_{t=1}^T$, where $\mathbf{z}_{{t_1}:{t_2}} =\left[\mathbf{z}_{t_1},\ldots,\mathbf{z}_{t_2}\right]$ and  \(\mathbf{x}_{t}\) are known time-dependent covariates (e.g., time of day, day of week) for all future time steps. In essence, the problem involves predicting the time series values for \(Q\) future time steps using all available covariates and \(P\) steps of historical time series data:
\begin{equation}\label{eqn:prob1}
    p\left(\mathbf{z}_{{T+1}:{T+Q}} \mid \mathbf{z}_{{T-P+1}:{T}}; \mathbf{x}_{{T-P+1}:{T+Q}}\right)  =\prod\nolimits_{t=T+1}^{T+Q} p\left(\mathbf{z}_{t} \mid \mathbf{z}_{{t-P}:{t-1}}; \mathbf{x}_{{t-P}:{t}}\right),
\end{equation}
which becomes an autoregressive model that can be used for either one-step-ahead (\(Q=1\)) or multistep-ahead forecasting in a rolling manner. When performing multistep-ahead forecasting, samples are drawn in the prediction range (\(t\geq T+1\)) and fed back for the next time step until the end of the desired prediction range. In neural networks, the conditioning information is commonly encoded into a state vector \(\mathbf{h}_t\). Hence, Eq.~\eqref{eqn:prob1} can be expressed more concisely: 
\begin{equation}\label{eqn:prob2}
    p\left(\mathbf{z}_{{T+1}:{T+Q}} \mid \mathbf{z}_{{T-P+1}:{T}}; \mathbf{x}_{{T-P+1}:{T+Q}}\right)  =\prod\nolimits_{t=T+1}^{T+Q} p\left(\mathbf{z}_{t} \mid \mathbf{h}_{t}\right),
\end{equation}
where \(\mathbf{h}_{t}\) is mapped to the parameters of a parametric distribution (e.g., multivariate Gaussian). 

\begin{figure}[!t]
  \centering
  \includegraphics[width=0.9\textwidth, interpolate=false]{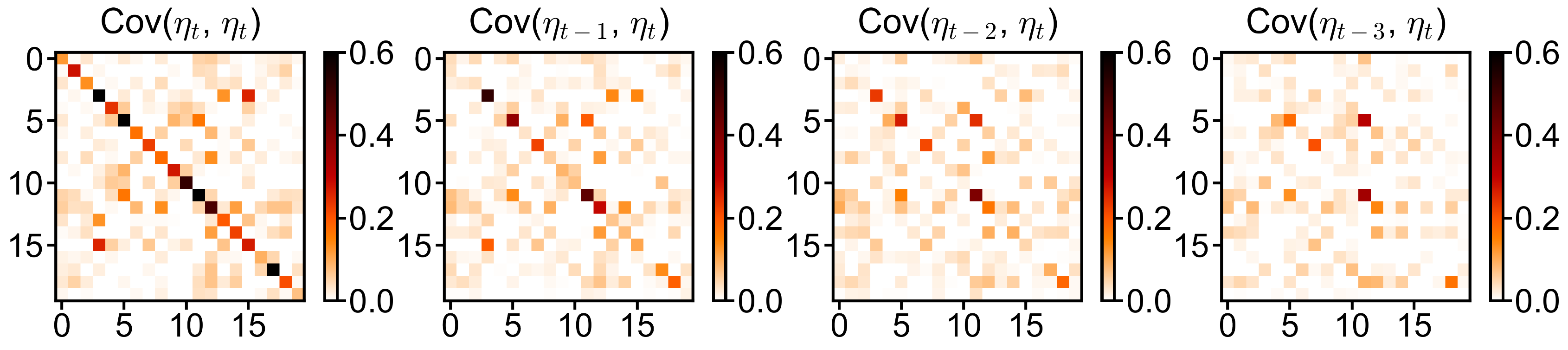}
  \caption{Contemporaneous covariance matrix $\operatorname{Cov}(\boldsymbol{\eta}_{t}, \boldsymbol{\eta}_{t})$ and cross-covariance matrix $\operatorname{Cov}(\boldsymbol{\eta}_{t-\Delta}, \boldsymbol{\eta}_{t}), \Delta=1,2,3$, calculated based on the one-step-ahead prediction residuals of GPVar on a batch of time series from the $\mathtt{m4\_hourly}$ dataset. For visualization clarity, covariance are clipped to the range $[0,0.6]$.}
\label{fig:corr_struc}
\end{figure}


Existing autoregressive models typically assume that the error at each time step is independent, meaning that $\mathbf{z}_t$ follows a multivariate Gaussian distribution:
\begin{equation}
    \left.\mathbf{z}_{t} \mid \mathbf{h}_{t}\right.\sim\mathcal{N}\left( \boldsymbol{\mu}(\mathbf{h}_{t}), \boldsymbol{\Sigma}(\mathbf{h}_{t})\right),
\end{equation}
where $\boldsymbol{\mu}(\cdot)$ and $\boldsymbol{\Sigma}(\cdot)$ map $\mathbf{h}_{t}$ to the mean and covariance parameters of a multivariate Gaussian distribution. This formulation can be decomposed as $\mathbf{z}_{t}=\boldsymbol{\mu}_{t}+\boldsymbol{\eta}_{t}$ with $\boldsymbol{\eta}_{t} \sim \mathcal{N}(\mathbf{0},\boldsymbol{\Sigma}_{t})$. The temporally independent error assumption corresponds to $\operatorname{Cov}(\boldsymbol{\eta}_{s}, \boldsymbol{\eta}_{t})=\mathbf{0}$ for any time points $s$ and $t$ where  $s\neq t$. Fig.~\ref{fig:corr_struc} provides an empirical example of the contemporaneous covariance matrix $\operatorname{Cov}(\boldsymbol{\eta}_{t}, \boldsymbol{\eta}_{t})$ and cross-covariance matrix $\operatorname{Cov}(\boldsymbol{\eta}_{t-\Delta}, \boldsymbol{\eta}_{t}), \Delta=1,2,3$. The results are calculated based on the prediction residuals of GPVar \citep{salinas2019high} on the $\mathtt{m4\_hourly}$ dataset. While multivariate models primarily focus on contemporaneous covariance, the residuals clearly exhibit temporal dependence, as $\operatorname{Cov}(\boldsymbol{\eta}_{t-\Delta}, \boldsymbol{\eta}_{t})\neq \mathbf{0}$. This non-zero cross-covariance suggests that residuals still contain valuable information, which can be leveraged to improve predictions.

\section{Related Work}

\subsection{Probabilistic Time Series Forecasting}
Probabilistic forecasting aims to model the probability distribution of target variables, unlike deterministic forecasting, which produces only point estimates. There are two main approaches: parametric probability density functions (PDFs) and quantile functions \citep{benidis2022deep}. For example, MQ-RNN \citep{wen2017multi} generates quantile forecasts using a sequence-to-sequence (Seq2Seq) RNN architecture. In contrast, PDF-based approaches assume a specific distribution (e.g., Gaussian, Poisson) and use neural networks to generate the distribution parameters. DeepAR \citep{salinas2020deepar}, for instance, uses an RNN to model hidden state transitions, while its multivariate version, GPVar \citep{salinas2019high}, employs a Gaussian copula to transform observations into Gaussian variables, assuming a joint multivariate Gaussian distribution.

Neural networks can also generate probabilistic model parameters. The deep state space model (SSM) \citep{rangapuram2018deep} uses an RNN to generate SSM parameters. The normalizing Kalman filter (NKF) \citep{de2020normalizing} combines normalizing flows (NFs) with the linear Gaussian state space model (LGM) to model nonlinear dynamics and evaluate the PDF of observations. NKF uses RNNs to produce LGM parameters at each time step, then transforms the LGM output into observations using NFs. \citet{wang2019deep} proposed the deep factor model, which includes a deterministic global component parameterized by an RNN and a random component from any classical probabilistic model (e.g., Gaussian white noise) to represent random effects. Some methods improve expressive conditioning for probabilistic forecasting by using Transformer instead of RNNs to model latent state dynamics, thus breaking the Markovian assumption in RNNs \citep{tang2021probabilistic}. Other approaches adopt more flexible distribution forms, including normalizing flows \citep{rasul2020multivariate}, diffusion models \citep{rasul2021autoregressive}, and copulas \citep{drouin2022tactis,ashok2023tactis}. For a recent and comprehensive review, we refer readers to \citet{benidis2022deep}.

\subsection{Modeling Correlated Errors}
Error correlation in time series has been extensively studied in econometrics and statistics \citep{prado2021time, hyndman2018forecasting, hamilton2020time}. In multivariate time series, correlation structure is characterized by contemporaneous covariance $\operatorname{Var}(\boldsymbol{\eta}_{t})=\operatorname{Cov}(\boldsymbol{\eta}_{t},\boldsymbol{\eta}_{t})$ and cross-covariance $\operatorname{Cov}(\boldsymbol{\eta}_{t-\Delta},\boldsymbol{\eta}_{t})$. Cross-covariance includes both the autocovariance of errors $\operatorname{Cov}(\eta_{i,t-\Delta},\eta_{i,t})$ and the cross-lag covariance $\operatorname{Cov}(\eta_{i,t-\Delta},\eta_{j,t})$ between pairs of components in the multivariate series. Contemporaneous covariance captures the correlation among individual time series at a specific point in time. In the univariate setting, DeepAR \citep{salinas2020deepar} achieves probabilistic forecasting by modeling the contemporaneous covariance, assuming that errors are independent over time. To address autocorrelation, \citet{sun2021adjusting} re-parameterized the input and output of neural networks to model first-order error autocorrelation, effectively capturing serially correlated errors using an AR$(1)$ process. This method improves the performance of one-step-ahead neural forecasting models, allowing joint optimization of base and error regressors, but is limited to deterministic models. In spatial modeling, \citet{saha2023random} introduced the RF-GLS model, which uses random forests to estimate nonlinear covariate effects and Gaussian processes (GP) to model spatial random effects. The RF-GLS model assumes that the error process follows an AR$(p)$ process to accommodate autocorrelated errors. \citet{zheng2024better} proposed training a probabilistic forecasting model with a GLS loss that explicitly models the time-varying autocorrelation of batched error terms, extending DeepAR to incorporate autocorrelated errors.

In the multivariate setting, most existing work focuses on modeling contemporaneous covariance, assuming that $\boldsymbol{\eta}_{t}$ is independently distributed, which implies $\operatorname{Cov}(\boldsymbol{\eta}_{t-\Delta},\boldsymbol{\eta}_{t})=\boldsymbol{0}$. For example, GPVar \citep{salinas2019high} generalizes DeepAR \citep{salinas2020deepar} to account for correlations between time series by viewing the distribution of time series variables as a Gaussian process. In Seq2Seq models, correlations can span across series and forecasting steps, as predictions for future time steps are generated simultaneously. Since predictions for future time steps are generated simultaneously, we refer to these correlations as contemporaneous correlations within the scope of this study. \citet{choi2022scalable} introduced a dynamic mixture of matrix Gaussian distributions to capture contemporaneous covariance of errors in Seq2Seq models. One exception that explicitly models error cross-correlation is \citep{zheng2023enhancing}, where the authors assume that the matrix-variate error term of a multivariate Seq2Seq model follows a matrix autoregressive (AR) process with seasonal lags. However, applying this technique to probabilistic forecasting models is not straightforward.

To the best of our knowledge, our work is the first to model cross-covariance in multivariate probabilistic time series forecasting. The closest related studies are by \citet{zheng2024better} and \citet{zheng2023enhancing}. \citet{zheng2024better} applies GLS loss in the temporal domain to model autocorrelated errors, but their approach is tailored for univariate time series. \citet{zheng2023enhancing} models cross-covariance in multivariate forecasting models, but their method is limited to deterministic models and requires predefined seasonal lags in the error autoregressive process. Our work extends \cite{zheng2024better} to the multivariate setting, enabling the modeling of the correlation structure of multivariate errors across multiple steps. In addition, we distinguish our approach from methods that directly model the distribution of time series variables, such as Copulas \citep{drouin2022tactis,ashok2023tactis}, where no decomposition of the error term is provided.

\section{Our Method}\label{sec:methods}
Our methodology builds upon the formulation outlined in Eq.~\eqref{eqn:prob2}, employing an autoregressive model as its foundational framework. Using an RNN as an example, a probabilistic forecasting model consists of two components. Firstly, it incorporates a transition model \(f_{\Theta}\) to capture the dynamics of state transitions $\mathbf{h}_t = f_{\Theta}\left(\mathbf{h}_{t-1}, \mathbf{z}_{t-1}, \mathbf{x}_{t}\right)$, thus inherently having autoregressive properties. Second, it integrates a distribution head, represented by $\theta$, which maps $\mathbf{h}_t$ to the parameters of the desired probability distribution. Following  GPVar \citep{salinas2019high}, our approach employs the multivariate Gaussian distribution as the distribution head. The time series variable can be decomposed into a deterministic mean component and a random error component $\mathbf{z}_{t} = \boldsymbol{\mu}_{t} + \boldsymbol{\eta}_{t}$, where $\boldsymbol{\eta}_t \sim \mathcal{N}(\mathbf{0},\boldsymbol{\Sigma}_t)$. To efficiently model the covariance $\boldsymbol{\Sigma}_t$ for large $N$, GPVar adopts a low-rank-plus-diagonal parameterization $\boldsymbol{\Sigma}_t=\boldsymbol{L}_{t}\boldsymbol{L}_{t}^\top+\diag{(\mathbf{d}_{t})}$, where $\boldsymbol{L}_{t}\in\mathbb{R}^{N\times R}$ ($R\ll N)$ and $\boldsymbol{d}_t\in\mathbb{R}_{+}^N$. Autoregressive models based on Gaussian likelihood typically assume that $\boldsymbol{\eta}_{t}$ are independently distributed following a multivariate Gaussian distribution. The log-likelihood of the distribution serves as the loss function for optimizing the model:
\begin{equation}\label{eqn:gls_ll}
    \mathcal{L}=\sum\nolimits_{t=1}^{T} \log p\left(\mathbf{z}_{t} \mid \theta\left(\mathbf{h}_{t}\right)\right) 
    \propto \sum\nolimits_{t=1}^{T} -\frac{1}{2}[\ln \lvert \boldsymbol{\Sigma}_{t} \rvert + \boldsymbol{\eta}_{t}^\top\boldsymbol{\Sigma}_{t}^{-1}\boldsymbol{\eta}_{t}].
\end{equation}

The parameters $\theta(\mathbf{h}_{t})$ are parameterized as $(\boldsymbol{\mu}_{t}, \boldsymbol{L}_{t}, \mathbf{d}_{t})$, where $\boldsymbol{\mu}_{t} \in \mathbb{R}^{N}$ represents the mean vector of the distribution. $\boldsymbol{L}_{t}$ and $\mathbf{d}_{t}$ correspond to the covariance factor and diagonal elements in the low-rank parameterization of the multivariate Gaussian distribution. We use shared mapping functions for all time series:
\begin{equation}
\begin{aligned}
    \mu_{i}(\mathbf{h}_{i,t}) & = \Tilde{\mu}(\mathbf{h}_{i,t}) = \mathbf{w}_{\mu}^\top\mathbf{h}_{i,t}, \\
    d_{i}(\mathbf{h}_{i,t}) & = \Tilde{d}(\mathbf{h}_{i,t}) = \log(1+\exp(\mathbf{w}_{d}^\top\mathbf{h}_{i,t})),\\\
    l_{i}(\mathbf{h}_{i,t}) & = \Tilde{l}(\mathbf{h}_{i,t}) = W_{l}\mathbf{h}_{i,t}, \\ 
\end{aligned}
\end{equation}
where $\mathbf{h}_{i,t} \in \mathbb{R}^{H}$, $\mathbf{w}_{\mu} \in \mathbb{R}^{H}$, $\mathbf{w}_{d} \in \mathbb{R}^{H}$, and $W_{l} \in \mathbb{R}^{R \times H}$ are parameters. Since the parameters of the mapping functions are shared across all time series, we can use a random subset of time series to compute the Gaussian likelihood-based loss in each optimization step, as any subset of \(\mathbf{z}_{t}\) will still follow a multivariate Gaussian distribution. In other words, we can train the model with a substantially reduced batch size $B<N$. 

\begin{figure}[!t]
  \centering\includegraphics[width=0.7\textwidth]{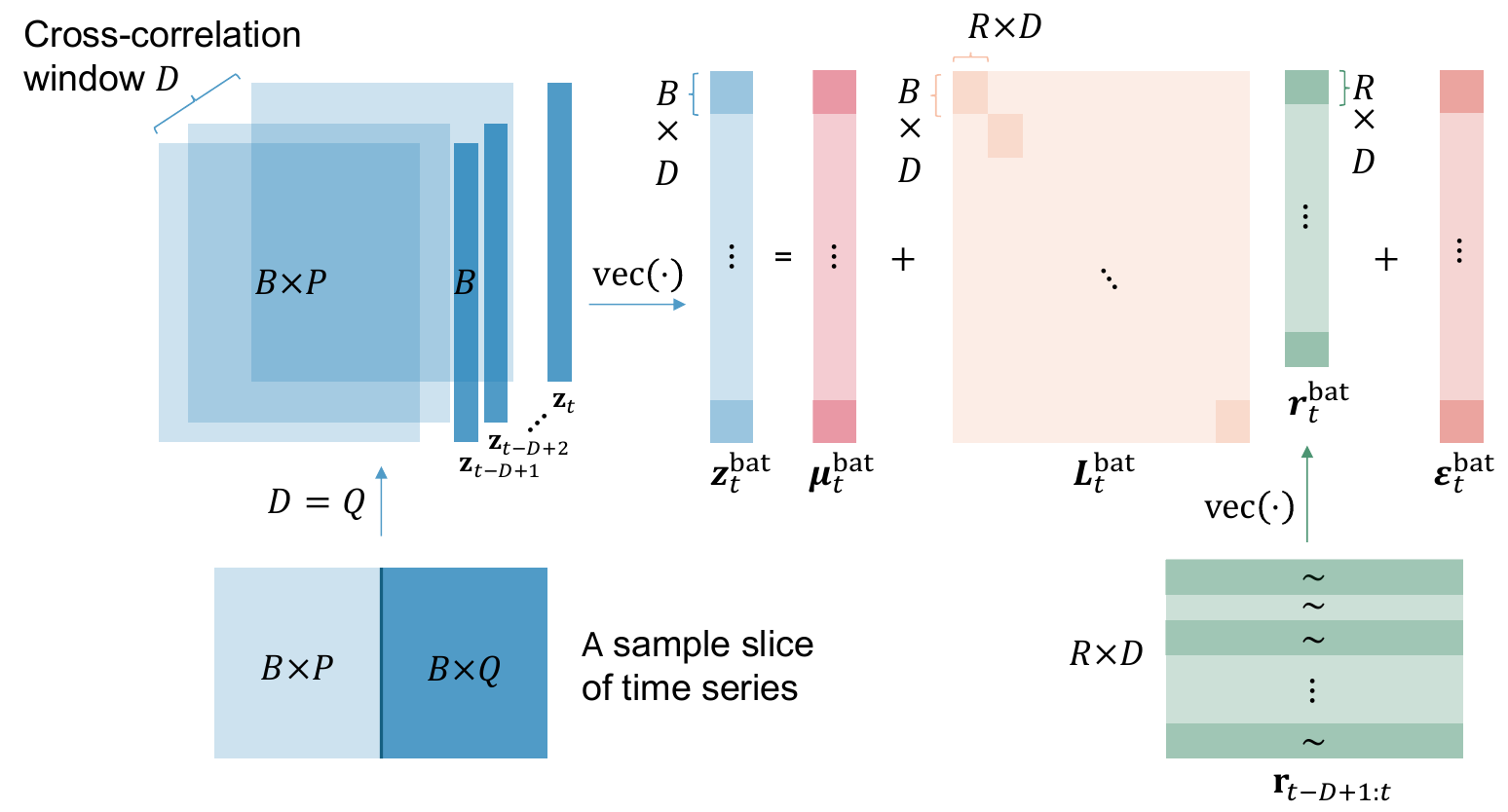}
  \caption{Graphic illustration of Eq.~\eqref{eqn:eq_bat}, where $B$ is the number of time series in a batch, $R$ is the rank of the covariance factor, $D$ is the time window we consider cross-correlation, $P$ and $Q$ are the conditioning range and prediction range. Cross-correlation is modeled by introducing correlation in each row of matrix $\mathbf{r}_{t-D+1:t}$.}
\label{fig:eq_bat}
\end{figure}

\subsection{Training with Correlated Errors}\label{sec:training}
We build upon the approach introduced in \citep{zheng2024better} to address cross-correlated errors in a multivariate context by introducing time-dependent error terms \(\boldsymbol{\eta}_{t}\) into the GLS loss. In many existing deep probabilistic forecasting models, such as GPVar \citep{salinas2019high}, a training batch typically consists of a sample slice of \(B\) time series spanning a temporal length of \(P+Q\), where \(P\) is the conditioning range and \(Q\) is the prediction range. The Gaussian likelihood is evaluated independently at each time step within the prediction range through one-step-ahead predictions. However, this approach overlooks the serial correlation of errors across consecutive time steps. To address this limitation, we propose modifying the likelihood function by introducing a dynamic covariance that accommodates the temporal dependence of the error term, as illustrated in Fig.~\ref{fig:eq_bat}. To achieve this, we organize \(D\) smaller slices of time series with a temporal length of \(P+1\) (i.e., \(Q=1\)), sorted by the prediction start time in sequential order, where \(D\) represents the time horizon over which we consider cross-correlation. The new batch structure effectively reconstructs the conventional training batch, covering the same time horizon when \(D=Q\). An example of the collection of target time series variables in a batch covering cross-correlation horizon $D$ is given by
\begin{equation}
\begin{aligned}
    \mathbf{z}_{t-D+1} & = \boldsymbol{\mu}_{t-D+1} + \boldsymbol{\eta}_{t-D+1}, \\
    \mathbf{z}_{t-D+2} & = \boldsymbol{\mu}_{t-D+2}+ \boldsymbol{\eta}_{t-D+2},\\\
    & \ldots\\
    \mathbf{z}_{t}& = \boldsymbol{\mu}_{t}+ \boldsymbol{\eta}_{t}, \\ 
\end{aligned}
\end{equation}
where for time point $t'$, $\boldsymbol{\mu}_{t'}$, $\boldsymbol{L}_{t'}$ and $\mathbf{d}_{t'}$ are the outputs of the model. The covariance parameterization in GPVar corresponds to
\begin{equation}    \boldsymbol{\eta}_{t'}=\boldsymbol{L}_{t'}\mathbf{r}_{t'}+\boldsymbol{\varepsilon}_{t'},
\end{equation}
where $\mathbf{r}_{t'}\sim \mathcal{N}(\mathbf{0},\mathbf{I}_{R})$ is a low-dimensional latent variable, and $\boldsymbol{\varepsilon}_{t'} \sim \mathcal{N}(\mathbf{0},\diag{(\mathbf{d}_{t'})})$ is an additional error independent of $\mathbf{r}_{t'}$. We denote $\boldsymbol{z}_t^{\text{bat}}= \operatorname{vec}\left(\mathbf{z}_{t-D+1:t}\right) \in \mathbb{R}^{DB}$ as the collection of target time series variables in a batch, where $\operatorname{vec}(\cdot)$ is an operator that stacks all the columns of a matrix into a vector. Similarly, we define  $\boldsymbol{\mu}_t^{\text{bat}} \in \mathbb{R}^{DB}$, $\boldsymbol{r}_t^{\text{bat}} \in \mathbb{R}^{DR}$, $\boldsymbol{\varepsilon}_t^{\text{bat}} \in \mathbb{R}^{DB}$, $\boldsymbol{d}_t^{\text{bat}} \in \mathbb{R}_{+}^{DB}$, and $\boldsymbol{L}_t^{\text{bat}} = \operatorname{blkdiag}(\{\boldsymbol{L}_{t'}\}_{t'=t-D+1}^{t}) \in \mathbb{R}^{DB \times DR}$, where \(\boldsymbol{L}_t^{\text{bat}}\) has a block diagonal structure (see Fig.~\ref{fig:eq_bat}). The batch-wise decomposition is then expressed as
\begin{equation}\label{eqn:eq_bat}
\boldsymbol{z}_t^{\text{bat}}=\boldsymbol{\mu}_t^{\text{bat}}+\boldsymbol{L}_t^{\text{bat}}\boldsymbol{r}_t^{\text{bat}}+\boldsymbol{\varepsilon}_t^{\text{bat}}.
\end{equation}

The default GPVar model assumes the latent variable $\boldsymbol{r}_t$ is temporally independent, meaning $\operatorname{Cov}\left(\boldsymbol{r}_s,\boldsymbol{r}_t\right)=\boldsymbol{0}, \forall s\neq t$. However, this assumption cannot capture the potential cross-correlation in the errors. To address this, we introduce temporal dependencies in the latent variable within a batch by assuming $\boldsymbol{r}_t^{\text{bat}} \sim \mathcal{N} \left( \boldsymbol{0}, \boldsymbol{C}_t \otimes \mathbf{I}_{R} \right)$, where
$\boldsymbol{C}_t$ is a dynamic $D \times D$ correlation matrix. This approach assumes that the rows in the matrix $\mathbf{r}_{t-D+1:t}=\left[\mathbf{r}_{t-D+1},\ldots,\mathbf{r}_{t}\right]$ are independent and identically distributed, following $\mathcal{N}\left(\boldsymbol{0},\boldsymbol{C}_t\right)$. To efficiently capture dynamic patterns over time, we follow \citet{zheng2024better} and express $\boldsymbol{C}_t$ as a dynamic weighted sum base kernel matrices: $\boldsymbol{C}_t = \sum_{m=1}^M w_{m,t}\boldsymbol{K}_m$, where $w_{m,t}\ge 0$ (with $\sum_m w_{m,t}=1$) represents the weights for each component. For simplicity, we model each component $\boldsymbol{K}_m$ using a kernel matrix generated from a squared-exponential (SE) kernel function, where the $(i,j)$-th entry is $\boldsymbol{K}^{ij}_m = \exp(-\frac{(i-j)^2}{l_m^2})$, with different lengthscales $l_m$ (e.g., $l=1,2,3,\dots$). In addition, we incorporate an identity matrix into the additive structure to account for the independent noise process. This parameterization ensures that $\boldsymbol{C}_t$ is a positive definite symmetric matrix with unit diagonals, making it a valid correlation matrix. The weights for these components are derived from the hidden state $\mathbf{h}_t$ at each time step through a small neural network, with the number of nodes in the output layer set to $M$ (i.e., the number of components). A softmax layer is used to ensure that these weights are summed up to 1. Note that the parameters of this network will be learned simultaneously with those of the base model.

\begin{figure}[!t]
  \centering\includegraphics[width=0.81\textwidth]{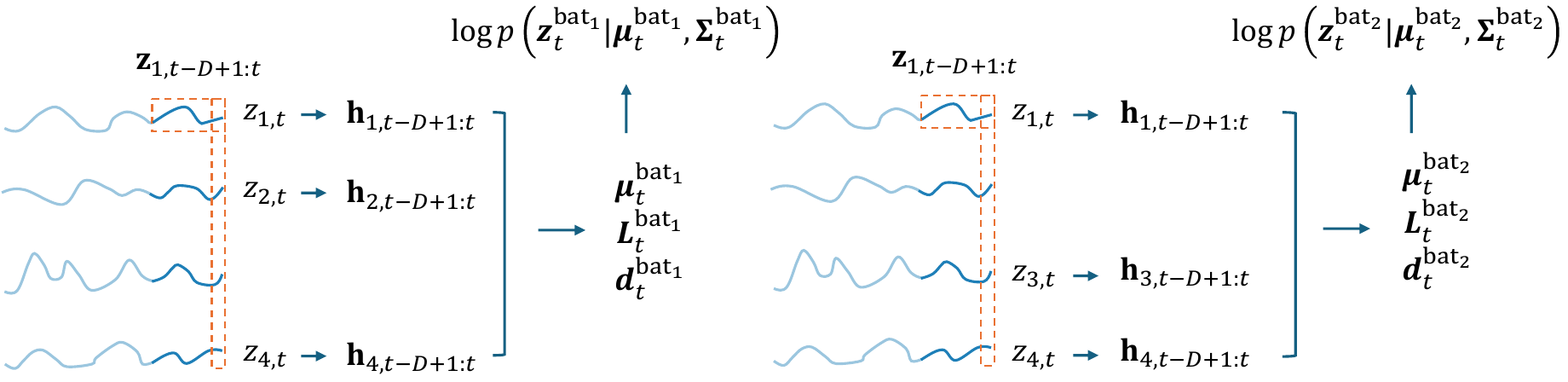}
  \caption{Illustration of the training process. Following \citep{salinas2019high}, time series dimensions are randomly sampled, and the base model (e.g., RNNs) is unrolled for each dimension individually (e.g., 1, 2, 4, followed by 1, 3, 4 as depicted). The model parameters are shared across all time series dimensions. A batch of time series variables \(\boldsymbol{z}_t^{\text{bat}}\) contains time series vectors \(\boldsymbol{z}_t\) covering time steps from \(t-D+1\) to \(t\). In contrast to \citep{salinas2019high}, our approach explicitly models dependencies over the extended temporal window from $t-D+1$ to $t$ during training.}
\label{fig:components}
\end{figure}

Marginalizing out $\boldsymbol{r}_t^{\text{bat}}$ in Eq.~\eqref{eqn:eq_bat}, we have $\boldsymbol{z}_t^{\text{bat}}\sim \mathcal{N}\left(\boldsymbol{\mu}_t^{\text{bat}}, \boldsymbol{\Sigma}_t^{\text{bat}}\right)$ with covariance 
\begin{equation}
\boldsymbol{\Sigma}_t^{\text{bat}}=(\boldsymbol{L}_t^{\text{bat}})(\boldsymbol{C}_t \otimes \mathbf{I}_{R})(\boldsymbol{L}_t^{\text{bat}})^\top+\diag(\boldsymbol{d}_t^{\text{bat}}).
\end{equation}
It is straightforward to derive that for any $i,j\in \{0,1,\ldots,D-1\}$ and $i\neq j$, the proposed model creates cross-covariance $\operatorname{Cov}\left(\boldsymbol{\eta}_{t-i},\boldsymbol{\eta}_{t-j}\right)=\boldsymbol{C}_t^{ij}\boldsymbol{L}_{t-i}\boldsymbol{L}_{t-j}^{\top}$ between times $t-i$ and $t-j$, which is no longer $\boldsymbol{0}$. While this parameterization results in a non-stationary multivariate process through varying coregionalization \cite{gelfand2004nonstationary, meng2021nonstationary}, a key difference is that both the coregionalization coefficient matrix $\boldsymbol{L}_t$ and 
the temporal correlation $\boldsymbol{C}_t$ are generated by a deep neural network. In this sense, our model can better characterize the empirical cross-covariance matrices of the residuals (see empirical examples in Fig.~\ref{fig:corr_struc}). As $\boldsymbol{\mu}_t^{\text{bat}}$, $\boldsymbol{L}_t^{\text{bat}}$, and $\boldsymbol{d}_t^{\text{bat}}$ are default outputs of the base probabilistic model, we can compute the overall likelihood (with overlapped data) as
\begin{equation}
\mathcal{L}=\sum\nolimits_{t=D}^{T} \log p\left(\boldsymbol{z}_{t}^{\text{bat}} \mid \boldsymbol{\mu}_t^{\text{bat}}, \boldsymbol{\Sigma}_t^{\text{bat}}\right).
\end{equation}
Here, computing the log-likelihood involves evaluating the inverse and the determinant of $\boldsymbol{\Sigma}_t^{\text{bat}}$ with size $DB\times DB$, for which a naive implementation has a prohibitive time complexity of $\mathcal{O}\left(D^3B^3\right)$. However, our parameterization of $\boldsymbol{\Sigma}_t^{\text{bat}}$ as $\boldsymbol{E}+\boldsymbol{A}\boldsymbol{C}\boldsymbol{A}^\top$, where $\boldsymbol{E}=\diag(\boldsymbol{d}_t^{\text{bat}})$,  $\boldsymbol{A}=\boldsymbol{L}_t^{\text{bat}}$, and $\boldsymbol{C}=\boldsymbol{C}_t \otimes \mathbf{I}_{R}$, allows us to leverage the Sherman–Morrison–Woodbury identity (matrix inversion lemma) and the companion matrix determinant lemma to simplify the computation:
\begin{equation}\label{eqn:sigma_inv}
\begin{aligned}
(\boldsymbol{E}+\boldsymbol{A}\boldsymbol{C}\boldsymbol{A}^\top)^{-1}
    &=\boldsymbol{E}^{-1}-\boldsymbol{E}^{-1}\boldsymbol{A}(\boldsymbol{C}^{-1}+\boldsymbol{A}^\top\boldsymbol{E}^{-1}\boldsymbol{A})^{-1}\boldsymbol{A}^\top\boldsymbol{E}^{-1},\\
    \det{(\boldsymbol{E}+\boldsymbol{A}\boldsymbol{C}\boldsymbol{A}^\top)}&=\det{(\boldsymbol{C}^{-1}+\boldsymbol{A}^\top\boldsymbol{E}^{-1}\boldsymbol{A})}\det{(\boldsymbol{C})}\det{(\boldsymbol{E})}.
\end{aligned}
\end{equation}
Then, the likelihood calculation only requires computing the inverse and determinant of a \(DR \times DR\) matrix, specifically \(\boldsymbol{C}^{-1}+\boldsymbol{A}^{\top}\boldsymbol{E}^{-1}\boldsymbol{A}\). These computations can be efficiently performed using Cholesky factorization. Detailed computations are provided in Appendix \S \ref{apx:ed_nll}.

Modeling the latent process \(\boldsymbol{r}_t\) offers several advantages. Firstly, because \(\boldsymbol{r}_t\) has a much lower dimension than \(\boldsymbol{\varepsilon}_t\), modeling the cross-correlation of \(\boldsymbol{r}_t\) results in a significantly smaller \(DR \times DR\) covariance matrix compared to the \(DB \times DB\) covariance matrix of \(\boldsymbol{\varepsilon}_t\). Secondly, since \(\boldsymbol{r}_t\) follows an isotropic Gaussian distribution, the covariance of \(\boldsymbol{r}_t^{\text{bat}}\) can be parameterized with a Kronecker structure \(\boldsymbol{C}_t \otimes \mathbf{I}_{R}\). This greatly simplifies the task into learning a \(D \times D\) correlation matrix shared by all time series in a batch. Lastly, similar to GPVar, we can still train the model in an end-to-end manner using a subset of time series in each iteration to ensure computational efficiency (Fig.~\ref{fig:components}).

\subsection{Multistep-ahead Rolling Prediction}\label{sec:prediction}

Autoregressive models perform multistep-ahead forecasting in an iterative manner, where the model generates a sample at each time step during prediction, using it as input for the subsequent step, and continuing this process until the desired prediction range is reached. Our approach enhances this process, similar to \citet{zheng2024better}, by offering additional calibration based on the learned correlation matrix $\boldsymbol{C}_t$. Assuming observations are available up to time step $t$, the conditional distribution of $\boldsymbol{\eta}_{t+1}$ given errors in the past $(D-1)$ steps, can be derived as
\begin{equation}\label{eqn:cond_dist}
\boldsymbol{\eta}_{t+1} \mid \boldsymbol{\eta}_{t}, \boldsymbol{\eta}_{t-1},\ldots,\boldsymbol{\eta}_{t-D+2}  \sim \mathcal{N} \left( \boldsymbol{\Sigma}_*\boldsymbol{\Sigma}_{\text{obs}}^{-1}\boldsymbol{\eta}_{\text{obs}}, \boldsymbol{\Sigma}_{t+1}- \boldsymbol{\Sigma}_*\boldsymbol{\Sigma}_{\text{obs}}^{-1} \boldsymbol{\Sigma}_*^{\top}\right), 
\end{equation}
where $\boldsymbol{\eta}_{\text{obs}}=\operatorname{vec}\left(\left[\boldsymbol{\eta}_{t-D+2},\ldots,\boldsymbol{\eta}_{t-1},\boldsymbol{\eta}_t\right]\right) \in \mathbb{R}^{(D-1)B}$ represents the set of residuals, accessible at forecasting step $t+1$. Here, $\boldsymbol{\Sigma}_{\text{obs}}$ is a $(D-1)B\times (D-1)B$ partition of $\boldsymbol{\Sigma}_{t+1}^{\text{bat}}$ that captures the covariance of $\boldsymbol{\eta}_{\text{obs}}$, and $\boldsymbol{\Sigma}_{*}$ is a $B \times (D-1)B$ partition of $\boldsymbol{\Sigma}_{t+1}^{\text{bat}}$ representing the covariance between $\boldsymbol{\eta}_{t+1}$ and $\boldsymbol{\eta}_{\text{obs}}$,  i.e., $\boldsymbol{\Sigma}_{t+1}^{\text{bat}}=\begin{bmatrix}
\boldsymbol{\Sigma}_{\text{obs}} & \boldsymbol{\Sigma}_*^\top \\
\boldsymbol{\Sigma}_* & \boldsymbol{\Sigma}_{t+1} 
\end{bmatrix}$. For conciseness, we omit the time index $t$ in $\boldsymbol{\Sigma}_{\text{obs}}$, $\boldsymbol{\Sigma}_{*}$ and $\boldsymbol{\eta}_{\text{obs}}$. Since $\boldsymbol{\mu}_{t+1}$ is a deterministic output from the base model, a sample of the target variables $\Tilde{\mathbf{z}}_{t+1}$ can be derived by first drawing a sample $\Tilde{\boldsymbol{\eta}}_{t+1}$ from Eq.~\eqref{eqn:cond_dist}, then combining it with the predicted mean vector $\boldsymbol{\mu}_{t+1}$ as $\Tilde{\mathbf{z}}_{t+1}=\boldsymbol{\mu}_{t+1}+\Tilde{\boldsymbol{\eta}}_{t+1}$. It should be noted that we can still leverage the Sherman-Morrison-Woodbury identity when computing the inverse $\boldsymbol{\Sigma}_{\text{obs}}^{-1}$.


By taking the sample $\Tilde{\boldsymbol{\eta}}_{t+1}$ as an observed residual, we can iteratively apply the process described in Eq.~\eqref{eqn:cond_dist} to derive a trajectory of $\{\Tilde{\mathbf{z}}_{t+q}\}_{q=1}^Q$. Repeating this procedure allows us to generate multiple samples, characterizing the predictive distribution at each time step.

\section{Experiments}\label{sec:results}
\subsection{Evaluation of Predictive Performance}

\textbf{Datasets}. We use widely recognized time series benchmarking datasets from GluonTS \citep{alexandrov2020gluonts}. The prediction range (\(Q\)) for each dataset follows the configurations provided by GluonTS. We applied a sequential split into training, validation, and testing sets for each dataset. Each dataset was standardized using the mean and standard deviation from the training set, and predictions were rescaled to their original values for evaluation. Further details on the datasets can be found in Appendix \S \ref{apx:ed_data}.

\textbf{Base probabilistic models}. We integrated the proposed method into two distinct autoregressive models: the RNN-based GPVar \citep{salinas2019high} and the decoder-only Transformer \citep{radford2018improving}. These models are trained to generate distribution parameters as described in \S \ref{sec:methods}. Our approach can be applied to other autoregressive multivariate models with minimal adjustments, provided the final prediction follows a multivariate Gaussian distribution. The implementation is based on using PyTorch Forecasting \citep{pytorchforecasting}. Both models use lagged time series values and additional features or covariates as inputs. Details on model training (\S \ref{apx:ed_train}), hyperparameter tuning (\S \ref{apx:ed_hyparam}), and the base model (\S \ref{apx:ed_model}) are provided in Appendix \S \ref{apx:ed}. The code is available at \href{https://github.com/rottenivy/mv_pts_correlatederr}{https://github.com/rottenivy/mv\_pts\_correlatederr}.

\textbf{Dynamic correlation matrix}. We introduce a limited number of additional parameters to project the state vector \(\mathbf{h}_{t}\) into component weights \(w_{m,t}\), which are used to generate the dynamic correlation matrix \(\boldsymbol{C}_t\). The number of base kernels (\(M\)) for generating \(\boldsymbol{C}_t\) and the associated lengthscale set \(\{l_m\}_{m=1}^{M-1}\) are treated as hyperparameters. We perform a grid search over \(M = 2,3,4\) and two sets of lengthscales—\(\{0.5, 1.5, \dots\}\) and \(\{1.0, 2.0, \dots\}\). Models with the best validation loss are selected. These different lengthscales capture varying correlation decay rates, enabling the model to account for different temporal patterns. The time-varying component weights enable dynamic adaptation to changing correlation structures over time. 

\textbf{Baselines}. We evaluate the proposed method by comparing it with a baseline model trained without accounting for error cross-correlation (Eq.~\eqref{eqn:gls_ll}). The baseline model represents a special case of our model with \(\boldsymbol{C}_t = \boldsymbol{I}_D\). To ensure a straightforward and fair comparison, we align the cross-correlation range (\(D\)) with the prediction range (\(Q\)), ensuring identical data sampling processes for both methods. Additionally, we set \(P = Q\) following the default configuration in GluonTS. We also include VAR and GARCH as naive baseline models (see Appendix \S \ref{apx:ed_naive}).









\textbf{Metrics}. We use the Continuous Ranked Probability Score (CRPS) \citep{gneiting2007strictly} as the main metric:
\begin{equation}
    \operatorname{CRPS}(F, z)=\mathbb{E}_F|Z-z|-\frac{1}{2} \mathbb{E}_F\left|Z-Z^{\prime}\right|,
\end{equation}
where $F$ is the cumulative distribution function (CDF) of the predicted variable, $z$ is the observation, $Z$ and $Z^{\prime}$ are independent copies of the prediction samples associated with the distribution $F$. To evaluate multivariate dependencies in the time series data, we compute $\operatorname{CRPS}_{\text{sum}}$ by first summing both the forecast and ground-truth values across all time series and then calculating the $\operatorname{CRPS}$ over the resulting sums \citep{salinas2019high,drouin2022tactis,ashok2023tactis}. As \(\operatorname{CRPS}_{\text{sum}}\) may overlook model performance on individual dimensions \citep{koochali2022random}, we also report additional metrics, e.g., the energy score \citep{gneiting2007strictly,marcotte2023regions}, in Appendix \S \ref{apx:mar_metric}.

\textbf{Training dynamics}. Our approach incurs additional training costs per optimization step due to the more complex likelihood function. As shown in Appendix \S \ref{apx:mar_training}, the training time per epoch for models using our method is generally longer than that of baseline methods. However, our parameterization allows for scalability to large time series datasets by using a small random subset of time series at each optimization step during training.

\begin{table*}[!t]
\scriptsize
\caption{$\operatorname{CRPS}_{\text{sum}}$ accuracy comparison. ``w/o'' denotes methods without time-dependent errors, while ``w/'' indicates our method. Bold values show models with time-dependent errors performing better. Mean and standard deviation are obtained from 10 runs of each model.}
\label{tab:crps_sum}
\begin{center}
\begin{tabular}{lcccccc}
\toprule
                          & VAR     & GARCH       & \multicolumn{2}{c}{GPVar}              & \multicolumn{2}{c}{Transformer} \\
\cmidrule(lr){2-7}
                          &       &      & w/o              & w/           & w/o              & w/   \\
\midrule
$\mathtt{exchange\_rate}$ & 0.0033±0.0000  & 0.0435±0.0001 & 0.0068±0.0004 & 0.0117±0.0004          & 0.0055±0.0002 & \textbf{0.0042±0.0002} \\
$\mathtt{solar}$          & 0.7663±0.0050  & 0.8752±0.0015 & 0.7103±0.0065 & \textbf{0.6929±0.0039} & 0.4960±0.0034 & \textbf{0.4132±0.0027} \\
$\mathtt{electricity}$    & 0.1264±0.0006  & 0.2847±0.0015 & 0.0430±0.0005 & \textbf{0.0403±0.0004} & 0.0494±0.0004 & 0.0638±0.0003          \\
$\mathtt{traffic}$        & 3.5241±0.0084  & 0.4459±0.0005 & 0.1095±0.0002 & \textbf{0.0649±0.0002} & 0.0717±0.0002 & 0.0981±0.0002          \\
$\mathtt{wikipedia}$      & 26.2025±0.0389 & 0.6699±0.0045 & 0.1745±0.0008 & \textbf{0.0743±0.0009} & 0.0841±0.0013 & \textbf{0.0500±0.0005} \\
$\mathtt{m4\_hourly}$     & 0.2352±0.0008  & 0.2758±0.0006 & 0.0613±0.0004 & \textbf{0.0358±0.0002} & 0.0651±0.0004 & \textbf{0.0616±0.0003} \\
$\mathtt{m1\_quarterly}$  & N/A            & N/A           & 0.3942±0.0030 & \textbf{0.3538±0.0017} & 0.4448±0.0027 & \textbf{0.4367±0.0028} \\
$\mathtt{pems03}$         & 0.0598±0.0002  & 0.3202±0.0007 & 0.0503±0.0001 & \textbf{0.0491±0.0002} & 0.0490±0.0001 & \textbf{0.0386±0.0001} \\
$\mathtt{uber\_hourly}$   & N/A            & N/A           & 0.0342±0.0006 & \textbf{0.0222±0.0004} & 0.0632±0.0003 & \textbf{0.0513±0.0005}   \\
\cmidrule(lr){1-7}
&   &    &  \multicolumn{1}{r}{avg. rel. impr.} & \multicolumn{1}{l}{13.79\%}  & \multicolumn{1}{r}{avg. rel. impr.} & \multicolumn{1}{l}{6.91\%}  \\
\bottomrule
\end{tabular}
\end{center}
\end{table*}

\textbf{Benchmark results}. The \(\operatorname{CRPS}_{\text{sum}}\) results are presented in Table~\ref{tab:crps_sum}. Our method achieves an average improvement of 13.79\% for GPVar and 6.91\% for the Transformer model. It is important to note that the degree of performance enhancement varies across different base models and datasets, influenced by factors such as the inherent data characteristics and the performance of different model architectures. The alignment between the actual correlation structure and our kernel assumption also plays a crucial role in the effectiveness of our method. Additionally, our approach demonstrates consistent improvements across five different metrics, with significant gains in multivariate metrics such as the energy score (Appendix \S \ref{apx:mar_results}). 

To provide further insights, we compare the residual autocorrelation and cross-lag correlation with and without applying our method in Appendix \S \ref{apx:extra_exp_crc}, showing that our method effectively reduces cross-correlations in many scenarios. We use ACF plot comparisons to illustrate the reduction in autocorrelation and cross-correlation plot comparisons to demonstrate the decrease in cross-lag correlation. The residuals generated by the model with our method exhibit weaker cross-correlations, which is particularly enhanced by the calibration process during prediction (\S \ref{sec:prediction}). 

Furthermore, Appendix \S \ref{apx:extra_exp_pbefs} separates the accuracy improvement over forecast steps for each dataset. The performance improvement is shown to be related to both the absolute time across the temporal span of the dataset (especially for time series with strong periodic patterns) and the relative time over the prediction horizon.

\subsection{Model Interpretation}

\begin{figure}[!t]
  \centering
  \includegraphics[width=0.95\textwidth]{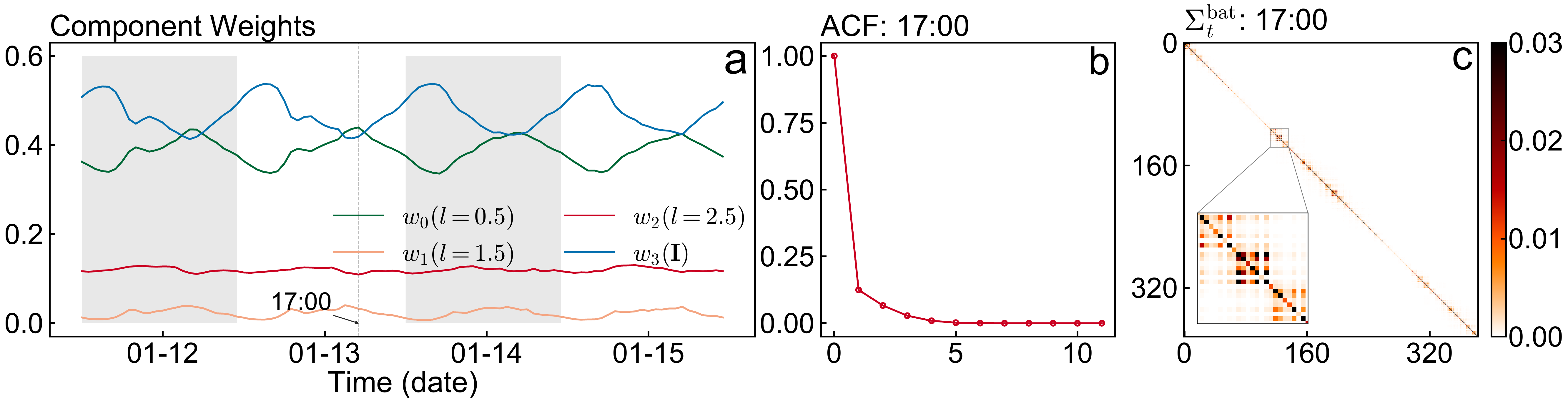}
  \caption{(a) Component weights for generating $\boldsymbol{C}_t$ for a batch of time series ($B=8$) from the $\mathtt{m4\_hourly}$ dataset obtained by the GPVar model. Parameters $w_0, w_1, w_2$ represent the component weights of the kernel matrices associated with lengthscales $l=0.5,1.5,2.5$, and $w_3$ is the component weight of the identity matrix. Shaded areas distinguish different days; (b) The autocorrelation function (ACF) indicated by the correlation matrix $\boldsymbol{C}_t$ at 17:00. Given the rapid decay of the ACF, we only plot 12 lags to enhance visualization; (c) The corresponding covariance matrix of the associated target variables $\boldsymbol{\Sigma}_t^{\text{bat}}$ at 17:00. A zoom-in view of a $3B \times 3B$ region is illustrated in the plot, where the diagonal blocks represent $B\times B$ covariance matrices $\boldsymbol{\Sigma}_{t^\prime}$ of $\mathbf{z}_{t^\prime}$ over three consecutive time steps. The off-diagonal blocks describe the cross-covariance $\operatorname{Cov}(\mathbf{z}_{t-\Delta}, \mathbf{z}_{t})$, $ \forall \Delta \neq 0$. For visualization clarity, covariance values are clipped to the range $[0,0.03]$.}
\label{fig:mix_weights}
\vspace{-0.3cm}
\end{figure}

Our method captures error cross-correlation through the dynamic construction of a covariance matrix, achieved by combining kernel matrices with varying lengthscales in a dynamically weighted sum. A small lengthscale corresponds to short-range positive correlations, while a large lengthscale captures positive correlations over longer lags.

In Fig.~\ref{fig:mix_weights}, we depict the dynamic component weights and the resulting autocorrelation function (the first row of the correlation matrix \(\boldsymbol{C}_t\)) for a batch of time series from the $\mathtt{m4\_hourly}$ dataset spanning a four-day window. We also provide the covariance matrix of \(\boldsymbol{z}_t^{\text{bat}}\) using the correlation matrix and model outputs at a specific time of day. The component weight \(w_3\), corresponding to the identity matrix, dominates throughout the observation period. This suggests that the error correlation is generally mild over time. This behavior is influenced by the Kronecker structure used to parameterize the covariance over the low-dimensional latent variables \(\boldsymbol{r}_{t}\), which assumes all latent processes share the same autocorrelation structure. Given the Kronecker structure, the model tends to learn the mildest temporal correlation among the time series in a batch.

Moreover, we observe that the dynamic component weights adjust the correlation strengths. Specifically, when the weight assigned to the identity matrix (\(w_3\)) increases, the error process tends to be more independent. In contrast, when the weights assigned to the other kernel matrices (\(w_0\), \(w_1\), and \(w_2\)) are larger, the error process becomes more correlated, as the kernel matrices with different lengthscales combine to formulate a specific correlation structure. Fig.~\ref{fig:mix_weights}(a) demonstrates pronounced daily patterns in temporal correlation, particularly when errors exhibit increased correlation around 17:00 each day. The corresponding autocorrelation function is shown in Fig.~\ref{fig:mix_weights}(b). Fig.~\ref{fig:mix_weights}(c) illustrates the corresponding covariance matrix of the associated target variables within the cross-correlation horizon. The diagonal blocks represent the contemporaneous covariance \(\boldsymbol{\Sigma}_{t}\) of \(\mathbf{z}_{t}\) at each time step, while the off-diagonal blocks capture the cross-covariance \(\operatorname{Cov}(\mathbf{z}_{t-\Delta}, \mathbf{z}_{t})\) for \(\forall \Delta \neq 0\), effectively modeled by our approach. The zoomed-in view provides a \(3B \times 3B\) region that illustrates the cross-covariance within two lags. We observe that the cross-covariance is most pronounced at lag 1, consistent with the observation in Fig.~\ref{fig:mix_weights}(a) that the component weight \(w_0\), assigned to the base kernel matrix with lengthscale \(l=0.5\), is more pronounced than \(w_1\) and \(w_2\).


\section{Discussion}

In this section, we discuss factors that influence the performance of our method. Specifically, we highlight the effectiveness of our model in long-term forecasting across various scenarios. We also discuss the effect of scaling up to larger batch sizes during prediction. Additionally, we examine the impact of non-Gaussian errors on model performance.

\textbf{Long-term forecasting}. The advantage of modeling error correlation can vary in long-term forecasting, especially in autoregressive predictions where errors accumulate and propagate over time. Using residuals from previous time steps to calibrate forecasts may be beneficial for non-stationary segments of the time series. However, for time series with strong periodic effects, the model may also rely on seasonal lags. As shown in Fig.~\ref{fig:deepar_crps_sum_step} and Fig.~\ref{fig:gpt_crps_sum_step} of the Appendix, the advantage of modeling error correlation can decrease in longer-term forecasts compared to shorter-term forecasts for some datasets with strong periodic effects (e.g., the $\mathtt{traffic}$ dataset in Fig.~\ref{fig:deepar_crps_sum_step}). It is not necessarily true that the advantage diminishes for long-horizon predictions, as the effectiveness of our method depends on the quality of predictions during inference. In cases where the model provides accurate long-term forecasts, the benefit of modeling correlated errors may be less pronounced.

\textbf{Scalability}. Increasing the number of time series \(B\) in a batch leads to higher training costs. Because the model requires numerous iterations over the dataset for optimization, using a large \(B\) during training is not feasible. However, during prediction, the batch size can be increased to leverage more information. This may enhance both prediction accuracy and error calibration, provided sufficient memory is available. We demonstrate the effect of increasing batch size during prediction in Appendix \S \ref{apx:num_b_pred} through additional experiments. Both models, with and without our method, show improvement from increased batch sizes during prediction, as reflected by a decrease in $\operatorname{CRPS}_{\text{sum}}$.

\textbf{Non-Gaussian errors}. For the baseline model, assuming Gaussian errors may lead to model misspecification, resulting in more correlated residuals. To address this issue, we also trained the baseline models using the likelihood of a multivariate \(t\)-distribution; the results are presented in Table~\ref{tab:crps_sum_tdist} of the Appendix. Although using an alternative distribution can lead to better performance on some datasets without our method, we observed that our method effectively closes the performance gap when the \(t\)-distribution outperforms the Gaussian assumption. We chose the Gaussian distribution for its beneficial properties, including its marginalization rule and well-defined conditional distributions, both essential for statistically consistent model training and reliable inference. Thus, a more effective approach could involve first transforming the original observations into Gaussian-distributed data using a Gaussian Copula \citep{salinas2019high}, followed by applying our method.

\section{Conclusion and Broader Impacts}
This paper presents a novel approach for addressing error cross-correlation in multivariate probabilistic time series forecasting, specifically for models with autoregressive properties and Gaussian distribution outputs. We construct a dynamic covariance matrix using a small set of independent and identically distributed latent temporal processes. These latent processes effectively model temporal correlation and integrate seamlessly into the base model, where the contemporaneous covariance is parameterized by a low-rank-plus-diagonal structure. This approach enables the modeling and prediction of a time-varying covariance matrix for the target time series variables. The experimental results demonstrate its effectiveness in enhancing uncertainty quantification. 

Our contributions are two-fold. First, our approach relaxes the time-independent error assumption during the training process for probabilistic forecasting models, addressing the reality that residuals are typically time-dependent. Second, the learned cross-correlation improves multistep-ahead predictions by refining the distribution output at each forecasting step. These enhancements to existing models have broader implications for fields such as finance, healthcare, and energy, where improved forecasts and uncertainty quantification can lead to more informed decisions.

There are several avenues for future research. First, the Kronecker structure \(\boldsymbol{C}_t \otimes \mathbf{I}_{R}\) for the covariance matrix of the latent variable \(\boldsymbol{r}_t^{\text{bat}}\) may be too restrictive for multivariate time series problems. Exploring more flexible covariance structures, such as employing different \(\boldsymbol{C}_{r,t}\) matrices for each latent temporal process as in the linear model of coregionalization (LMC, \cite{gelfand2004nonstationary,meng2021nonstationary}), could be a promising direction for further investigation. Second, the parameterization of \(\boldsymbol{C}_t\) could be expanded. Instead of using SE kernels, \(\boldsymbol{C}_t\) could be parameterized as fully learnable positive definite symmetric Toeplitz matrices. For example, an AR(\(p\)) process has a covariance structure in Toeplitz form, allowing for the modeling of negative correlations. This alternative approach could offer greater flexibility in capturing complex correlation patterns in multivariate time series data. 

\begin{ack}

We acknowledge the support from the Natural Sciences and Engineering Research Council (NSERC) of Canada (Discovery Grant). Vincent Zhihao Zheng also acknowledges the support received from the FRQNT B2X Doctoral Scholarship Program.

\end{ack}


%

{
\medskip
\small
\bibliographystyle{unsrtnat}
\bibliography{refs}

\begin{thebibliography}{46}
\providecommand{\natexlab}[1]{#1}
\providecommand{\url}[1]{\texttt{#1}}
\expandafter\ifx\csname urlstyle\endcsname\relax
  \providecommand{\doi}[1]{doi: #1}\else
  \providecommand{\doi}{doi: \begingroup \urlstyle{rm}\Url}\fi

\bibitem[Gneiting and Katzfuss(2014)]{gneiting2014probabilistic}
Tilmann Gneiting and Matthias Katzfuss.
\newblock Probabilistic forecasting.
\newblock \emph{Annual Review of Statistics and Its Application}, 1\penalty0 (1):\penalty0 125--151, 2014.

\bibitem[Benidis et~al.(2022)Benidis, Rangapuram, Flunkert, Wang, Maddix, Turkmen, Gasthaus, Bohlke-Schneider, Salinas, Stella, et~al.]{benidis2022deep}
Konstantinos Benidis, Syama~Sundar Rangapuram, Valentin Flunkert, Yuyang Wang, Danielle Maddix, Caner Turkmen, Jan Gasthaus, Michael Bohlke-Schneider, David Salinas, Lorenzo Stella, et~al.
\newblock Deep learning for time series forecasting: Tutorial and literature survey.
\newblock \emph{ACM Computing Surveys}, 55\penalty0 (6):\penalty0 1--36, 2022.

\bibitem[Salinas et~al.(2019)Salinas, Bohlke-Schneider, Callot, Medico, and Gasthaus]{salinas2019high}
David Salinas, Michael Bohlke-Schneider, Laurent Callot, Roberto Medico, and Jan Gasthaus.
\newblock High-dimensional multivariate forecasting with low-rank gaussian copula processes.
\newblock \emph{Advances in Neural Information Processing Systems}, 32, 2019.

\bibitem[Rasul et~al.(2021{\natexlab{a}})Rasul, Sheikh, Schuster, Bergmann, and Vollgraf]{rasul2020multivariate}
Kashif Rasul, Abdul-Saboor Sheikh, Ingmar Schuster, Urs Bergmann, and Roland Vollgraf.
\newblock Multivariate probabilistic time series forecasting via conditioned normalizing flows.
\newblock In \emph{International Conference on Learning Representations}, 2021{\natexlab{a}}.

\bibitem[Rasul et~al.(2021{\natexlab{b}})Rasul, Seward, Schuster, and Vollgraf]{rasul2021autoregressive}
Kashif Rasul, Calvin Seward, Ingmar Schuster, and Roland Vollgraf.
\newblock Autoregressive denoising diffusion models for multivariate probabilistic time series forecasting.
\newblock In \emph{International Conference on Machine Learning}, pages 8857--8868. PMLR, 2021{\natexlab{b}}.

\bibitem[Hyndman and Athanasopoulos(2018)]{hyndman2018forecasting}
Rob~J Hyndman and George Athanasopoulos.
\newblock \emph{Forecasting: Principles and Practice}.
\newblock OTexts, 2018.

\bibitem[Sun et~al.(2021)Sun, Lang, and Boning]{sun2021adjusting}
Fan-Keng Sun, Chris Lang, and Duane Boning.
\newblock Adjusting for autocorrelated errors in neural networks for time series.
\newblock \emph{Advances in Neural Information Processing Systems}, 34:\penalty0 29806--29819, 2021.

\bibitem[Zheng et~al.(2023)Zheng, Choi, and Sun]{zheng2023enhancing}
Vincent~Zhihao Zheng, Seongjin Choi, and Lijun Sun.
\newblock Enhancing deep traffic forecasting models with dynamic regression.
\newblock \emph{arXiv preprint arXiv:2301.06650}, 2023.

\bibitem[Zheng et~al.(2024)Zheng, Choi, and Sun]{zheng2024better}
Vincent~Zhihao Zheng, Seongjin Choi, and Lijun Sun.
\newblock Better batch for deep probabilistic time series forecasting.
\newblock In \emph{International Conference on Artificial Intelligence and Statistics}, pages 91--99. PMLR, 2024.

\bibitem[Salinas et~al.(2020)Salinas, Flunkert, Gasthaus, and Januschowski]{salinas2020deepar}
David Salinas, Valentin Flunkert, Jan Gasthaus, and Tim Januschowski.
\newblock Deepar: Probabilistic forecasting with autoregressive recurrent networks.
\newblock \emph{International Journal of Forecasting}, 36\penalty0 (3):\penalty0 1181--1191, 2020.

\bibitem[Wen et~al.(2017)Wen, Torkkola, Narayanaswamy, and Madeka]{wen2017multi}
Ruofeng Wen, Kari Torkkola, Balakrishnan Narayanaswamy, and Dhruv Madeka.
\newblock A multi-horizon quantile recurrent forecaster.
\newblock \emph{arXiv preprint arXiv:1711.11053}, 2017.

\bibitem[Rangapuram et~al.(2018)Rangapuram, Seeger, Gasthaus, Stella, Wang, and Januschowski]{rangapuram2018deep}
Syama~Sundar Rangapuram, Matthias~W Seeger, Jan Gasthaus, Lorenzo Stella, Yuyang Wang, and Tim Januschowski.
\newblock Deep state space models for time series forecasting.
\newblock \emph{Advances in Neural Information Processing Systems}, 31, 2018.

\bibitem[de~B{\'e}zenac et~al.(2020)de~B{\'e}zenac, Rangapuram, Benidis, Bohlke-Schneider, Kurle, Stella, Hasson, Gallinari, and Januschowski]{de2020normalizing}
Emmanuel de~B{\'e}zenac, Syama~Sundar Rangapuram, Konstantinos Benidis, Michael Bohlke-Schneider, Richard Kurle, Lorenzo Stella, Hilaf Hasson, Patrick Gallinari, and Tim Januschowski.
\newblock Normalizing kalman filters for multivariate time series analysis.
\newblock \emph{Advances in Neural Information Processing Systems}, 33:\penalty0 2995--3007, 2020.

\bibitem[Wang et~al.(2019)Wang, Smola, Maddix, Gasthaus, Foster, and Januschowski]{wang2019deep}
Yuyang Wang, Alex Smola, Danielle Maddix, Jan Gasthaus, Dean Foster, and Tim Januschowski.
\newblock Deep factors for forecasting.
\newblock In \emph{International Conference on Machine Learning}, pages 6607--6617. PMLR, 2019.

\bibitem[Tang and Matteson(2021)]{tang2021probabilistic}
Binh Tang and David~S Matteson.
\newblock Probabilistic transformer for time series analysis.
\newblock \emph{Advances in Neural Information Processing Systems}, 34:\penalty0 23592--23608, 2021.

\bibitem[Drouin et~al.(2022)Drouin, Marcotte, and Chapados]{drouin2022tactis}
Alexandre Drouin, {\'E}tienne Marcotte, and Nicolas Chapados.
\newblock Tactis: Transformer-attentional copulas for time series.
\newblock In \emph{International Conference on Machine Learning}, pages 5447--5493. PMLR, 2022.

\bibitem[Ashok et~al.(2024)Ashok, Marcotte, Zantedeschi, Chapados, and Drouin]{ashok2023tactis}
Arjun Ashok, {\'E}tienne Marcotte, Valentina Zantedeschi, Nicolas Chapados, and Alexandre Drouin.
\newblock Tactis-2: Better, faster, simpler attentional copulas for multivariate time series.
\newblock In \emph{International Conference on Learning Representations}, 2024.

\bibitem[Prado et~al.(2021)Prado, Ferreira, and West]{prado2021time}
Raquel Prado, Marco~AR Ferreira, and Mike West.
\newblock \emph{Time Series: Modeling, Computation, and Inference}.
\newblock CRC Press, 2021.

\bibitem[Hamilton(2020)]{hamilton2020time}
James~Douglas Hamilton.
\newblock \emph{Time Series Analysis}.
\newblock Princeton University Press, 2020.

\bibitem[Saha et~al.(2023)Saha, Basu, and Datta]{saha2023random}
Arkajyoti Saha, Sumanta Basu, and Abhirup Datta.
\newblock Random forests for spatially dependent data.
\newblock \emph{Journal of the American Statistical Association}, 118\penalty0 (541):\penalty0 665--683, 2023.

\bibitem[Choi et~al.(2022)Choi, Saunier, Zheng, Trepanier, and Sun]{choi2022scalable}
Seongjin Choi, Nicolas Saunier, Vincent~Zhihao Zheng, Martin Trepanier, and Lijun Sun.
\newblock Scalable dynamic mixture model with full covariance for probabilistic traffic forecasting.
\newblock \emph{arXiv preprint arXiv:2212.06653}, 2022.

\bibitem[Gelfand et~al.(2004)Gelfand, Schmidt, Banerjee, and Sirmans]{gelfand2004nonstationary}
Alan~E Gelfand, Alexandra~M Schmidt, Sudipto Banerjee, and CF~Sirmans.
\newblock Nonstationary multivariate process modeling through spatially varying coregionalization.
\newblock \emph{Test}, 13:\penalty0 263--312, 2004.

\bibitem[Meng et~al.(2021)Meng, Soper, Lee, Liu, Greene, and Ray]{meng2021nonstationary}
Rui Meng, Braden Soper, Herbert~KH Lee, Vincent~X Liu, John~D Greene, and Priyadip Ray.
\newblock Nonstationary multivariate gaussian processes for electronic health records.
\newblock \emph{Journal of Biomedical Informatics}, 117:\penalty0 103698, 2021.

\bibitem[Alexandrov et~al.(2020)Alexandrov, Benidis, Bohlke-Schneider, Flunkert, Gasthaus, Januschowski, Maddix, Rangapuram, Salinas, Schulz, et~al.]{alexandrov2020gluonts}
Alexander Alexandrov, Konstantinos Benidis, Michael Bohlke-Schneider, Valentin Flunkert, Jan Gasthaus, Tim Januschowski, Danielle~C Maddix, Syama Rangapuram, David Salinas, Jasper Schulz, et~al.
\newblock Gluonts: Probabilistic and neural time series modeling in python.
\newblock \emph{The Journal of Machine Learning Research}, 21\penalty0 (1):\penalty0 4629--4634, 2020.

\bibitem[Radford et~al.(2018)Radford, Narasimhan, Salimans, Sutskever, et~al.]{radford2018improving}
Alec Radford, Karthik Narasimhan, Tim Salimans, Ilya Sutskever, et~al.
\newblock Improving language understanding by generative pre-training.
\newblock 2018.

\bibitem[Beitner(2020)]{pytorchforecasting}
Jan Beitner.
\newblock Pytorch forecasting.
\newblock \url{https://pytorch-forecasting.readthedocs.io}, 2020.

\bibitem[Gneiting and Raftery(2007)]{gneiting2007strictly}
Tilmann Gneiting and Adrian~E Raftery.
\newblock Strictly proper scoring rules, prediction, and estimation.
\newblock \emph{Journal of the American Statistical Association}, 102\penalty0 (477):\penalty0 359--378, 2007.

\bibitem[Koochali et~al.(2022)Koochali, Schichtel, Dengel, and Ahmed]{koochali2022random}
Alireza Koochali, Peter Schichtel, Andreas Dengel, and Sheraz Ahmed.
\newblock Random noise vs. state-of-the-art probabilistic forecasting methods: A case study on crps-sum discrimination ability.
\newblock \emph{Applied Sciences}, 12\penalty0 (10):\penalty0 5104, 2022.

\bibitem[Marcotte et~al.(2023)Marcotte, Zantedeschi, Drouin, and Chapados]{marcotte2023regions}
{\'E}tienne Marcotte, Valentina Zantedeschi, Alexandre Drouin, and Nicolas Chapados.
\newblock Regions of reliability in the evaluation of multivariate probabilistic forecasts.
\newblock In \emph{International Conference on Machine Learning}, pages 23958--24004. PMLR, 2023.

\bibitem[Dua and Graff(2017)]{Dua:2019}
Dheeru Dua and Casey Graff.
\newblock Uci machine learning repository.
\newblock \emph{URL https://archive.ics.uci.edu/ml/index.php}, 2017.

\bibitem[Makridakis et~al.(2020)Makridakis, Spiliotis, and Assimakopoulos]{makridakis2020m4}
Spyros Makridakis, Evangelos Spiliotis, and Vassilios Assimakopoulos.
\newblock The m4 competition: 100,000 time series and 61 forecasting methods.
\newblock \emph{International Journal of Forecasting}, 36\penalty0 (1):\penalty0 54--74, 2020.

\bibitem[Lai et~al.(2018)Lai, Chang, Yang, and Liu]{lai2018modeling}
Guokun Lai, Wei-Cheng Chang, Yiming Yang, and Hanxiao Liu.
\newblock Modeling long-and short-term temporal patterns with deep neural networks.
\newblock In \emph{The 41st International ACM SIGIR Conference on Research \& Development in Information Retrieval}, pages 95--104, 2018.

\bibitem[Makridakis et~al.(1982)Makridakis, Andersen, Carbone, Fildes, Hibon, Lewandowski, Newton, Parzen, and Winkler]{makridakis1982accuracy}
Spyros Makridakis, Allan Andersen, Robert Carbone, Robert Fildes, Michele Hibon, Rudolf Lewandowski, Joseph Newton, Emanuel Parzen, and Robert Winkler.
\newblock The accuracy of extrapolation (time series) methods: Results of a forecasting competition.
\newblock \emph{Journal of Forecasting}, 1\penalty0 (2):\penalty0 111--153, 1982.

\bibitem[Chen et~al.(2001)Chen, Petty, Skabardonis, Varaiya, and Jia]{chen2001freeway}
Chao Chen, Karl Petty, Alexander Skabardonis, Pravin Varaiya, and Zhanfeng Jia.
\newblock Freeway performance measurement system: mining loop detector data.
\newblock \emph{Transportation Research Record}, 1748\penalty0 (1):\penalty0 96--102, 2001.

\bibitem[Caltrans(2015)]{traffic}
Caltrans.
\newblock Caltrans performance measurement system.
\newblock \emph{URL https://pems.dot.ca.gov/}, 2015.

\bibitem[Commission(2015)]{uber2015tlc}
NYC Taxi~Limousine Commission.
\newblock Uber tlc foil response, 2015.

\bibitem[Gasthaus et~al.(2019)Gasthaus, Benidis, Wang, Rangapuram, Salinas, Flunkert, and Januschowski]{gasthaus2019probabilistic}
Jan Gasthaus, Konstantinos Benidis, Yuyang Wang, Syama~Sundar Rangapuram, David Salinas, Valentin Flunkert, and Tim Januschowski.
\newblock Probabilistic forecasting with spline quantile function rnns.
\newblock In \emph{The 22nd International Conference on Artificial Intelligence and Statistics}, pages 1901--1910. PMLR, 2019.

\bibitem[L{\"u}tkepohl(2005)]{lutkepohl2005new}
Helmut L{\"u}tkepohl.
\newblock \emph{New Introduction to Multiple Time Series Analysis}.
\newblock Springer Science \& Business Media, 2005.

\bibitem[Van~der Weide(2002)]{van2002go}
Roy Van~der Weide.
\newblock Go-garch: a multivariate generalized orthogonal garch model.
\newblock \emph{Journal of Applied Econometrics}, 17\penalty0 (5):\penalty0 549--564, 2002.

\bibitem[Engle(2002)]{engle2002dynamic}
Robert Engle.
\newblock Dynamic conditional correlation: A simple class of multivariate generalized autoregressive conditional heteroskedasticity models.
\newblock \emph{Journal of Business \& Economic Statistics}, 20\penalty0 (3):\penalty0 339--350, 2002.

\bibitem[Seabold and Perktold(2010)]{seabold2010statsmodels}
Skipper Seabold and Josef Perktold.
\newblock Statsmodels: econometric and statistical modeling with python.
\newblock \emph{SciPy}, 7:\penalty0 1, 2010.

\bibitem[Srivastava(2022)]{mgarch}
Prashant Srivastava.
\newblock mgarch.
\newblock \url{https://pypi.org/project/mgarch/}, 2022.

\bibitem[Vaswani et~al.(2017)Vaswani, Shazeer, Parmar, Uszkoreit, Jones, Gomez, Kaiser, and Polosukhin]{vaswani2017attention}
Ashish Vaswani, Noam Shazeer, Niki Parmar, Jakob Uszkoreit, Llion Jones, Aidan~N Gomez, {\L}ukasz Kaiser, and Illia Polosukhin.
\newblock Attention is all you need.
\newblock \emph{Advances in Neural Information Processing Systems}, 30, 2017.

\bibitem[Bai et~al.(2020)Bai, Yao, Li, Wang, and Wang]{bai2020adaptive}
Lei Bai, Lina Yao, Can Li, Xianzhi Wang, and Can Wang.
\newblock Adaptive graph convolutional recurrent network for traffic forecasting.
\newblock \emph{Advances in Neural Information Processing Systems}, 33:\penalty0 17804--17815, 2020.

\bibitem[Shih et~al.(2019)Shih, Sun, and Lee]{shih2019temporal}
Shun-Yao Shih, Fan-Keng Sun, and Hung-yi Lee.
\newblock Temporal pattern attention for multivariate time series forecasting.
\newblock \emph{Machine Learning}, 108:\penalty0 1421--1441, 2019.

\bibitem[Guidolin and Pedio(2018)]{guidolin2018essentials}
Massimo Guidolin and Manuela Pedio.
\newblock \emph{Essentials of Time Series for Financial Applications}.
\newblock Academic Press, 2018.

\end{thebibliography}
}







\newpage
\appendix

\section*{Appendix}

\textbf{Table of Contents}\\
\noindent\makebox[\textwidth]{\rule{\textwidth}{0.4pt}}
\textbf{A Experimental Details} \dotfill \pageref{apx:ed}

\hspace{0.5cm} A.1 Datasets \dotfill \pageref{apx:ed_data} 

\hspace{0.5cm} A.2 Multivariate Likelihood with Autocorrelated Errors \dotfill \pageref{apx:ed_nll} 

\hspace{0.5cm} A.3 Training Procedure \dotfill \pageref{apx:ed_train} 

\hspace{0.5cm} A.4 Naive Baseline Description \dotfill \pageref{apx:ed_naive} 

\hspace{0.5cm} A.5 Hyperparameter Search \dotfill \pageref{apx:ed_hyparam} 

\hspace{0.5cm} A.6 Base Model Description and Input Features \dotfill \pageref{apx:ed_model} 

\textbf{B Metrics and Additional Results} \dotfill \pageref{apx:mar}

\hspace{0.5cm} B.1 Metric Definition \dotfill \pageref{apx:mar_metric} 

\hspace{1.0cm} B.1.1 Continuous Ranked Probability Score \dotfill \pageref{apx:mar_metric_crps} 

\hspace{1.0cm} B.1.2 Quantile Loss \dotfill \pageref{apx:mar_metric_ql} 

\hspace{1.0cm} B.1.3 Energy Score \dotfill \pageref{apx:mar_metric_es} 

\hspace{1.0cm} B.1.4 Root Relative Mean Squared Error \dotfill \pageref{apx:mar_metric_rrmse} 

\hspace{0.5cm} B.2 Results on Other Metrics \dotfill \pageref{apx:mar_results} 

\hspace{0.5cm} B.3 Training Dynamics \dotfill \pageref{apx:mar_training} 

\hspace{0.5cm} B.4 Effect of the Number of Time Series during Prediction \dotfill \pageref{apx:num_b_pred} 

\hspace{0.5cm} B.5 Additional Model Interpretation \dotfill \pageref{apx:extra_exp} 

\hspace{1.0cm} B.5.1 Comparison of Residual Correlation \dotfill \pageref{apx:extra_exp_crc} 

\hspace{1.0cm} B.5.2 Performance Breakdown at Each Forecast Step \dotfill \pageref{apx:extra_exp_pbefs} 

\hspace{0.5cm} B.6 Alternative Parametrization of $\boldsymbol{C}_t$ \dotfill \pageref{apx:ap} 

\hspace{1.0cm} B.6.1 Learnable Lengthscales \dotfill \pageref{apx:ap_ll} 

\hspace{1.0cm} B.6.2 Using Autocorrelations of an AR\((p)\) process \dotfill \pageref{apx:ap_ar} 

\hspace{0.5cm} B.7 Alternative Error Assumptions \dotfill \pageref{apx:ana} 

\hspace{0.5cm} B.8 Qualitative Results on Forecasting \dotfill \pageref{apx:mar_qualitative} 
\noindent\makebox[\textwidth]{\rule{\textwidth}{0.4pt}}

\section{Experimental Details}\label{apx:ed}

\subsection{Datasets}\label{apx:ed_data}

We performed experiments on a diverse set of real-world datasets obtained from GluonTS \citep{alexandrov2020gluonts}. These datasets include:
\begin{itemize}[nosep, noitemsep]
    \item $\mathtt{electricity}$ \citep{Dua:2019}: Hourly electricity consumption data collected from a total of 370 households over the period spanning from January 2012 to June 2014.
    \item $\mathtt{m4\_hourly}$ \citep{makridakis2020m4}: Hourly time series data from various domains, covering microeconomics, macroeconomics, finance, industry, demographics, and various other fields, are sourced from the M4-competition.
    \item $\mathtt{exchange\_rate}$ \citep{lai2018modeling}: Daily exchange rate information for eight different countries spanning the period from 1990 to 2016.
    \item $\mathtt{m1\_quarterly}$ \citep{makridakis1982accuracy}: Quarterly time series data spanning seven different domains.
    \item $\mathtt{pems03}$ \citep{chen2001freeway}: Traffic flow records obtained from Caltrans District 3 and accessed through the Caltrans Performance Measurement System (PeMS). The records are aggregated at a 5-minute interval.
    \item $\mathtt{solar}$ \citep{lai2018modeling}: Hourly time series representing solar power production data in the state of Alabama for the year 2006.
    \item $\mathtt{traffic}$ \citep{traffic}: Hourly traffic occupancy rates recorded by sensors installed in the San Francisco freeway system between January 2008 and June 2008.
    \item $\mathtt{uber\_hourly}$ \citep{uber2015tlc}: Hourly time series of Uber pickups in New York City spanning from February to July 2015.
    \item $\mathtt{wikipedia}$ \citep{gasthaus2019probabilistic}: Daily page views for 2,000 Wikipedia pages spanning from January 2012 to March 2014.
\end{itemize}

These datasets are widely employed for benchmarking time series forecasting models, following their default configurations in GluonTS, including granularity, prediction range ($Q$), and the number of rolling evaluations. For each dataset, we performed a sequential split into training, validation, and testing sets, with the temporal length of the validation set matching that of the testing set. The temporal length of the testing set was determined by considering the prediction range and the required number of rolling evaluations. For instance, the testing horizon for the $\mathtt{traffic}$ dataset is computed as $24+7-1=30$ time steps. As a result, the model will predict 24 steps ($Q$) sequentially, with 7 distinct consecutive prediction start timestamps, also known as 7 forecast instances. In our experiments, we align the conditioning range ($P$) with the prediction range ($Q$), maintaining consistency with the default setting in GluonTS. For simplicity, we set the autocorrelation horizon ($D$) to also match the prediction range ($Q$). Essentially, in this paper, we have $P=Q=D$. Each dataset was standardized using the mean and standard deviation from the training set. Predictions were rescaled to their original values for computing evaluation metrics. The statistics of all datasets are summarized in Table~\ref{tab:datasets}.

\begin{table}[!h]
\small
  \caption{Dataset summary.}
  \label{tab:datasets}
  \centering
  \begin{tabular}{lccccc}
    \toprule
    Dataset     &   Granularity   &  \# of time series & \# of time steps & $Q$ & Rolling evaluation \\
    \midrule
    $\mathtt{electricity}$   &  hourly & 370   & 5,857 & 24 & 7 \\
    $\mathtt{m4\_hourly}$     &  hourly & 414   & 1,008 & 48 & 7 \\
    $\mathtt{exchange\_rate}$     &  workday & 8   & 6,101 & 30 & 5 \\
    $\mathtt{m1\_quarterly}$     &  quarterly & 281  & 48 & 8 & 1 \\
    $\mathtt{pems03}$     &  5min & 358  & 26,208 & 12 & 24 \\
    $\mathtt{solar}$     &  hourly & 137  & 7,033 & 24 & 7 \\
    $\mathtt{traffic}$       &  hourly & 963   & 4,025 & 24 & 7 \\
    $\mathtt{uber\_hourly}$           &  hourly  & 262   & 8,343 & 24 & 7   \\
    $\mathtt{wikipedia}$           &  daily  & 2,000   & 792 & 30 & 5   \\
    \bottomrule
  \end{tabular}
\end{table}

\subsection{Multivariate Likelihood with Correlated Errors}\label{apx:ed_nll}
The probability density function of a multivariate normal distribution with autocorrelated errors, as described in \S \ref{sec:training}, is defined in Eq.~\eqref{eqn:bat_pdf}. For simplicity, we omit the subscript \(t\) and superscript \(\text{bat}\) for all notations:
\begin{equation}\label{eqn:bat_pdf}
    f(\boldsymbol{z})=(2\pi)^{-B/2}\lvert\boldsymbol{\Sigma}\rvert^{-1/2}\exp{\left(-\frac{1}{2}\left(\boldsymbol{z}-\boldsymbol{\mu}\right)^\top\boldsymbol{\Sigma}^{-1}\left(\boldsymbol{z}-\boldsymbol{\mu}\right)\right)}.
\end{equation}

We use the negative log likelihood (NLL) of an observed $\boldsymbol{z}$ as the loss function for training our model. The NLL can be calculated as the negative log of the probability density function in Eq.~\eqref{eqn:bat_pdf}:
\begin{equation}\label{eqn:bat_nll}
    \mathcal{L}_{NLL}=-\ln{L\left(\boldsymbol{z}\right)}=\frac{1}{2}\left[\ln\lvert\boldsymbol{\Sigma}\rvert+\left(\boldsymbol{z}-\boldsymbol{\mu}\right)^\top\boldsymbol{\Sigma}^{-1}\left(\boldsymbol{z}-\boldsymbol{\mu}\right)+B\ln\left(2\pi\right)\right],
\end{equation}
where \(B\) is the number of time series in a batch. The covariance matrix is parameterized as \(\boldsymbol{\Sigma}=\boldsymbol{L}(\boldsymbol{C} \otimes \mathbf{I}_{R})\boldsymbol{L}^\top+\boldsymbol{E}\). In this parameterization, \(\boldsymbol{L} \in \mathbb{R}^{DB \times DR}\) is the covariance factor, \(\boldsymbol{C} \in \mathbb{R}^{D \times D}\) is the autocorrelation matrix, \(\boldsymbol{E}=\diag(\boldsymbol{d})\), and \(\boldsymbol{d} \in \mathbb{R}_{+}^{DB}\) are the diagonal elements. The bottleneck in evaluating this NLL lies in the calculation of the inverse and determinant of \(\boldsymbol{\Sigma}\). Therefore, we can simplify the calculation using the Sherman–Morrison–Woodbury identity (matrix inversion lemma) and the companion matrix determinant lemma:
\begin{equation}\label{eqn:bat_inv}
\begin{aligned}
    \boldsymbol{\Sigma}^{-1}&=\left(\boldsymbol{E}+\boldsymbol{L}\left(\boldsymbol{C} \otimes \mathbf{I}_{R}\right)\boldsymbol{L}^\top\right)^{-1}\\
    &=\boldsymbol{E}^{-1}-\boldsymbol{E}^{-1}\boldsymbol{L}\left(\left(\boldsymbol{C} \otimes \mathbf{I}_{R}\right)^{-1}+\boldsymbol{L}^\top\boldsymbol{E}^{-1}\boldsymbol{L}\right)^{-1}\boldsymbol{L}^\top\boldsymbol{E}^{-1}.
\end{aligned}
\end{equation}

Consequently, the Mahalanobis term in Eq.~\eqref{eqn:bat_nll} becomes:
\begin{equation}\label{eqn:bat_mahalanobis}
\begin{aligned}
    \boldsymbol{\eta}^\top\boldsymbol{\Sigma}^{-1}\boldsymbol{\eta}=&\boldsymbol{\eta}^\top\boldsymbol{E}^{-1}\boldsymbol{\eta}-\boldsymbol{\eta}^\top\boldsymbol{E}^{-1}\boldsymbol{L}\left(\left(\boldsymbol{C} \otimes \mathbf{I}_{R}\right)^{-1}+\boldsymbol{L}^\top\boldsymbol{E}^{-1}\boldsymbol{L}\right)^{-1}\boldsymbol{L}^\top\boldsymbol{E}^{-1}\boldsymbol{\eta}\\
    =&\boldsymbol{\eta}^\top\boldsymbol{E}^{-1}\boldsymbol{\eta}-\boldsymbol{\eta}^\top\boldsymbol{E}^{-1}\boldsymbol{L}\left(\boldsymbol{L}_{cap}\boldsymbol{L}_{cap}^\top\right)^{-1}\boldsymbol{L}^\top\boldsymbol{E}^{-1}\boldsymbol{\eta}\\
    =&\boldsymbol{\eta}^\top\boldsymbol{E}^{-1}\boldsymbol{\eta}-\left(\boldsymbol{L}_{cap}^{-1}\boldsymbol{L}^\top\boldsymbol{E}^{-1}\boldsymbol{\eta}\right)^\top\left(\boldsymbol{L}_{cap}^{-1}\boldsymbol{L}^\top\boldsymbol{E}^{-1}\boldsymbol{\eta}\right)\\
    =&\boldsymbol{\eta}^\top\boldsymbol{E}^{-1}\boldsymbol{\eta}-\boldsymbol{k}^\top\boldsymbol{k},
\end{aligned}
\end{equation}
where \(\boldsymbol{k}=\boldsymbol{L}_{cap}^{-1}\boldsymbol{L}^\top\boldsymbol{E}^{-1}\boldsymbol{\eta}\). \(\boldsymbol{L}_{cap}\) is the Cholesky factor of the capacitance matrix \(\left(\left(\boldsymbol{C} \otimes \mathbf{I}_{R}\right)^{-1}+\boldsymbol{L}^\top\boldsymbol{E}^{-1}\boldsymbol{L}\right)\). The computation of \(\boldsymbol{k}\) can be efficiently resolved by solving the linear system of equations \(\boldsymbol{L}_{cap}\boldsymbol{k}=\boldsymbol{L}^\top\boldsymbol{E}^{-1}\boldsymbol{\eta}\). Since \(\boldsymbol{E}\) is a diagonal matrix, the only matrix inverse we need to calculate in Eq.~\eqref{eqn:bat_mahalanobis} is \(\left(\boldsymbol{C} \otimes \mathbf{I}_{R}\right)^{-1}\), which can be further simplified as \(\boldsymbol{C}^{-1} \otimes \mathbf{I}_{R}\). Recall that \(\boldsymbol{C}\) is a \(D \times D\) autocorrelation matrix. Therefore, calculating its inverse is much easier than computing the inverse of \(\boldsymbol{\Sigma}\), which is a \(DB \times DB\) matrix. Moreover, the computational cost does not scale with the number of time series \(B\) in a batch.

The calculation of the determinant can also be greatly simplified with our parameterization:
\begin{equation}\label{eqn:bat_det}
\begin{aligned}
    \ln\lvert\boldsymbol{\Sigma}\rvert&=\ln\lvert\boldsymbol{E}+\boldsymbol{L}\left(\boldsymbol{C} \otimes \mathbf{I}_{R}\right)\boldsymbol{L}^\top\rvert\\
    &=\ln\lvert\left(\boldsymbol{C} \otimes \mathbf{I}_{R}\right)^{-1}+\boldsymbol{L}^\top\boldsymbol{E}^{-1}\boldsymbol{L}\rvert + \ln\lvert\boldsymbol{C} \otimes \mathbf{I}_{R}\rvert + \ln\lvert\boldsymbol{E}\rvert\\
    &=2\sum^{DR}_i\ln\left[\boldsymbol{L}_{cap}\right]_{i,i}+2R\sum^{D}_i\ln\left[\boldsymbol{L}_{C}\right]_{i,i}+\sum^{DB}_i\ln\left[\boldsymbol{E}\right]_{i,i},
\end{aligned}
\end{equation}
where \(\boldsymbol{L}_{C}\) is the Cholesky factor of the autocorrelation matrix \(\boldsymbol{C}\).


\subsection{Training Procedure}\label{apx:ed_train}

\textbf{Compute used} All models in the paper were trained in an Anaconda environment with access to one AMD Ryzen Threadripper PRO 5955WX CPU and four NVIDIA RTX A5000 GPUs (each with 24 GB of memory). 

\textbf{Batch size} We adopt the approach of GPVar \citep{salinas2019high} by using $B=20$ time series in a sample slice and a batch size of 16. Because our data sampler selects one slice of time series as a batch instead of sampling 16 slices simultaneously, we set $\mathtt{accumulate\_grad\_batches}$ to 16 to achieve an effective batch size of 16.

\textbf{Training loop} Each epoch involves training the model on up to 400 batches from the training set, followed by computing the NLL on the validation set. Training stops when any of the following conditions are met:
\begin{itemize}[nosep, noitemsep, leftmargin=*]
    \item A total of 10,000 gradient updates have been performed during model training,
    \item No improvement in the best NLL value on the validation set is observed for 10 consecutive epochs.
\end{itemize}
We select the version of the model that achieved the best NLL value on the validation set.

\subsection{Naive Baseline Description}\label{apx:ed_naive}
In this paper, we employ VAR \citep{lutkepohl2005new} (Vector Autoregression) and GARCH \citep{van2002go} (Generalized Autoregressive Conditionally Heteroskedasticity) as two naive baseline models. The VAR(\(p\)) model is defined as
\begin{equation}\label{var}
    \mathbf{z}_{t} = \mathbf{c} + \boldsymbol{A}_1\mathbf{z}_{t-1} + \dots + \boldsymbol{A}_p\mathbf{z}_{t-p} + \boldsymbol{\epsilon}_t, \boldsymbol{\epsilon}_t \sim \mathcal{N}(\mathbf{0},\boldsymbol{\Sigma}_{\epsilon}),
\end{equation}
where \(A_i\) is an \(N \times N\) coefficient matrix, and \(\mathbf{c}\) is the intercept. We use a VAR model of lag 1 (i.e., a VAR(1) model) in the experiments. The parameters of Eq.~\eqref{var} are estimated using ordinary least squares (OLS), following the procedure in \citep{lutkepohl2005new}.

The GARCH model describes the conditional covariance matrix of the error term in a multivariate system. Suppose the model for the conditional mean is an AR(\(1\)) model:
\begin{equation}
    \mathbf{z}_{t} = \mathbf{c} + \boldsymbol{A}_1\mathbf{z}_{t-1} + \boldsymbol{\epsilon}_t,
\end{equation}
where the error term is modeled as
\begin{equation}
    \boldsymbol{\epsilon}_t = \boldsymbol{H}_t^{1/2}\mathbf{e}_{t},
\end{equation}
where \(\boldsymbol{H}_t\) is an \(N \times N\) conditional covariance matrix, and \(\mathbf{e}_{t}\) is an \(N \times 1\) standard normal vector, \(\mathbf{e}_{t} \sim \mathcal{N}(\mathbf{0},\mathbf{I}_{N})\). In the experiments, we use the DCC-GARCH\((1,1)\) model \citep{engle2002dynamic}, where the conditional covariance matrix  \(\boldsymbol{H}_t\) is defined as
\begin{equation}
    \boldsymbol{H}_t = \boldsymbol{D}_t\boldsymbol{R}_t\boldsymbol{D}_t,
\end{equation}
where \(\boldsymbol{D}_t=\diag{(\mathbf{h}_{t})}^{1/2}\), and \(\mathbf{h}_{t}\) contains the variances for each time series. \(\boldsymbol{R}_t\) is the conditional correlation matrix in the DCC-GARCH model. The parameters of the DCC-GARCH model are estimated with the log-likelihood function:
\begin{equation}
    \mathcal{L} = -\frac{1}{2}\sum_{t=1}^T\left[N\ln{(2\pi)}+2\ln\lvert\boldsymbol{D}_t\rvert+\ln\lvert\boldsymbol{R}_t\rvert+\mathbf{e}_{t}^\top\boldsymbol{R}_t^{-1}\mathbf{e}_{t}\right].
\end{equation}
In this paper, we implement the VAR model using $\mathtt{statsmodels}$ \citep{seabold2010statsmodels} and the DCC-GARCH model using $\mathtt{mgarch}$ \citep{mgarch}.

\subsection{Hyperparameter Search}\label{apx:ed_hyparam}
The hyperparameters and training configuration largely align with those used in the GPVar paper \citep{salinas2019high}. All DL models are trained using the Adam optimizer with \(l2\) regularization set to 1e-8, and gradients are clipped at 10.0. For all methods, we limit the total number of gradient updates to 10,000 and decay the learning rate by a factor of 2 after 500 consecutive updates without improvement. Table~\ref{tab:hyparam} lists the parameters that are tuned, as well as the hyperparameters that are kept constant across all datasets and not subject to tuning.

\begin{table}[htbp]
\small
\caption{Hyperparameters values that are fixed or searched over a range during hyperparameter tuning.}\label{tab:hyparam}
\begin{center}
\begin{tabular}{l|c}
\toprule
Hyperparameter             & Value or Range Searched   \\
\midrule
learning rate              & [1e-4, 1e-3, 1e-2]        \\
LSTM cells / $\mathtt{d\_model}$ of Transformer                 & [10, 20, 40]              \\
LSTM layers / Transformer decoder layers               & 2                \\
$\mathtt{n\_heads}$ (Transformer)      & 2                 \\
rank                       & 10                \\
sampling dimension         & 20                         \\
dropout                    & 0.01                         \\
batch size                 & 16                        \\
\bottomrule
\end{tabular}
\end{center}
\end{table}

To tune the hyperparameters of each model, we conduct a grid search over nine parameters on each dataset. The best hyperparameters for each base model–dataset combination are selected based on the lowest validation loss. Once the optimal learning rate and hidden size are determined, we apply the same hyperparameters to models both with and without our method. 

The number of base kernels (\(M\)) for generating $\boldsymbol{C}_t$ and the associated lengthscale set \(\{l_m\}_{m=1}^{M-1}\) are two additional hyperparameters when applying our method. The optimal values of \(M\) and \(\{l_m\}_{m=1}^{M-1}\) are selected in a similar manner via hyperparameter search. The values of \(M\) and \(\{l_m\}_{m=1}^{M-1}\) explored during hyperparameter tuning are shown in Table~\ref{tab:hyparam_auto}. There are six possible combinations. For example, if we set \(M=3\) and choose the initial lengthscale to be \(1.0\), the lengthscales for generating the component kernels will be \(\{1.0 ,2.0\}\) since the last weight corresponds to the identity matrix.

\begin{table}[htbp]
\small
\caption{Hyperparameters values of our method that are searched over a range during hyperparameter tuning.}\label{tab:hyparam_auto}
\begin{center}
\begin{tabular}{l|c}
\toprule
Hyperparameter                       & Value or Range Searched   \\
\midrule
number of kernels \(M\)              & [2, 3, 4]        \\
possible lengthscales \(\{l_m\}_{m=1}^{M-1}\)       & [\(\{0.5 ,1.5, \dots\}\), \(\{1.0 ,2.0, \dots\}\)]              \\
\bottomrule
\end{tabular}
\end{center}
\end{table}

\subsection{Base Model Description and Input Features}\label{apx:ed_model}

The input to the base models consists of lagged time series values and generic features that encode time and identify each time series. The number of lagged values used is determined by the time-frequency of each dataset. Specifically, we use lags [1, 24, 168] for hourly data; [1, 7, 14] for daily data; and [1, 2, 4, 12, 24, 48] for data with a granularity of less than one hour. For all other datasets, we only use the lag-1 values.

We use generic features to represent time. For datasets with a granularity of one hour or less, we include features for the hour of the day and the day of the week. For daily datasets, we use the day of the week feature. Additionally, each time series is distinguished by an identifier number. All features are encoded with a single value; for example, the hour of the day feature takes values in [0, 23]. These feature values are concatenated with the RNN or Transformer input at each time step to generate the model input vector \(\mathbf{y}_{t}\). 

As illustrated in \S \ref{sec:methods}, our method requires a state vector \(\mathbf{h}_{t}\) at each time step to generate the parameters for the predictive distribution and the dynamic weights for correlation matrix kernels. We use two different neural architectures for this purpose: RNN and Transformer, both of which preserve autoregressive properties. Specifically, we use an LSTM as our base model for the RNN and a decoder-only Transformer (i.e., the GPT model \citep{radford2018improving}) for the Transformer. Table~\ref{tab:gpvar_param} and Table~\ref{tab:gpt_param} summarize the number of parameters for the GPVar and Transformer models across each dataset.

\begin{table*}[htbp]
\scriptsize
\caption{Number of parameters of the GPVar model for each dataset.}
\label{tab:gpvar_param}
\begin{center}
\begin{tabular}{lcccc}
\toprule
                          & covariate embedding & rnn   & distribution proj & covariance proj (our method) \\
\cmidrule(lr){1-5}
$\mathtt{exchange\_rate}$ & 60        & 6.1k  & 252                    & 84                                \\
$\mathtt{solar}$          & 3.7k      & 26.6k & 492                    & 164                               \\
$\mathtt{electricity}$    & 16.5k     & 29.6k & 492                    & 164                               \\
$\mathtt{traffic}$        & 72.5k     & 34.6k & 492                    & 164                               \\
$\mathtt{wikipedia}$      & 200k      & 5.7k  & 132                    & 44                                \\
$\mathtt{m4\_hourly}$     & 19.7k     & 10.2k & 252                    & 84                                \\
$\mathtt{m1\_quarterly}$  & 6.3k      & 25k   & 492                    & 164                               \\
$\mathtt{pems03}$         & 26.4k     & 34.6k & 492                    & 164                               \\
$\mathtt{uber\_hourly}$   & 9.7k      & 28.3k & 492                    & 164                               \\
\bottomrule
\end{tabular}
\end{center}
\end{table*}

\begin{table*}[htbp]
\scriptsize
\caption{Number of parameters of the Transformer model for each dataset.}
\label{tab:gpt_param}
\begin{center}
\begin{tabular}{lcccccc}
\toprule
                          & target proj  & covariate proj & covariate embedding & transformer & distribution proj & covariance proj (our method) \\
\cmidrule(lr){1-7}
$\mathtt{exchange\_rate}$ & 160          & 400             & 60        & 26.5k       & 492                    & 164                               \\
$\mathtt{solar}$          & 40           & 400             & 3.7k      & 1.8k        & 132                    & 44                                \\
$\mathtt{electricity}$    & 160          & 2.4k            & 16.5k     & 26.5k       & 492                    & 164                               \\
$\mathtt{traffic}$        & 80           & 1.8k            & 72.5k     & 6.8k        & 252                    & 84                                \\
$\mathtt{wikipedia}$      & 160          & 4.2k            & 200k      & 26.5k       & 492                    & 164                               \\
$\mathtt{m4\_hourly}$     & 80           & 1.2k            & 19.7k     & 6.8k        & 252                    & 84                                \\
$\mathtt{m1\_quarterly}$  & 80           & 1.3k            & 6.3k      & 26.5k       & 492                    & 164                               \\
$\mathtt{pems03}$         & 70           & 870             & 26.4k     & 1.8k        & 132                    & 44                                \\
$\mathtt{uber\_hourly}$   & 160          & 2k              & 9.7k      & 26.5k       & 492                    & 164                               \\
\bottomrule
\end{tabular}
\end{center}
\end{table*}

LSTM, a type of RNN architecture, is designed to model sequences and time series data. Unlike traditional RNNs, LSTMs can learn long-term dependencies, making them effective for tasks requiring context and memory over long sequences. A decoder-only Transformer is primarily used for sequence generation tasks, such as text generation, language modeling, and machine translation. It is a simplified version of the original Transformer model introduced by \citet{vaswani2017attention}, consisting of only the decoder component. The LSTM model can be formulated as

\begin{equation}\label{eqn:lstm}
\begin{aligned}
\mathbf{f}_{t} &= \sigma(\mathbf{W}_f \cdot [\mathbf{h}_{t-1}, \mathbf{y}_{t}] + \mathbf{b}_f), \\
\mathbf{i}_{t} &= \sigma(\mathbf{W}_i \cdot [\mathbf{h}_{t-1}, \mathbf{y}_{t}] + \mathbf{b}_i), \\
\tilde{\mathbf{C}}_{t} &= \tanh(\mathbf{W}_C \cdot [\mathbf{h}_{t-1}, \mathbf{y}_{t}] + \mathbf{b}_C), \\
\mathbf{C}_{t} &= \mathbf{f}_{t} \odot \mathbf{C}_{t-1} + \mathbf{i}_{t} \odot \tilde{\mathbf{C}}_{t}, \\
\mathbf{o}_{t} &= \sigma(\mathbf{W}_o \cdot [\mathbf{h}_{t-1}, \mathbf{y}_{t}] + \mathbf{b}_o), \\
\mathbf{h}_{t} &= \mathbf{o}_{t} \odot \tanh(\mathbf{C}_{t}),
\end{aligned}
\end{equation}
where \(\mathbf{y}_{t}\) is the input at each time step. The decoder-only Transformer can be formulated as
\begin{equation}
\begin{aligned}
\mathbf{Q} &= \mathbf{Y}_{t} \mathbf{W}_Q, \\
\mathbf{K} &= \mathbf{Y}_{t} \mathbf{W}_K, \\
\mathbf{V} &= \mathbf{Y}_{t} \mathbf{W}_V, \\
\mathbf{M} &= \text{Mask}(\mathbf{K}), \\
\mathbf{Z} &= \text{Softmax} \left( \frac{\mathbf{Q} \mathbf{K}^T}{\sqrt{d_k}} + \mathbf{M} \right) \mathbf{V}, \\
\mathbf{H}_t &= \text{LayerNorm}(\mathbf{Y}_t + \mathbf{Z}), \\
\mathbf{FFN} &= \text{ReLU}(\mathbf{H}_t \mathbf{W}_1 + \mathbf{b}_1) \mathbf{W}_2 + \mathbf{b}_2, \\
\mathbf{H}_{t} &= \text{LayerNorm}(\mathbf{H}_t + \mathbf{FFN}),
\end{aligned}
\end{equation}
where \(\mathbf{H}_{t}\) is the output containing state vectors for all time steps, and \(\mathbf{M}\) is a square causal mask for the sequence to preserve autoregressive properties.

\section{Metrics and Additional Results}\label{apx:mar}

\subsection{Metric Definition}\label{apx:mar_metric}
In this paper, we repeated the evaluation process on the testing set ten times to compute the mean and standard deviation of all metrics. Metrics calculated in each independent evaluation are based on the average results from all forecast instances in the testing set. For example, the \(\operatorname{CRPS}_{\text{sum}}\) reported for $\mathtt{traffic}$ is the average \(\operatorname{CRPS}_{\text{sum}}\) of seven forecast instances in its testing set. 100 prediction samples were drawn for all evaluation processes. 

\subsubsection{Continuous Ranked Probability Score}\label{apx:mar_metric_crps}
The Continuous Ranked Probability Score (\(\operatorname{CRPS}\)) is defined as: 
\begin{equation}
    \operatorname{CRPS}\left(F, z\right)=\E_F\lvert Z-z\rvert-\frac{1}{2} \E_F\lvert Z-Z^{\prime}\rvert,
\end{equation}
where $F$ is the cumulative distribution function (CDF) of the predicted variable, $z$ is the observation, $Z$ and $Z^{\prime}$ are independent copies of a set of prediction samples associated with the distribution $F$. For a single forecast instance, we calculate the average \(\operatorname{CRPS}\) across time series and over the prediction horizon:
\begin{equation}
    \E_{i,t}\left[\operatorname{CRPS}\left(F_{i,t}, z_{i,t}\right)\right],
\end{equation}
where we use the empirical CDF to represent \(F_{i,t}\) when predicting \(z_{i,t}\). Since \(\operatorname{CRPS}\) only compares a single ground-truth value to its predicted distribution, we also calculate the \(\operatorname{CRPS}_{\text{sum}}\) \citep{salinas2019high,drouin2022tactis,ashok2023tactis} to assess multivariate dependencies in the time series data. \(\operatorname{CRPS}_{\text{sum}}\) is computed by summing both the forecasted and ground-truth values across all time series and then calculating the \(\operatorname{CRPS}\) over the resulting sums:
\begin{equation}
    \E_{t}\left[\operatorname{CRPS}\left(F_{t}, \sum_iz_{i,t}\right)\right],
\end{equation}
where the empirical \(F_{t}\) is obtained by summing samples across time series. 

\subsubsection{Quantile Loss}\label{apx:mar_metric_ql}
The Quantile Loss ($\rho$-risk) is another metric used in \citep{salinas2020deepar} to evaluate the performance of  probabilistic forecasting:
\begin{equation}
    L_\rho\left(z, \hat{z}^\rho\right)=2\left(\hat{z}^\rho-z\right)\left(\left(1-\rho\right) \mathrm{I}_{\hat{z}^\rho>z}-\rho \mathrm{I}_{\hat{z}^\rho \leq z}\right),
\end{equation}
where $\mathrm{I}$ is a binary indicator function that equals 1 when the condition is met, $\hat{z}^\rho$ represents the predicted $\rho$-quantile, and $z$ represents the ground truth value. The quantile loss serves as a metric to assess the accuracy of a given quantile $\rho$ from the predictive distribution. We summarize the quantile losses over the testing set across all time series segments by computing a normalized summation of these losses: $\left(\sum_{i,t} L_\rho\left(z_{i,t}, \hat{z}_{i,t}^\rho\right)\right) /\left(\sum_{i,t} z_{i,t}\right)$. In this paper, we evaluate the $0.5$-risk and the $0.9$-risk following \citet{salinas2020deepar}. 

\subsubsection{Energy Score}\label{apx:mar_metric_es}
The Energy Score (\(\operatorname{ES}\)) generalizes the \(\operatorname{CRPS}\) to evaluate distributional forecasts of a vector-valued random variable and is thus another multivariate metric used in this paper:
\begin{equation}
    \operatorname{ES}(P, \mathbf{z})=\mathop{\E}_{\boldsymbol{Z}\sim P}\lVert\boldsymbol{Z}-\mathbf{z}\rVert_2^{\beta}-\frac{1}{2} \mathop{\E}_{\substack{\boldsymbol{Z}\sim P\\\boldsymbol{Z}^{\prime}\sim P}}\lVert\boldsymbol{Z}-\boldsymbol{Z}^{\prime}\rVert_2^{\beta},
\end{equation}
where \(\lVert\mathbf{z}\rVert_2\) is the Euclidean norm. In this paper, we use \(\beta=1\), following \citep{ashok2023tactis}. Since we also want to aggregate over the prediction horizon, we calculate the Frobenius norm of the matrix \(\lVert\mathbf{z}_{t+1:t+Q}\rVert_F\) in practice.

\subsubsection{Root Relative Mean Squared Error}\label{apx:mar_metric_rrmse}
The Root Relative Mean Squared Error (\(\operatorname{RRMSE}\)) is a metric commonly used for point forecasts \citep{bai2020adaptive,lai2018modeling,shih2019temporal}. \(\operatorname{RRMSE}\) is defined as:
\begin{equation}
    \operatorname{RRMSE} = \frac{\sqrt{\sum_{t=1}^Q \lVert \mathbf{z}_t - \hat{\mathbf{z}}_t\rVert_2^2}}{\sqrt{\sum_{t=1}^Q \lVert\mathbf{z}_t - \bar{\mathbf{z}}\rVert_2^2}},
\end{equation}
where $\hat{\mathbf{z}}_t$ is obtained by taking the mean of our prediction samples, and $\bar{\mathbf{z}}$ is the mean value of the entire forecast instance. We use this metric to evaluate the mean prediction performance of our model.

\subsection{Results on Other Forecasting Metrics}\label{apx:mar_results}
We present the results for \(\operatorname{CRPS}\) (Table~\ref{tab:crps}), the $0.5$-risk (Table~\ref{tab:p05}), the $0.9$-risk (Table~\ref{tab:p09}), \(\operatorname{ES}\) (Table~\ref{tab:es}), and \(\operatorname{RRMSE}\) (Table~\ref{tab:rrmse}). An ``N/A'' entry in the tables indicates that the naive baseline models could not be properly fitted to this dataset. We observe consistent performance improvements in the base models using our method across different evaluation metrics. Notably, in the multivariate metric \(\operatorname{ES}\), our method shows significant improvement, reducing the score by an average of 5.58\% for GPVar and 3.21\% for the Transformer.

\begin{table*}[htbp]
\scriptsize
\caption{Comparison of $\operatorname{CRPS}$ accuracy. ``w/o'' denotes methods without time-dependent errors, while ``w/'' indicates our method. Boldface values indicate that models considering time-dependent errors have better performance. Mean and standard deviation are obtained from 10 runs of each model.}
\label{tab:crps}
\begin{center}
\begin{tabular}{lcccccc}
\toprule
                          & VAR     & GARCH       & \multicolumn{2}{c}{GPVar}              & \multicolumn{2}{c}{Transformer} \\
\cmidrule(lr){2-7}
                          &       &      & w/o              & w/           & w/o              & w/   \\
\midrule
$\mathtt{exchange\_rate}$ & 0.0070±0.0000   & 0.0438±0.0001 & 0.0171±0.0004 & 0.0141±0.0003 & 0.0092±0.0002 & 0.0081±0.0001 \\
$\mathtt{solar}$          & 0.9566±0.0022   & 0.9193±0.0010 & 0.7097±0.0047 & 0.7521±0.0027 & 0.5981±0.0021 & 0.5627±0.0018 \\
$\mathtt{electricity}$    & 0.1548±0.0003   & 0.2778±0.0010 & 0.0586±0.0004 & 0.0568±0.0002 & 0.0665±0.0003 & 0.0775±0.0001 \\
$\mathtt{traffic}$        & 19.9208±0.0495  & 0.4063±0.0002 & 0.1474±0.0001 & 0.1296±0.0001 & 0.1260±0.0001 & 0.1318±0.0001 \\
$\mathtt{wiki}$           & 334.6021±0.4936 & 3.0351±0.0048 & 0.3712±0.0003 & 0.3705±0.0004 & 0.3737±0.0003 & 0.2937±0.0002 \\
$\mathtt{m4\_hourly}$     & 0.2837±0.0004   & 0.3567±0.0004 & 0.1174±0.0002 & 0.1237±0.0002 & 0.1306±0.0002 & 0.1189±0.0002 \\
$\mathtt{m1\_quarterly}$  & N/A             & N/A           & 0.3942±0.0030 & 0.3538±0.0017 & 0.4448±0.0027 & 0.4367±0.002  \\
$\mathtt{pems03}$         & 0.1144±0.0001   & 0.3533±0.0002 & 0.0828±0.0000 & 0.0835±0.0001 & 0.0826±0.0001 & 0.0735±0.0000 \\
$\mathtt{uber\_hourly}$   & N/A             & N/A           & 0.1488±0.0003 & 0.1468±0.0002 & 0.1576±0.0003 & 0.1762±0.0003 \\
\cmidrule(lr){1-7}
&   &    &  \multicolumn{1}{r}{avg. rel. impr.} & \multicolumn{1}{l}{3.59\%}  & \multicolumn{1}{r}{avg. rel. impr.} & \multicolumn{1}{l}{3.13\%}  \\
\bottomrule
\end{tabular}
\end{center}
\end{table*}

\begin{table*}[htbp]
\scriptsize
\caption{Comparison of $0.5$-risk accuracy. ``w/o'' denotes methods without time-dependent errors, while ``w/'' indicates our method. Boldface values indicate that models considering time-dependent errors have better performance. Mean and standard deviation are obtained from 10 runs of each model.}
\label{tab:p05}
\begin{center}
\begin{tabular}{lcccccc}
\toprule
                          & VAR     & GARCH       & \multicolumn{2}{c}{GPVar}              & \multicolumn{2}{c}{Transformer} \\
\cmidrule(lr){2-7}
                          &       &      & w/o              & w/           & w/o              & w/   \\
\midrule
$\mathtt{exchange\_rate}$ & 0.0049±0.0000   & 0.0256±0.0001 & 0.0109±0.0003 & 0.0095±0.0004 & 0.0060±0.0001 & 0.0056±0.0001 \\
$\mathtt{solar}$          & 0.6140±0.0025   & 0.5621±0.0008 & 0.4998±0.0025 & 0.5246±0.0016 & 0.4233±0.0017 & 0.3958±0.0013 \\
$\mathtt{electricity}$    & 0.1113±0.0005   & 0.2014±0.0010 & 0.0405±0.0003 & 0.0397±0.0002 & 0.0449±0.0002 & 0.0505±0.0001 \\
$\mathtt{traffic}$        & 10.2654±0.0268  & 0.2722±0.0002 & 0.0933±0.0001 & 0.0859±0.0001 & 0.0803±0.0001 & 0.0794±0.0001 \\
$\mathtt{wiki}$           & 171.5009±0.2573 & 0.7225±0.0067 & 0.2231±0.0005 & 0.2236±0.0006 & 0.2030±0.0005 & 0.1487±0.0003 \\
$\mathtt{m4\_hourly}$     & 0.1992±0.0003   & 0.2365±0.0005 & 0.0807±0.0001 & 0.0849±0.0001 & 0.0880±0.0002 & 0.0808±0.0001 \\
$\mathtt{m1\_quarterly}$  & N/A             & N/A           & 0.2196±0.0023 & 0.1948±0.0005 & 0.2328±0.0008 & 0.2327±0.0014 \\
$\mathtt{pems03}$         & 0.0784±0.0001   & 0.2028±0.0002 & 0.0568±0.0000 & 0.0574±0.0001 & 0.0569±0.0000 & 0.0506±0.0000 \\
$\mathtt{uber\_hourly}$   & N/A             & N/A           & 0.1035±0.0002 & 0.1013±0.0002 & 0.1093±0.0003 & 0.1234±0.0002 \\
\cmidrule(lr){1-7}
&   &    &  \multicolumn{1}{r}{avg. rel. impr.} & \multicolumn{1}{l}{2.75\%}  & \multicolumn{1}{r}{avg. rel. impr.} & \multicolumn{1}{l}{3.88\%}  \\
\bottomrule
\end{tabular}
\end{center}
\end{table*}

\begin{table*}[htbp]
\scriptsize
\caption{Comparison of $0.9$-risk accuracy. ``w/o'' denotes methods without time-dependent errors, while ``w/'' indicates our method. Boldface values indicate that models considering time-dependent errors have better performance. Mean and standard deviation are obtained from 10 runs of each model.}
\label{tab:p09}
\begin{center}
\begin{tabular}{lcccccc}
\toprule
                          & VAR     & GARCH       & \multicolumn{2}{c}{GPVar}              & \multicolumn{2}{c}{Transformer} \\
\cmidrule(lr){2-7}
                          &       &      & w/o              & w/           & w/o              & w/   \\
\midrule
$\mathtt{exchange\_rate}$ & 0.0021±0.0000   & 0.0070±0.0000 & 0.0042±0.0001 & 0.0057±0.0001 & 0.0030±0.0000 & 0.0023±0.0001 \\
$\mathtt{solar}$          & 0.4676±0.0016   & 0.4393±0.0008 & 0.1617±0.0004 & 0.1597±0.0003 & 0.2744±0.0015 & 0.2710±0.0015 \\
$\mathtt{electricity}$    & 0.0414±0.0003   & 0.0744±0.0003 & 0.0211±0.0004 & 0.0185±0.0002 & 0.0281±0.0002 & 0.0366±0.0002 \\
$\mathtt{traffic}$        & 11.0170±0.0405  & 0.1689±0.0001 & 0.0666±0.0001 & 0.0580±0.0001 & 0.0609±0.0001 & 0.0698±0.0000 \\
$\mathtt{wiki}$           & 174.0756±0.3770 & 1.5906±0.0044 & 0.2136±0.0002 & 0.2048±0.0001 & 0.2117±0.0006 & 0.1764±0.0003 \\
$\mathtt{m4\_hourly}$     & 0.1029±0.0003   & 0.1309±0.0003 & 0.0452±0.0002 & 0.0463±0.0001 & 0.0525±0.0001 & 0.0475±0.0002 \\
$\mathtt{m1\_quarterly}$  & N/A             & N/A           & 0.3049±0.0044 & 0.2787±0.0027 & 0.3784±0.0031 & 0.3621±0.0037 \\
$\mathtt{pems03}$         & 0.0399±0.0000   & 0.1783±0.0001 & 0.0317±0.0000 & 0.0317±0.0001 & 0.0304±0.0000 & 0.0269±0.0000 \\
$\mathtt{uber\_hourly}$   & N/A             & N/A           & 0.0533±0.0002 & 0.0528±0.0001 & 0.0562±0.0001 & 0.0638±0.0002 \\
\cmidrule(lr){1-7}
&   &    &  \multicolumn{1}{r}{avg. rel. impr.} & \multicolumn{1}{l}{0.22\%}  & \multicolumn{1}{r}{avg. rel. impr.} & \multicolumn{1}{l}{0.91\%}  \\
\bottomrule
\end{tabular}
\end{center}
\end{table*}

\begin{table*}[htbp]
\tiny
\caption{Comparison of \(\operatorname{ES}\) accuracy. ``w/o'' denotes methods without time-dependent errors, while ``w/'' indicates our method. Boldface values indicate that models considering time-dependent errors have better performance. Mean and standard deviation are obtained from 10 runs of each model.}
\label{tab:es}
\begin{center}
\begin{tabular}{lcccccc}
\toprule
                          & VAR     & GARCH       & \multicolumn{2}{c}{GPVar}              & \multicolumn{2}{c}{Transformer} \\
\cmidrule(lr){2-7}
                          &       &      & w/o              & w/           & w/o              & w/   \\
\midrule
$\mathtt{exchange\_rate}$ & 0.1301±0.0002                 & 0.6085±0.0009            & 0.3674±0.0067          & 0.2613±0.0047          & 0.1798±0.0039          & 0.1438±0.0026          \\
$\mathtt{solar}$ (\(\times 10^3\))         & 1.7429±0.0043              & 1.7758±0.0015         & 1.6052±0.0095       & 1.6591±0.0050       & 1.5307±0.0049       & 1.4633±0.0044       \\
$\mathtt{electricity}$ (\(\times 10^5\))   & 1.0102±0.0052          & 1.9422±0.0127    & 0.3569±0.0050    & 0.3172±0.0028    & 0.4031±0.0040    & 0.4754±0.0025    \\
$\mathtt{traffic}$        & 3.3585±0.010 (\(\times 10^3\))            & 4.4198±0.0020            & 2.4008±0.0020          & 2.2408±0.0015          & 2.2240±0.0021          & 2.2566±0.0014          \\
$\mathtt{wiki}$ (\(\times 10^7\))       & 970.0242±2.5944 & 2.8857±0.0783 & 0.1149±0.0027 & 0.1155±0.0031 & 0.1236±0.004 & 0.1075±0.0046 \\
$\mathtt{m4\_hourly}$ (\(\times 10^3\))     & 4.5109±0.0084              & 5.1849±0.0089         & 2.2729±0.0062       & 2.3611±0.0060       & 2.5877±0.0098       & 2.3440±0.0081       \\
$\mathtt{m1\_quarterly}$ (\(\times 10^2\)) & N/A                               & N/A                          & 3.7565±0.0294        & 3.3676±0.0147        & 4.2149±0.0252        & 4.1596±0.0248        \\
$\mathtt{pems03}$  (\(\times 10^3\))       & 1.3951±0.0009              & 5.4642±0.0067         & 1.0535±0.0010       & 1.0736±0.0015       & 1.0673±0.0012       & 0.9394±0.0004        \\
$\mathtt{uber\_hourly}$  (\(\times 10^3\)) & N/A                               & N/A           & 0.9035±0.0041        & 0.8773±0.0027        & 0.9377±0.0035        & 1.0566±0.0033   \\
\cmidrule(lr){1-7}
&   &    &  \multicolumn{1}{r}{avg. rel. impr.} & \multicolumn{1}{l}{5.58\%}  & \multicolumn{1}{r}{avg. rel. impr.} & \multicolumn{1}{l}{3.21\%}  \\
\bottomrule
\end{tabular}
\end{center}
\end{table*}

\begin{table*}[htbp]
\tiny
\caption{Comparison of \(\operatorname{RRMSE}\) accuracy. ``w/o'' denotes methods without time-dependent errors, while ``w/'' indicates our method. Boldface values indicate that models considering time-dependent errors have better performance. Mean and standard deviation are obtained from 10 runs of each model.}
\label{tab:rrmse}
\begin{center}
\begin{tabular}{lcccccc}
\toprule
                          & VAR     & GARCH       & \multicolumn{2}{c}{GPVar}              & \multicolumn{2}{c}{Transformer} \\
\cmidrule(lr){2-7}
                          &       &      & w/o              & w/           & w/o              & w/   \\
\midrule
$\mathtt{exchange\_rate}$ & 0.0247±0.0000     & 0.0983±0.0002 & 0.0699±0.0012  & 0.0501±0.0010  & 0.0350±0.0008  & 0.0265±0.0007  \\
$\mathtt{solar}$          & 0.9365±0.0025     & 0.9556±0.0008 & 0.8195±0.0038  & 0.8334±0.0019  & 0.8114±0.0023  & 0.7761±0.0019  \\
$\mathtt{electricity}$    & 0.2732±0.0020     & 0.5584±0.0036 & 0.1010±0.0013  & 0.0912±0.0009  & 0.1130±0.0011  & 0.1293±0.0007  \\
$\mathtt{traffic}$        & 0.6312±0.0017 (\(\times 10^3\))   & 0.9894±0.0008 & 0.5383±0.0005  & 0.5061±0.0003  & 0.5025±0.0005  & 0.5052±0.0003  \\
$\mathtt{wiki}$           & 0.6519±0.0016 (\(\times 10^4\))   & 6.3386±0.2020 & 1.0288±0.0029  & 1.0393±0.0039  & 0.9292±0.0057  & 0.8752±0.0027  \\
$\mathtt{m4\_hourly}$     & 0.6163±0.0012     & 0.6848±0.0015 & 0.3072±0.0008  & 0.3168±0.0007  & 0.3420±0.0011  & 0.3179±0.0010  \\
$\mathtt{m1\_quarterly}$  & N/A               & N/A           & 19.1005±0.1246 & 17.0277±0.0845 & 20.2333±0.0830 & 20.2708±0.0924 \\
$\mathtt{pems03}$         & 0.3727±0.0003     & 0.8824±0.0013 & 0.2796±0.0003  & 0.2877±0.0005  & 0.2841±0.0003  & 0.2502±0.0001  \\
$\mathtt{uber\_hourly}$   & N/A               & N/A           & 0.2358±0.0012  & 0.2282±0.0008  & 0.2458±0.0010  & 0.2768±0.0009  \\
\cmidrule(lr){1-7}
&   &    &  \multicolumn{1}{r}{avg. rel. impr.} & \multicolumn{1}{l}{5.48\%}  & \multicolumn{1}{r}{avg. rel. impr.} & \multicolumn{1}{l}{2.85\%}  \\
\bottomrule
\end{tabular}
\end{center}
\end{table*}

\subsection{Training Dynamics}\label{apx:mar_training}

\begin{figure}[htbp]
  \centering\includegraphics[width=0.9\textwidth]{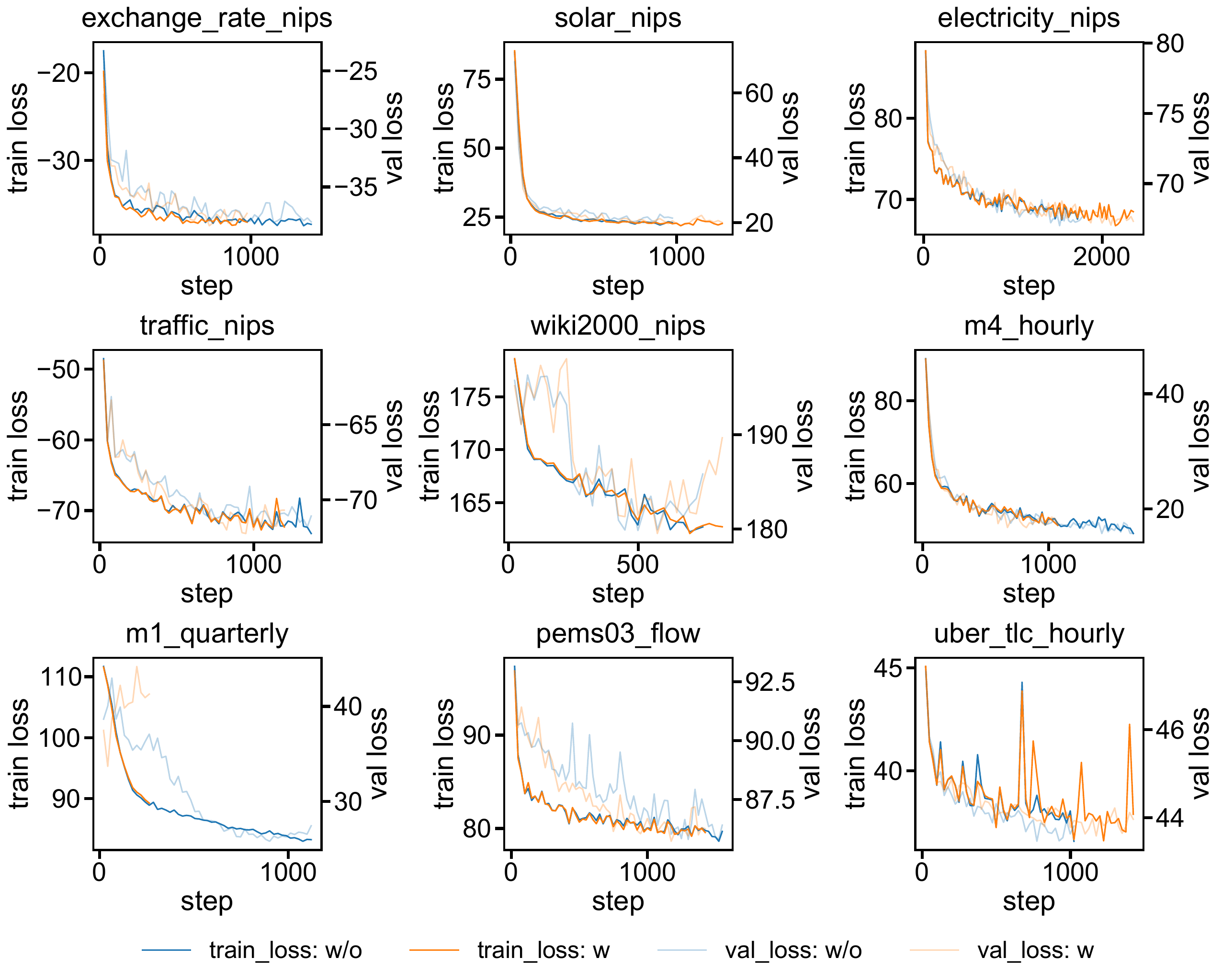}
  \caption{Training loss/validation loss vs training time of the GPVar model. ``w/o'' denotes methods without time-dependent errors, while ``w/'' indicates our method.}
\label{fig:apx_deepar_training}
\end{figure}

\begin{figure}[htbp]
  \centering\includegraphics[width=0.9\textwidth]{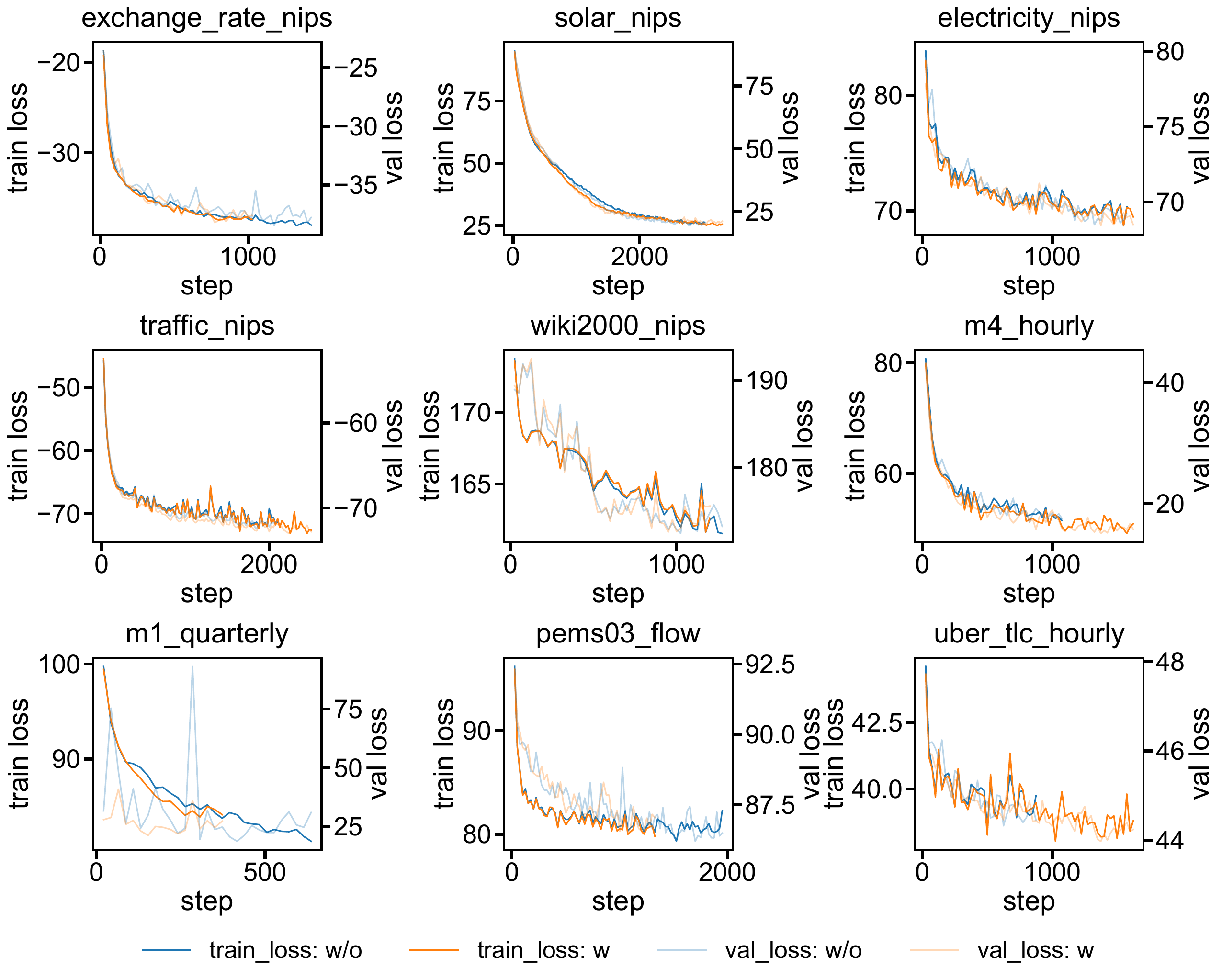}
  \caption{Training loss/validation loss vs training time of the Transformer model. ``w/o'' denotes methods without time-dependent errors, while ``w/'' indicates our method.}
\label{fig:apx_gpt_training}
\end{figure}

In Fig.~\ref{fig:apx_deepar_training} and Fig.~\ref{fig:apx_gpt_training}, we compare the training dynamics of the base models trained with and without our method. Note that the likelihood losses of the two methods are not directly comparable, even for the same dataset and base model, due to differences in the likelihood structures. We observe that while our method introduces more complexity into the likelihood function, there is no evidence that it significantly prolongs model convergence. On the contrary, our method can speed up convergence for some datasets in terms of the training steps used. We also report the training time in Table~\ref{tab:training_time}.

\begin{table}[htbp]
\scriptsize
\caption{Training cost comparison. ``w/o'' denotes methods without time-dependent errors, while ``w/'' indicates our method. }
\label{tab:training_time}
\begin{center}
\begin{tabular}{lcccccccc}
\toprule
                                 & \multicolumn{4}{c}{GPVar}              & \multicolumn{4}{c}{Transformer} \\
\cmidrule(lr){2-9}
                                 & \multicolumn{2}{c}{w/o}              & \multicolumn{2}{c}{w/}            & \multicolumn{2}{c}{w/o}              & \multicolumn{2}{c}{w/}       \\
                                 & sec./epoch   & epochs & sec./epoch & epochs & sec./epoch    & epochs & sec./epoch & epochs \\
\midrule
$\mathtt{exchange\_rate}$ & 4.60  & 56 & 200.27 & 39 & 9.73  & 57  & 206.49 & 41  \\
$\mathtt{solar}$          & 6.18  & 39 & 74.16  & 51 & 15.37 & 121 & 181.26 & 132 \\
$\mathtt{electricity}$    & 7.44  & 71 & 119.06 & 94 & 19.38 & 63  & 103.15 & 65  \\
$\mathtt{traffic}$        & 10.50 & 55 & 225.20 & 48 & 28.30 & 84  & 247.58 & 100 \\
$\mathtt{wiki}$           & 12.45 & 30 & 164.81 & 33 & 28.16 & 51  & 351.06 & 48  \\
$\mathtt{m4\_hourly}$     & 7.42  & 67 & 189.14 & 43 & 17.81 & 43  & 355.27 & 65  \\
$\mathtt{m1\_quarterly}$  & 4.24  & 51 & 25.82  & 12 & 9.84  & 29  & 24.99  & 17  \\
$\mathtt{pems03}$         & 11.75 & 62 & 143.34 & 57 & 36.31 & 78  & 88.05  & 53  \\
$\mathtt{uber\_hourly}$   & 6.82  & 41 & 174.90 & 57 & 17.28 & 35  & 188.08 & 65  \\
\bottomrule
\end{tabular}
\end{center}
\end{table}

\subsection{Effect of the Number of Time Series during Prediction}\label{apx:num_b_pred}

\begin{figure}[htbp]
  \centering\includegraphics[width=0.85\textwidth]{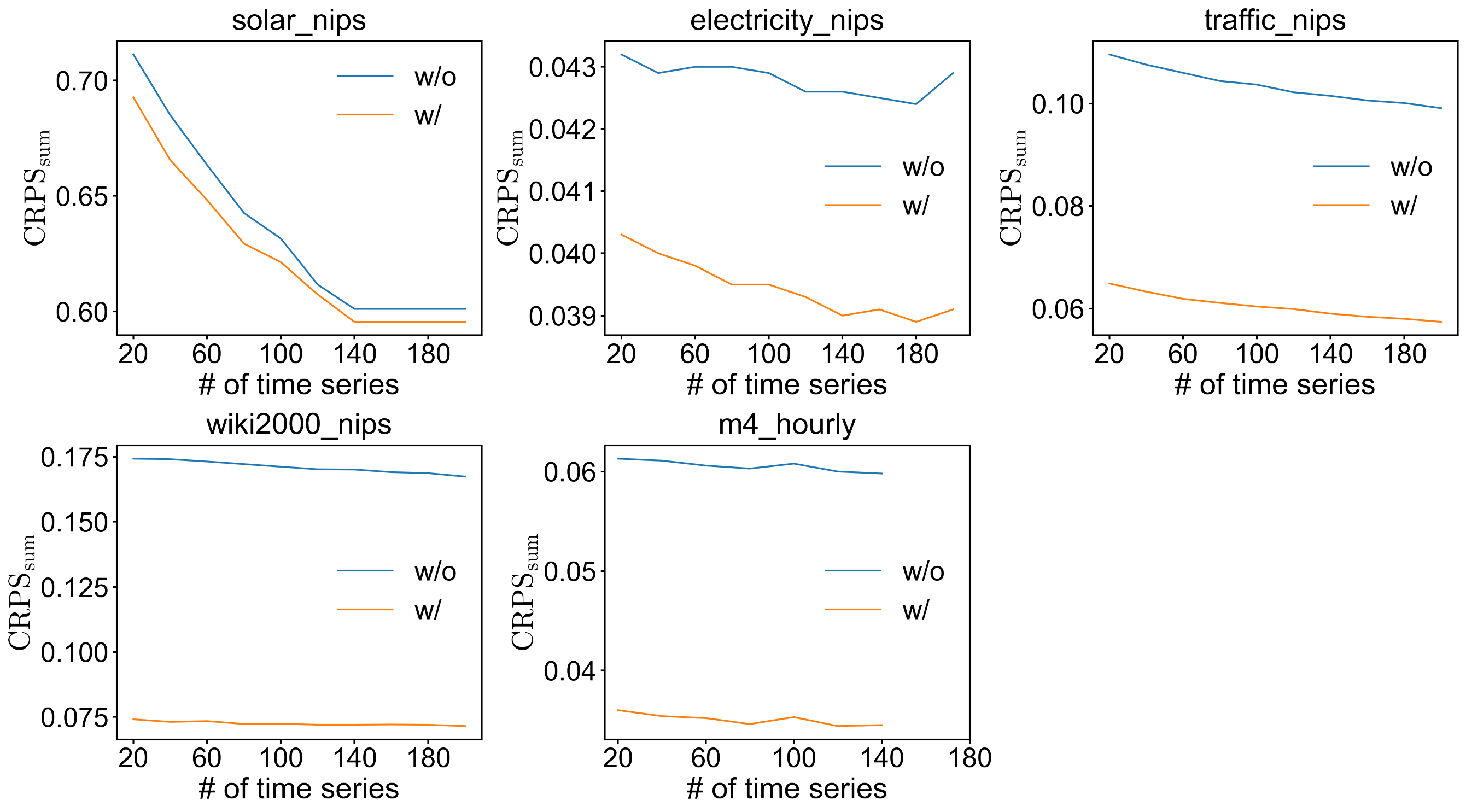}
  \caption{The influence of the number of time series in a batch on the performance of inference. "w/o" denotes methods without time-dependent errors, while "w/" indicates our method. We only show some datasets here because the remaining datasets have fewer than $B=20$ time series in the testing set.}
\label{fig:crps_batch_size}
\end{figure}

The number of time series does not impact training, as the model is trained using a random subset of \( B \) time series at a time, independent of the total number of time series \( N \). However, during prediction, the batch size can be increased beyond the training batch size of \( B=20 \) for multistep-ahead rolling predictions. This allows more information to be utilized, potentially improving both predictions and error calibration, provided that memory capacity permits. We conducted an additional experiment to demonstrate the effect of increasing the batch size during inference (Fig.~\ref{fig:crps_batch_size}).

\subsection{Additional Model Interpretation}\label{apx:extra_exp}
In this section, we provide further insights into how our method improves the base model. We illustrate these improvements by comparing the cross-correlations of the residuals from models with and without our method. Additionally, we demonstrate the performance of our method over the prediction horizon in multistep-ahead forecasting.

\subsubsection{Comparison of Residual Correlation}\label{apx:extra_exp_crc}
Recall that our method models both the autocovariance of errors $\operatorname{Cov}(\eta_{i,t-\Delta},\eta_{i,t})$ and the cross-lag covariance $\operatorname{Cov}(\eta_{i,t-\Delta},\eta_{j,t})$ between all pairs of components in the multivariate series. With the calibration process introduced in \S \ref{sec:prediction}, our method is expected to reduce error cross-correlations, including autocorrelation and cross-lag correlation. Here, we compare the empirical ACF of the residuals \(\eta_{i,t}\) of a single time series \(i\), as well as the empirical cross-correlations of $\boldsymbol{\eta}_{t}$ across multiple time series.

\begin{figure}[htbp]
  \centering\includegraphics[width=0.75\textwidth]{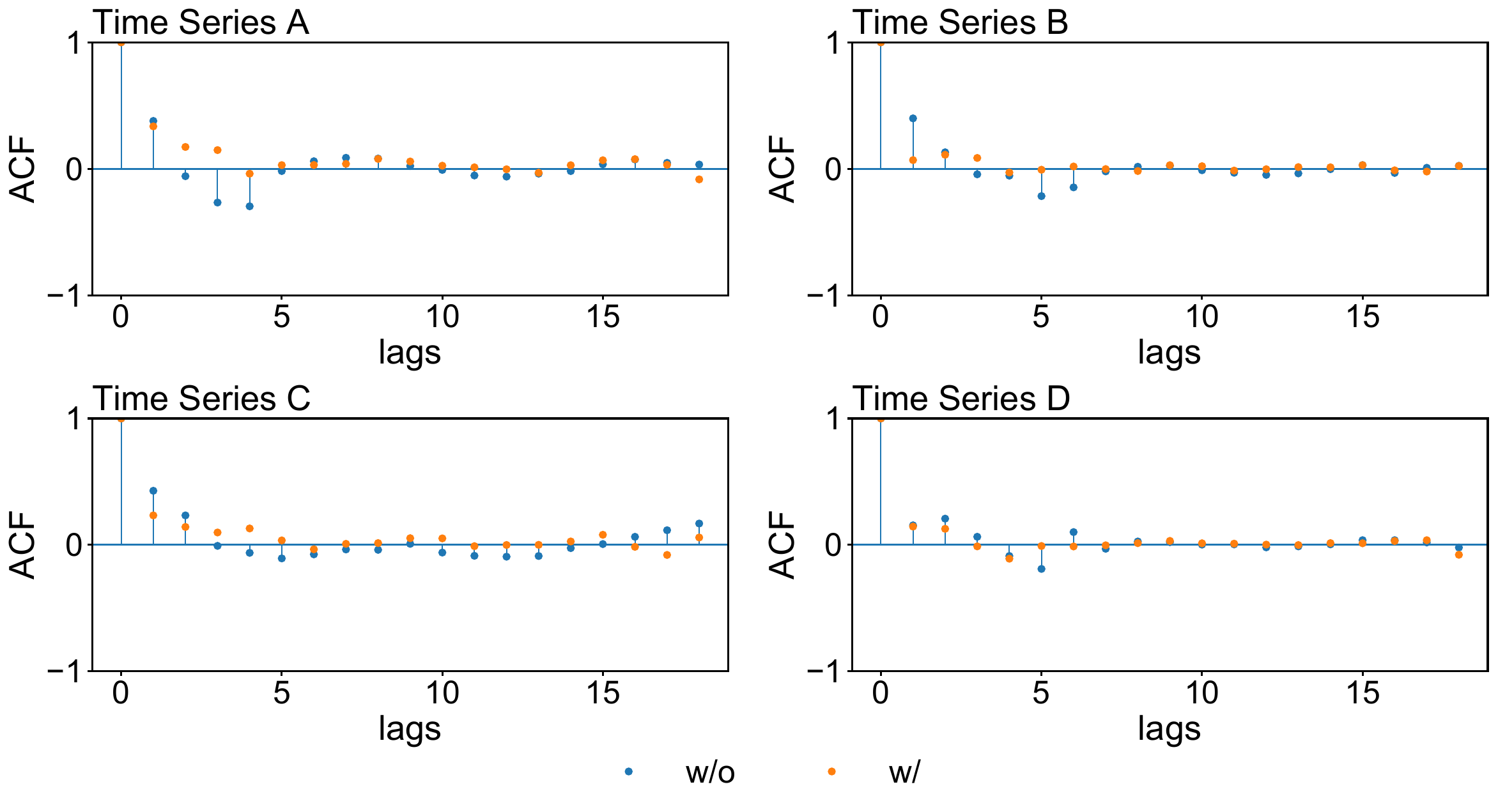}
  \caption{ACF comparison of the one-step-ahead prediction residuals with and without our method. The results depict the prediction outcomes generated by GPVar for four time series in the $\mathtt{solar}$ dataset. }
\label{fig:acf_compare_solar}
\end{figure}

\begin{figure}[htbp]
  \centering\includegraphics[width=0.75\textwidth]{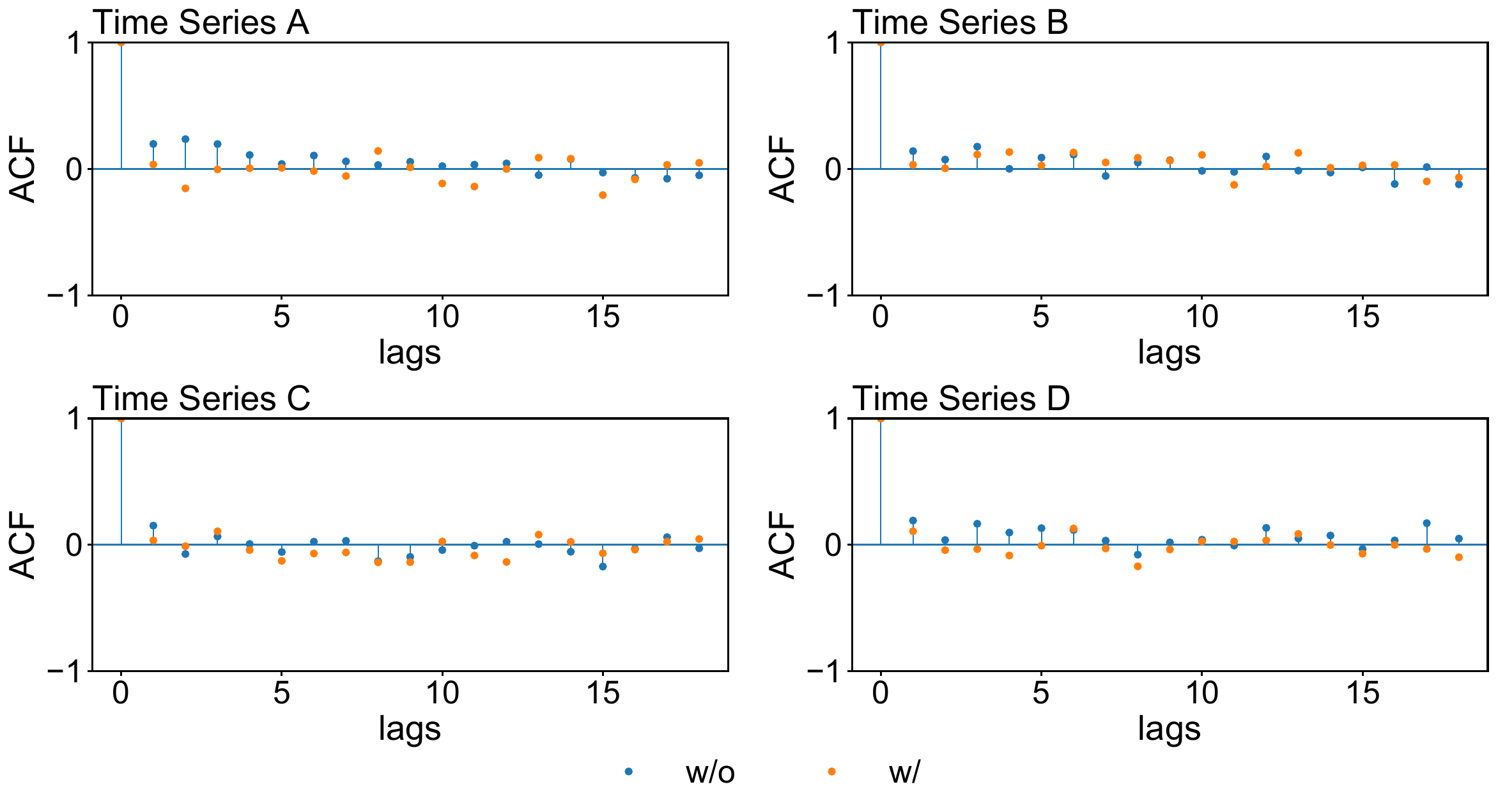}
  \caption{ACF comparison of the one-step-ahead prediction residuals with and without our method. The results depict the prediction outcomes generated by GPVar for four time series in the $\mathtt{electricity}$ dataset. }
\label{fig:acf_compare_electricity_nips}
\end{figure}

\begin{figure}[htbp]
  \centering\includegraphics[width=0.75\textwidth]{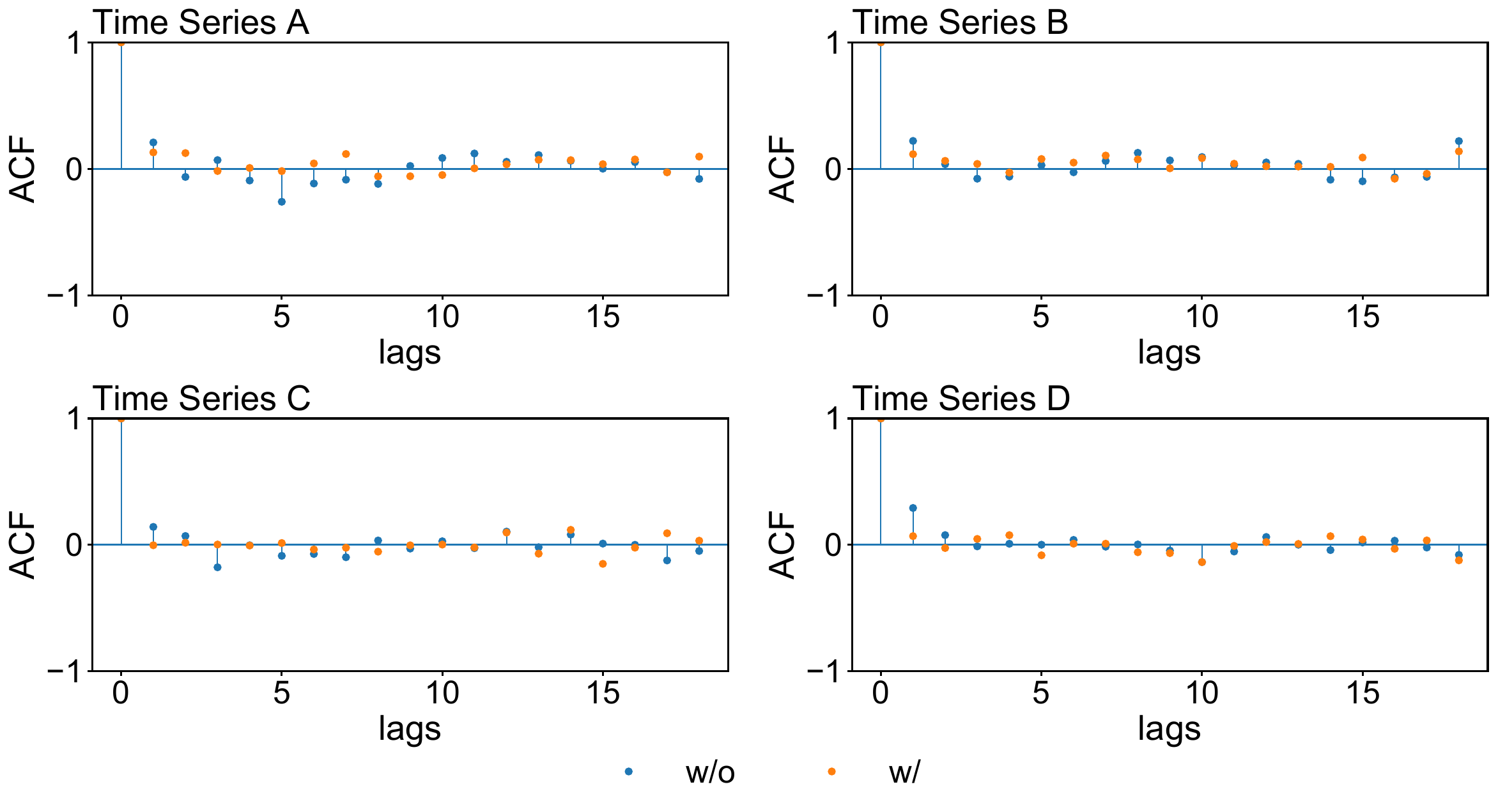}
  \caption{ACF comparison of the one-step-ahead prediction residuals with and without our method. The results depict the prediction outcomes generated by GPVar for four time series in the $\mathtt{traffic}$ dataset. }
\label{fig:acf_compare_traffic_nips}
\end{figure}

\begin{figure}[htbp]
  \centering\includegraphics[width=0.75\textwidth]{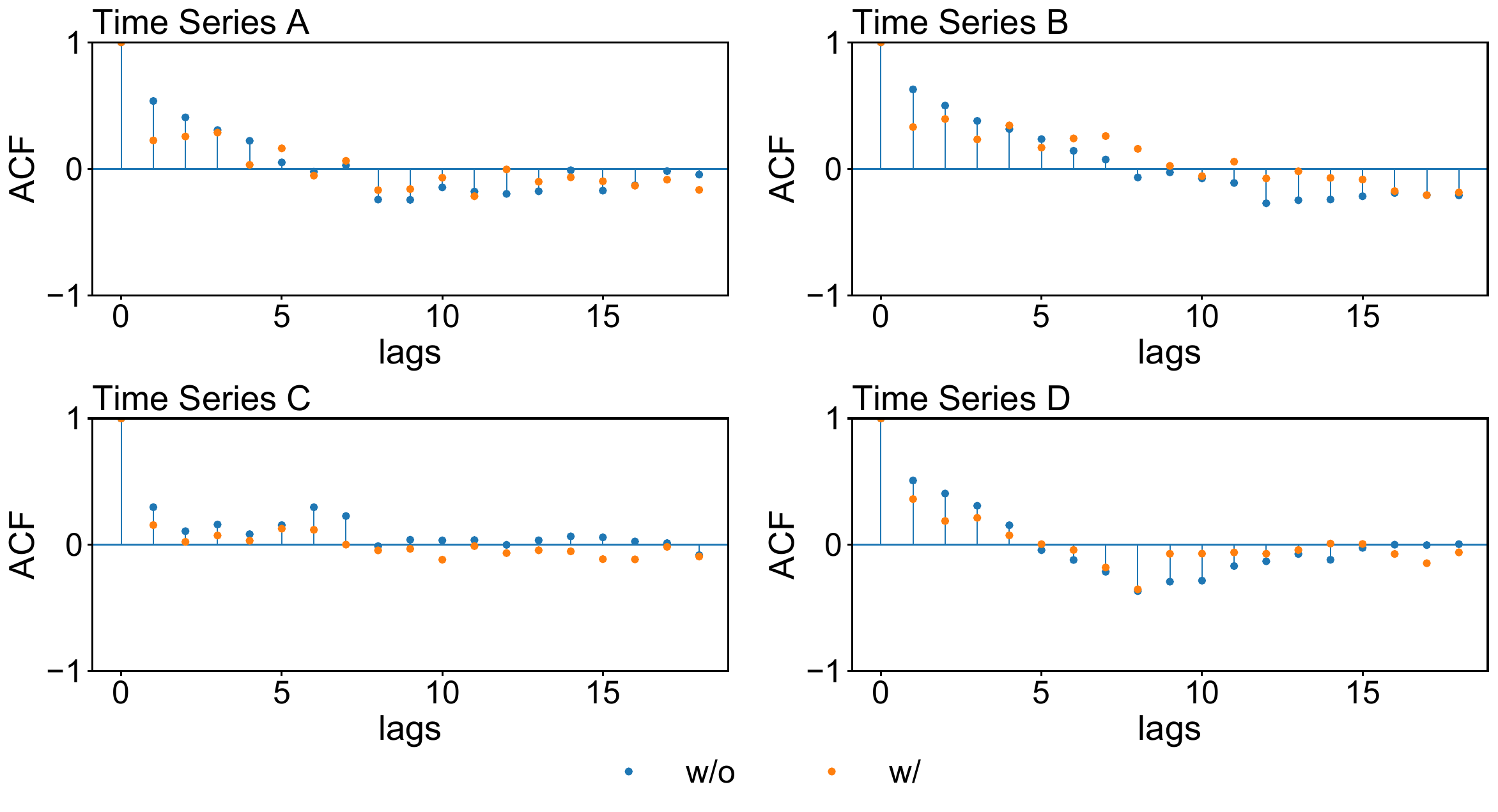}
  \caption{ACF comparison of the one-step-ahead prediction residuals with and without our method. The results depict the prediction outcomes generated by GPVar for four time series in the $\mathtt{wiki}$ dataset. }
\label{fig:acf_compare_wiki2000_nips}
\end{figure}

\begin{figure}[htbp]
  \centering\includegraphics[width=0.75\textwidth]{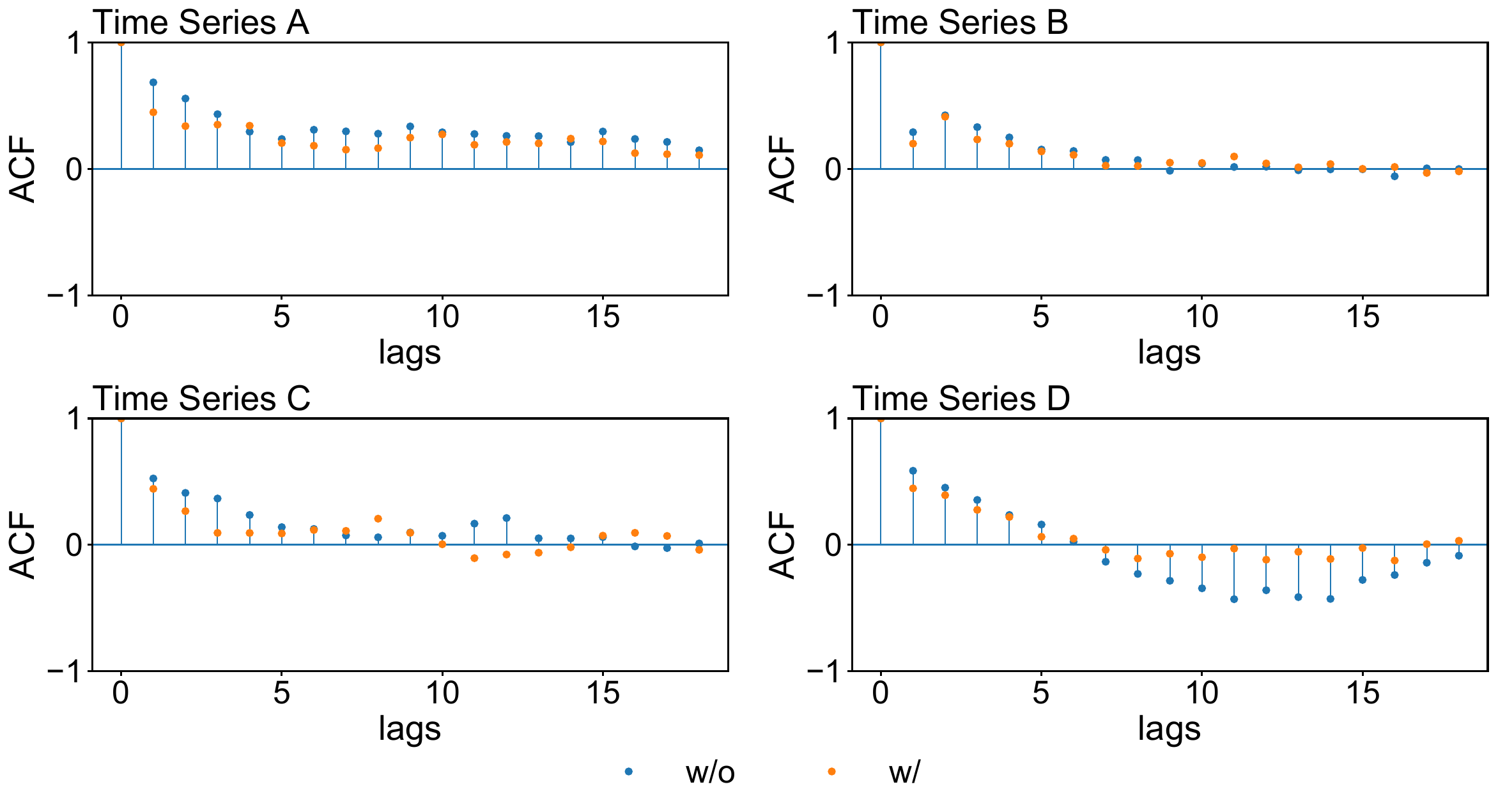}
  \caption{ACF comparison of the one-step-ahead prediction residuals with and without our method. The results depict the prediction outcomes generated by GPVar for four time series in the $\mathtt{m4\_hourly}$ dataset. }
\label{fig:acf_compare_m4_hourly}
\end{figure}

\begin{figure}[htbp]
  \centering\includegraphics[width=0.75\textwidth]{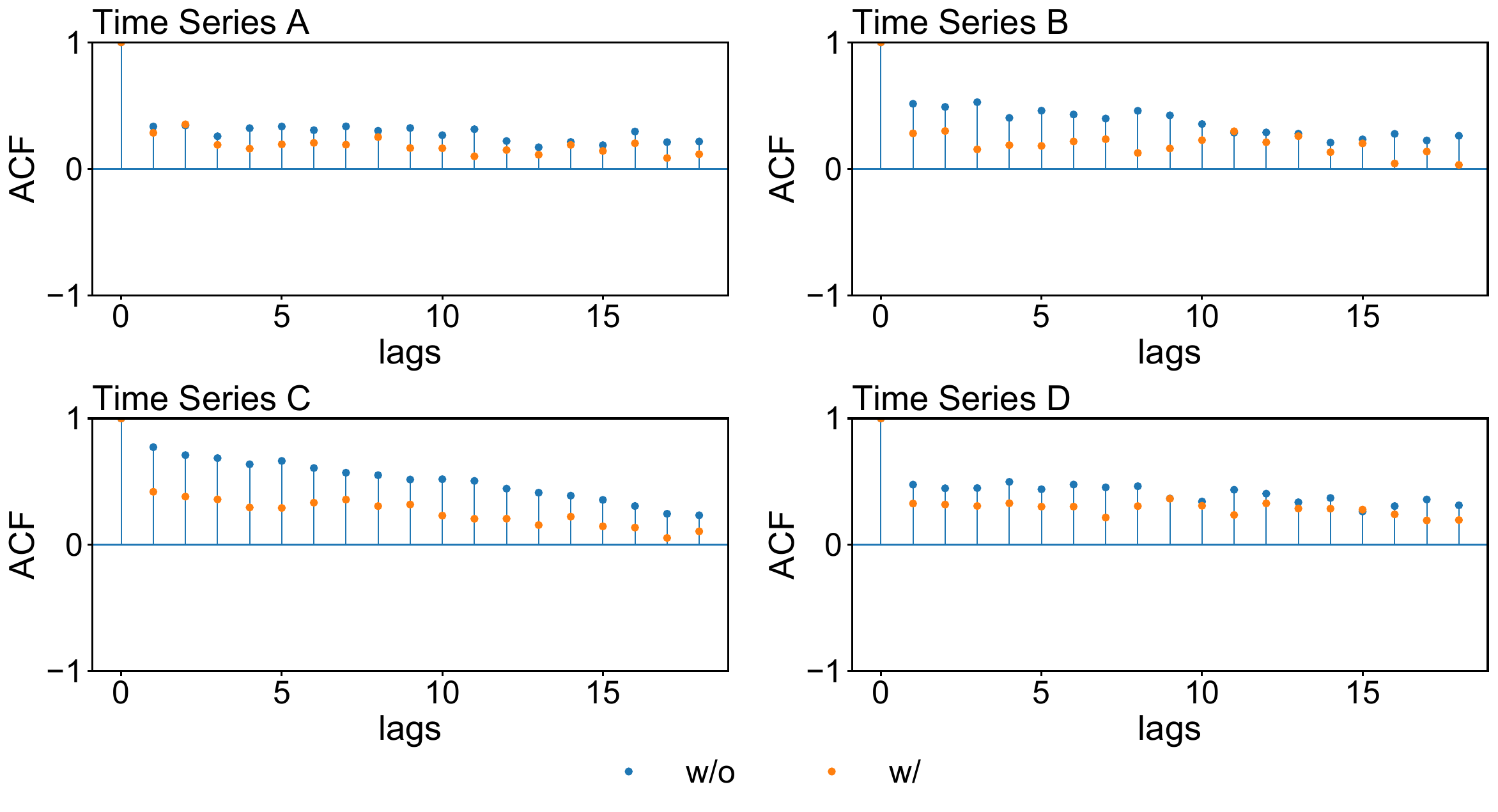}
  \caption{ACF comparison of the one-step-ahead prediction residuals with and without our method. The results depict the prediction outcomes generated by GPVar for four time series in the $\mathtt{pems03}$ dataset. }
\label{fig:acf_compare_pems03_flow}
\end{figure}

\begin{figure}[htbp]
  \centering\includegraphics[width=0.75\textwidth]{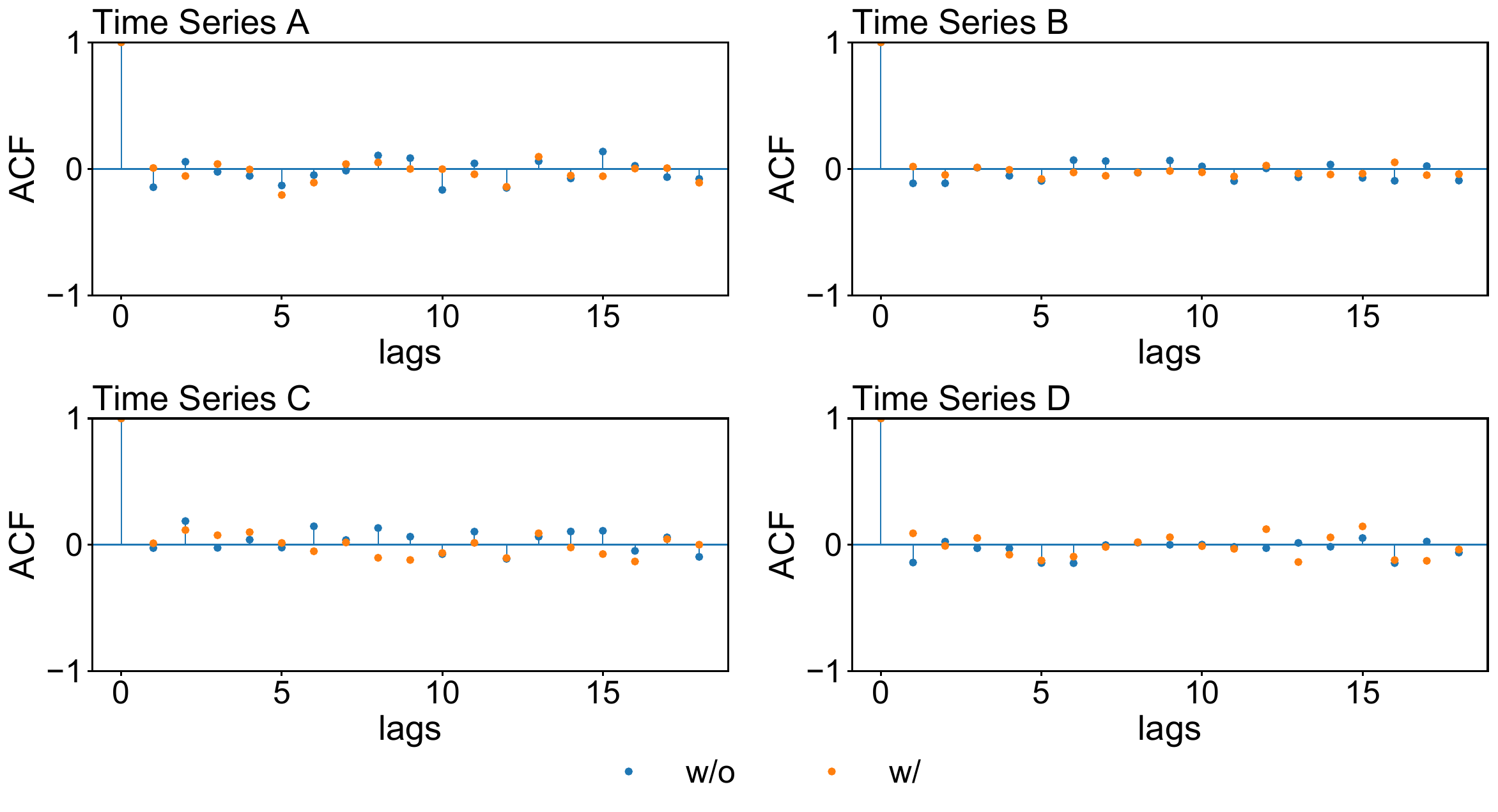}
  \caption{ACF comparison of the one-step-ahead prediction residuals with and without our method. The results depict the prediction outcomes generated by GPVar for four time series in the $\mathtt{uber\_hourly}$ dataset. }
\label{fig:acf_compare_uber_tlc_hourly}
\end{figure}

We begin by comparing the ACF of the one-step-ahead prediction residuals with and without our method. The comparisons are provided for the following datasets: $\mathtt{solar}$ (Fig.~\ref{fig:acf_compare_solar}), $\mathtt{electricity}$ (Fig.~\ref{fig:acf_compare_electricity_nips}), $\mathtt{traffic}$ (Fig.~\ref{fig:acf_compare_traffic_nips}), $\mathtt{wiki}$ (Fig.~\ref{fig:acf_compare_wiki2000_nips}), $\mathtt{m4\_hourly}$ (Fig.~\ref{fig:acf_compare_m4_hourly}), $\mathtt{pems03}$ (Fig.~\ref{fig:acf_compare_pems03_flow}), and $\mathtt{uber\_hourly}$ (Fig.~\ref{fig:acf_compare_uber_tlc_hourly}). We observe that the autocorrelation of the residuals is reduced after applying our method.

\begin{figure}[htbp]
  \centering\includegraphics[width=0.9\textwidth]{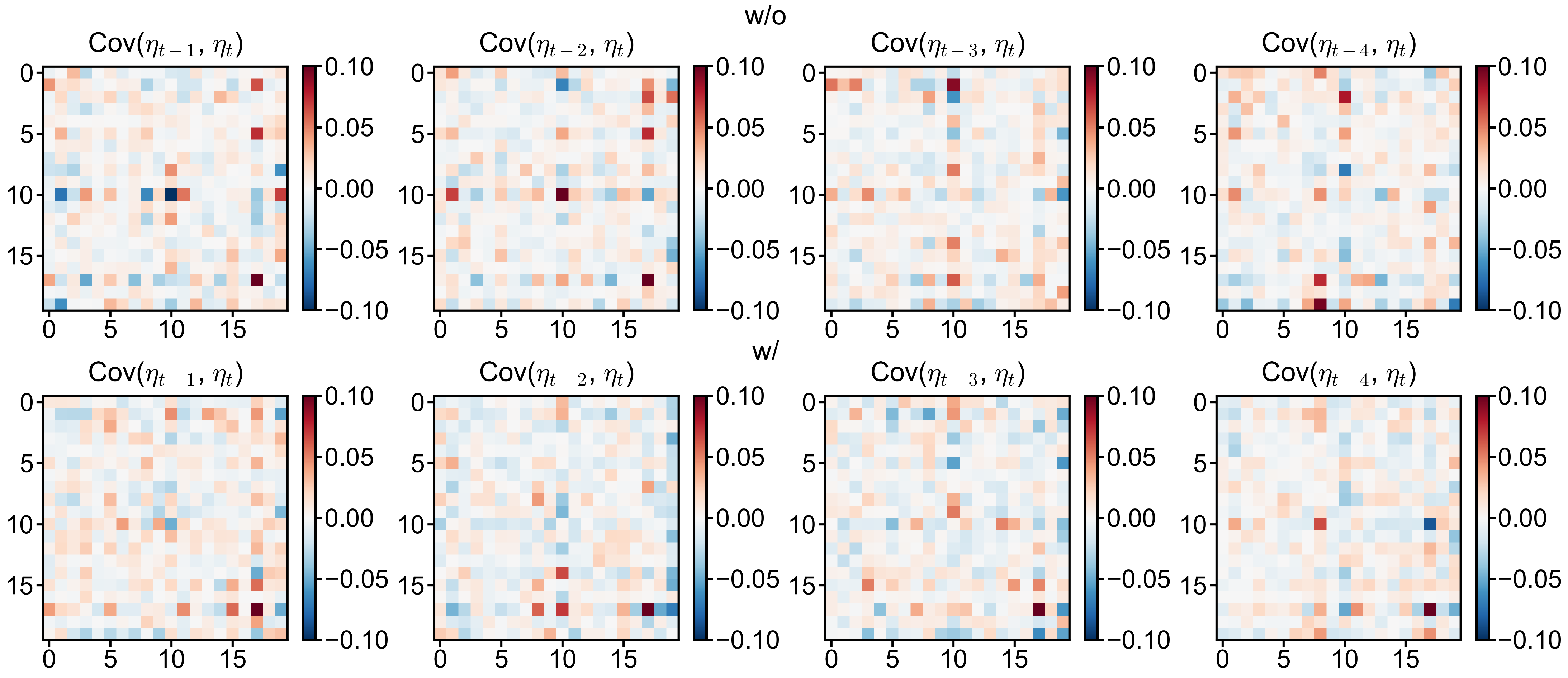}
  \caption{Cross-correlation comparison of the one-step-ahead prediction residuals with and without our method. The results depict the prediction outcomes generated by GPVar for four time series in the $\mathtt{electricity}$ dataset. }
\label{fig:ccf_compare_electricity_nips}
\end{figure}

\begin{figure}[htbp]
  \centering\includegraphics[width=0.9\textwidth]{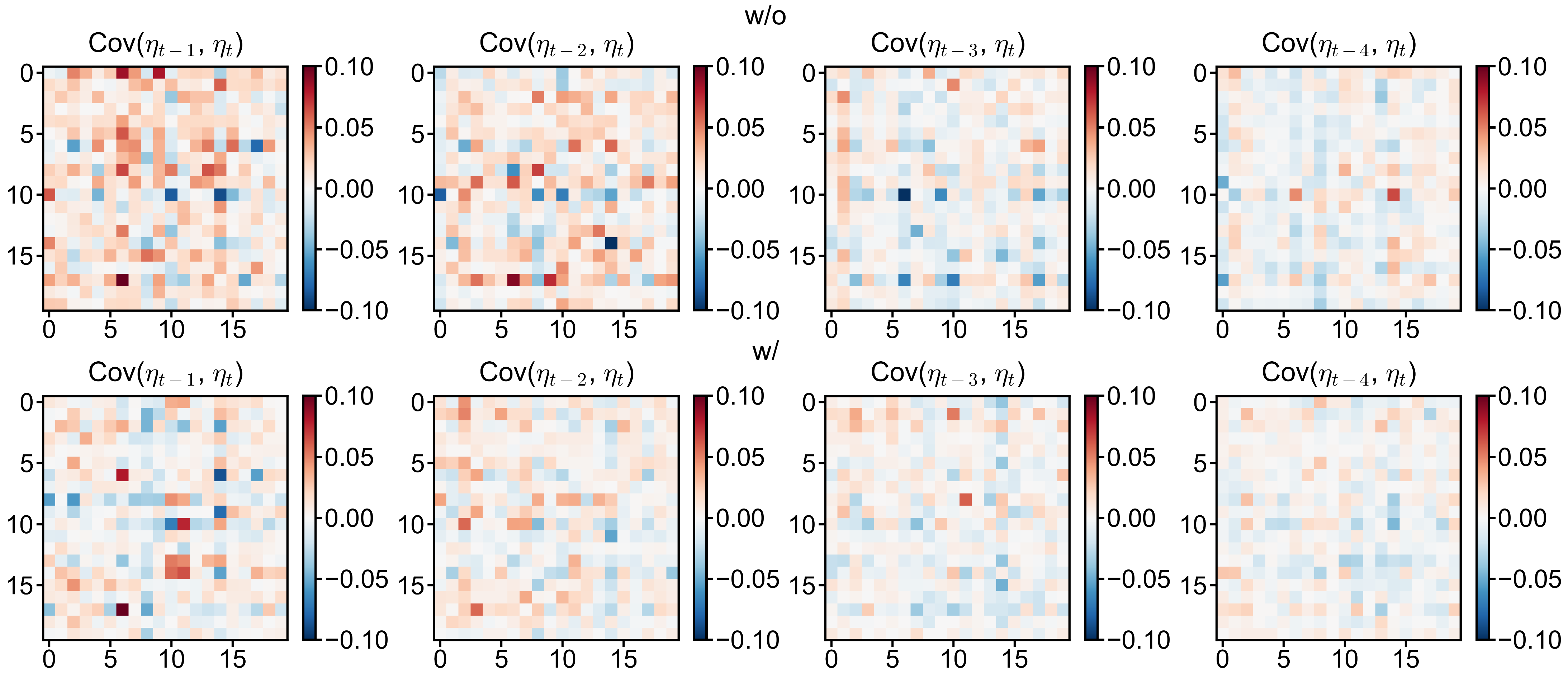}
  \caption{Cross-correlation comparison of the one-step-ahead prediction residuals with and without our method. The results depict the prediction outcomes generated by GPVar for four time series in the $\mathtt{traffic}$ dataset. }
\label{fig:ccf_compare_traffic_nips}
\end{figure}

\begin{figure}[htbp]
  \centering\includegraphics[width=0.9\textwidth]{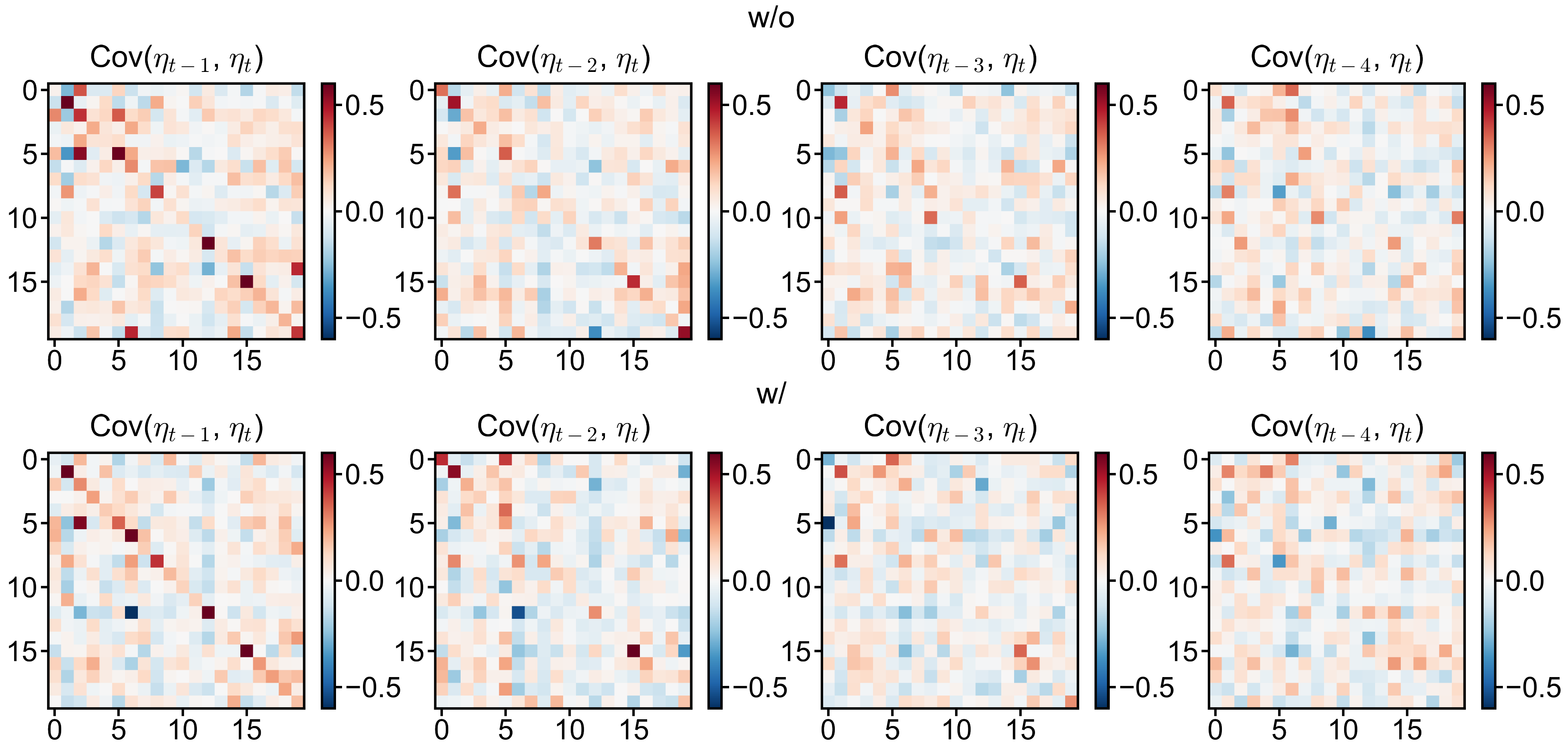}
  \caption{Cross-correlation comparison of the one-step-ahead prediction residuals with and without our method. The results depict the prediction outcomes generated by GPVar for four time series in the $\mathtt{wiki}$ dataset. }
\label{fig:ccf_compare_wiki2000_nips}
\end{figure}

\begin{figure}[htbp]
  \centering\includegraphics[width=0.9\textwidth]{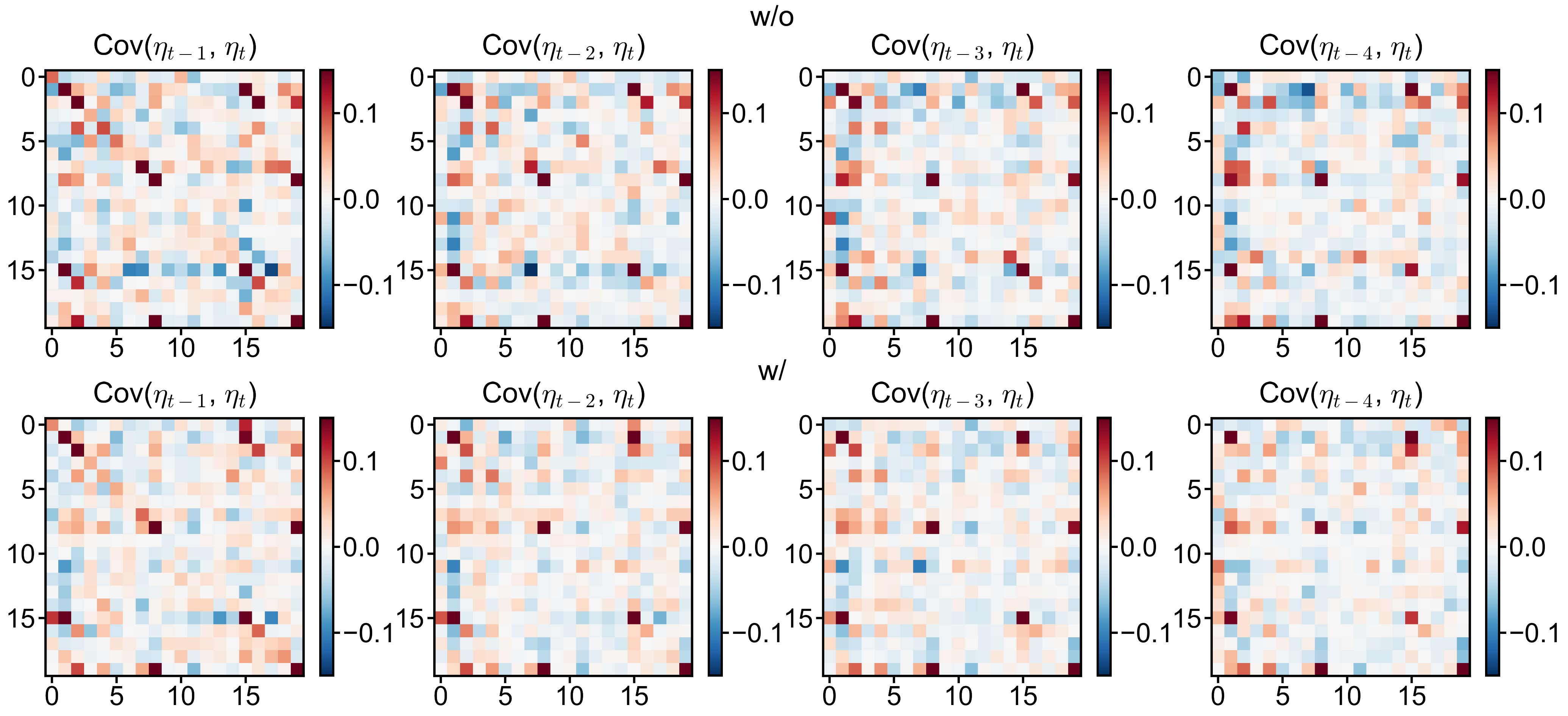}
  \caption{Cross-correlation comparison of the one-step-ahead prediction residuals with and without our method. The results depict the prediction outcomes generated by GPVar for four time series in the $\mathtt{m4\_hourly}$ dataset. }
\label{fig:ccf_compare_m4_hourly}
\end{figure}

\begin{figure}[htbp]
  \centering\includegraphics[width=0.9\textwidth]{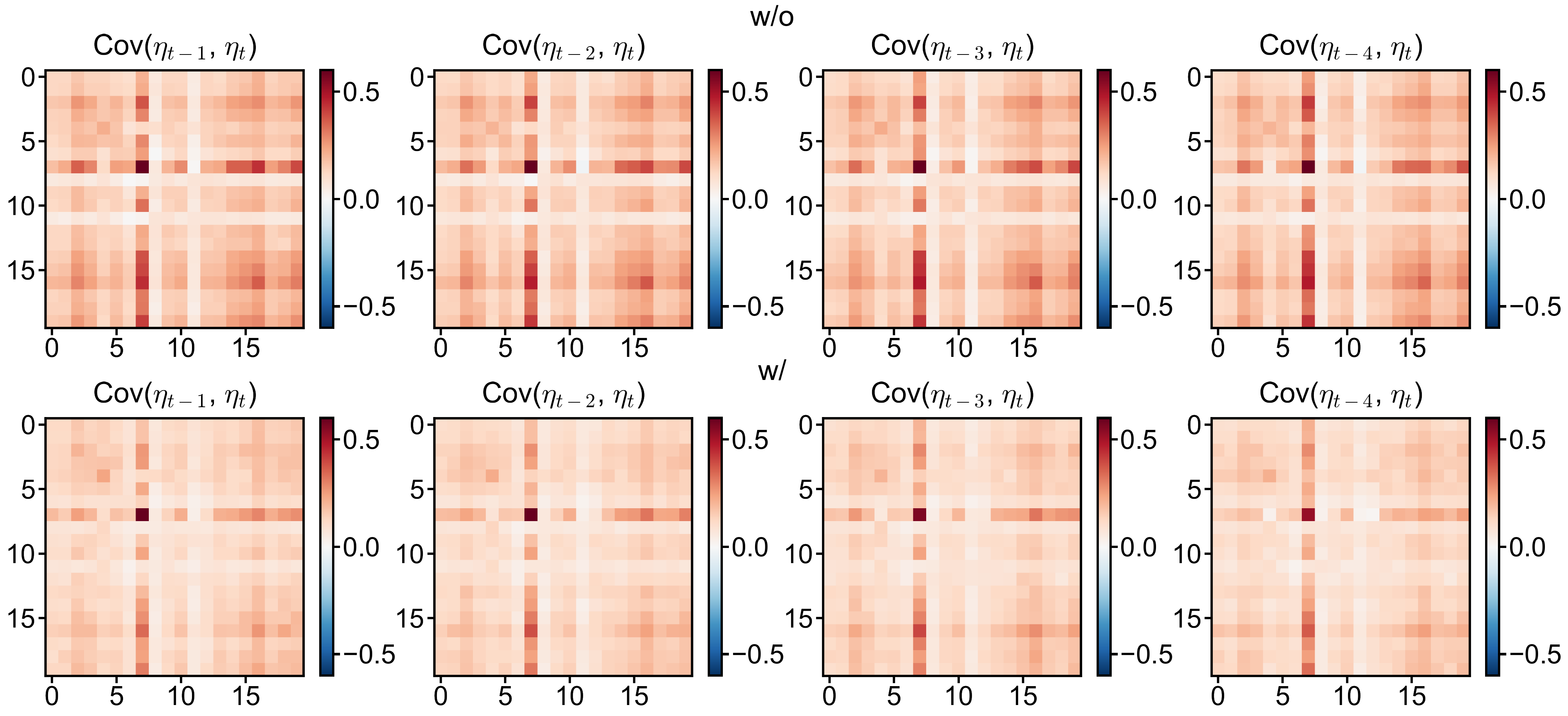}
  \caption{Cross-correlation comparison of the one-step-ahead prediction residuals with and without our method. The results depict the prediction outcomes generated by GPVar for four time series in the $\mathtt{pems03}$ dataset. }
\label{fig:ccf_compare_pems03_flow}
\end{figure}

\begin{figure}[htbp]
  \centering\includegraphics[width=0.9\textwidth]{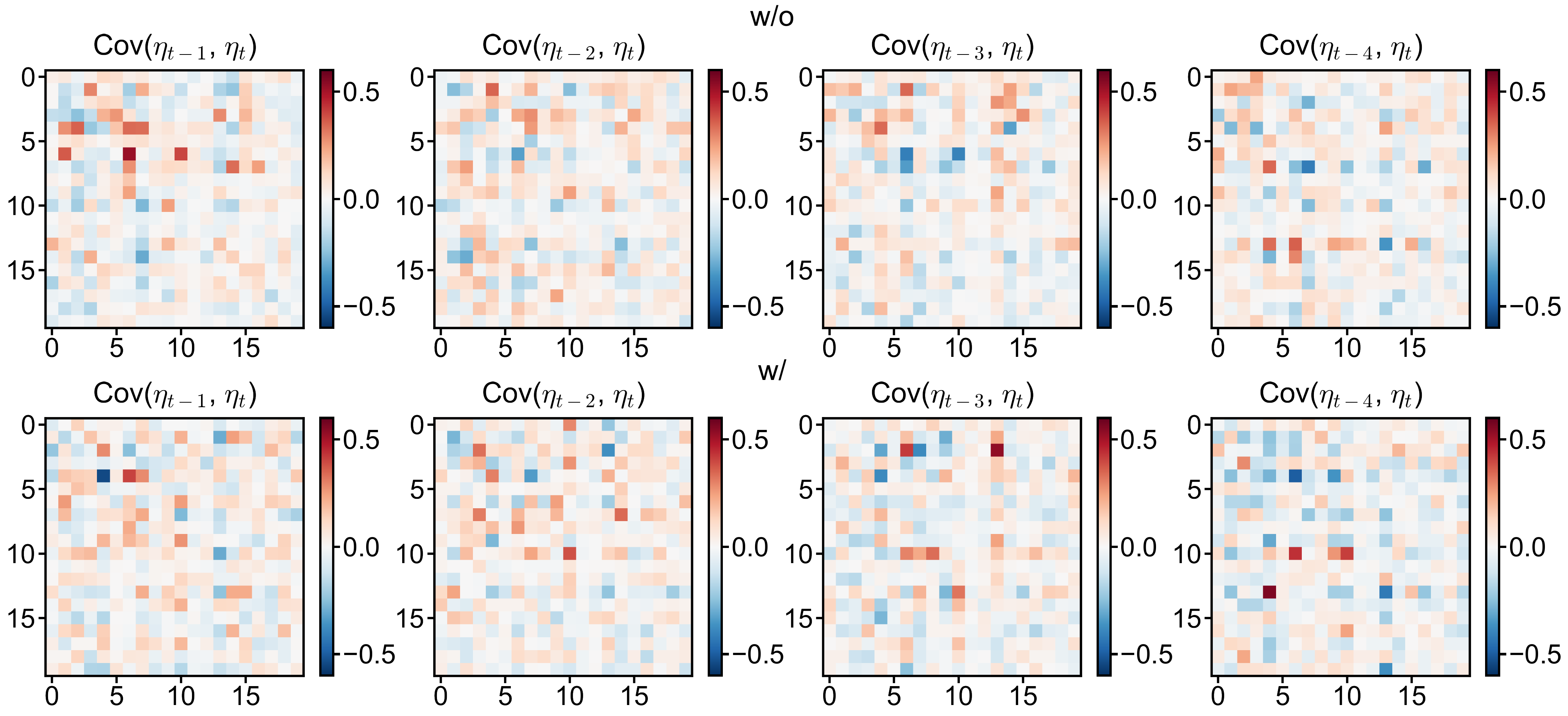}
  \caption{Cross-correlation comparison of the one-step-ahead prediction residuals with and without our method. The results depict the prediction outcomes generated by GPVar for four time series in the $\mathtt{uber\_hourly}$ dataset. }
\label{fig:ccf_compare_uber_tlc_hourly}
\end{figure}

Next, we compare the cross-correlations of the one-step-ahead prediction residuals with and without our method. The comparisons are provided for the following datasets: $\mathtt{electricity}$ (Fig.~\ref{fig:ccf_compare_electricity_nips}), $\mathtt{traffic}$ (Fig.~\ref{fig:ccf_compare_traffic_nips}), $\mathtt{wiki}$ (Fig.~\ref{fig:ccf_compare_wiki2000_nips}), $\mathtt{m4\_hourly}$ (Fig.~\ref{fig:ccf_compare_m4_hourly}), $\mathtt{pems03}$ (Fig.~\ref{fig:ccf_compare_pems03_flow}), and $\mathtt{uber\_hourly}$ (Fig.~\ref{fig:ccf_compare_uber_tlc_hourly}). We also observe that the cross-correlations of the residuals are reduced after applying our method.

\subsubsection{Performance Breakdown at Each Forecast Step}\label{apx:extra_exp_pbefs}

\begin{figure}[htbp]
  \centering\includegraphics[width=0.98\textwidth]{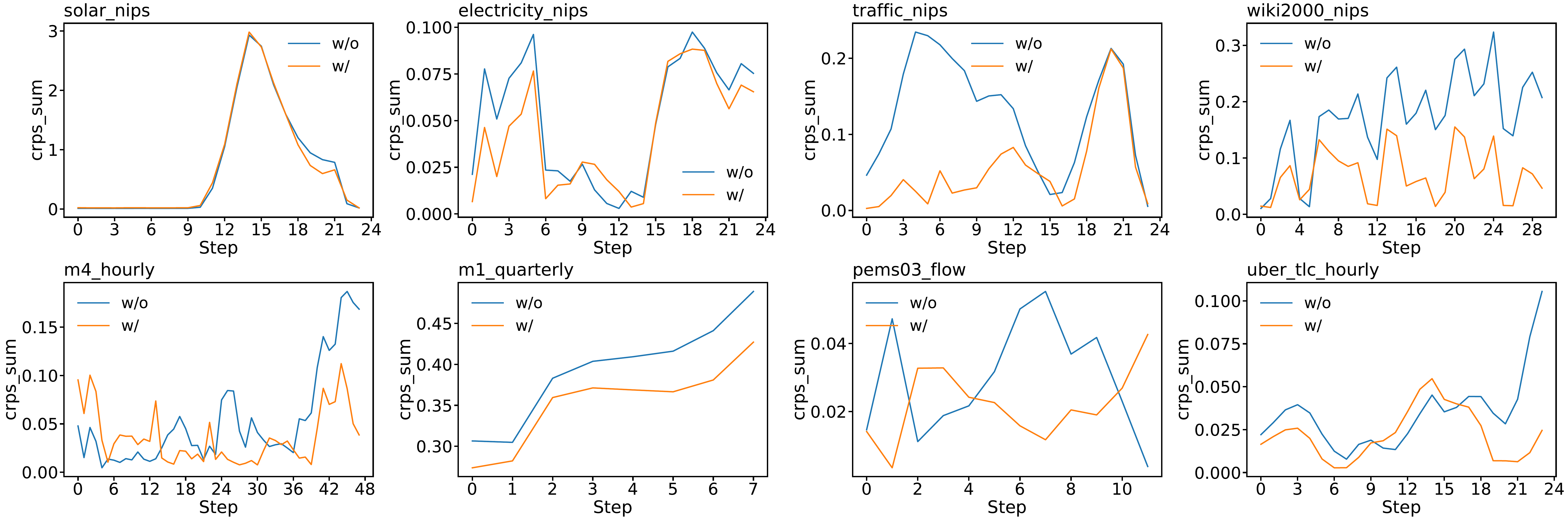}
  \caption{Step-wise $\operatorname{CRPS}_{\text{sum}}$ accuracy of GPVar. ``w/o'' denotes methods without time-dependent errors, while ``w/'' indicates our method. }
\label{fig:deepar_crps_sum_step}
\end{figure}

\begin{figure}[htbp]
  \centering\includegraphics[width=0.98\textwidth]{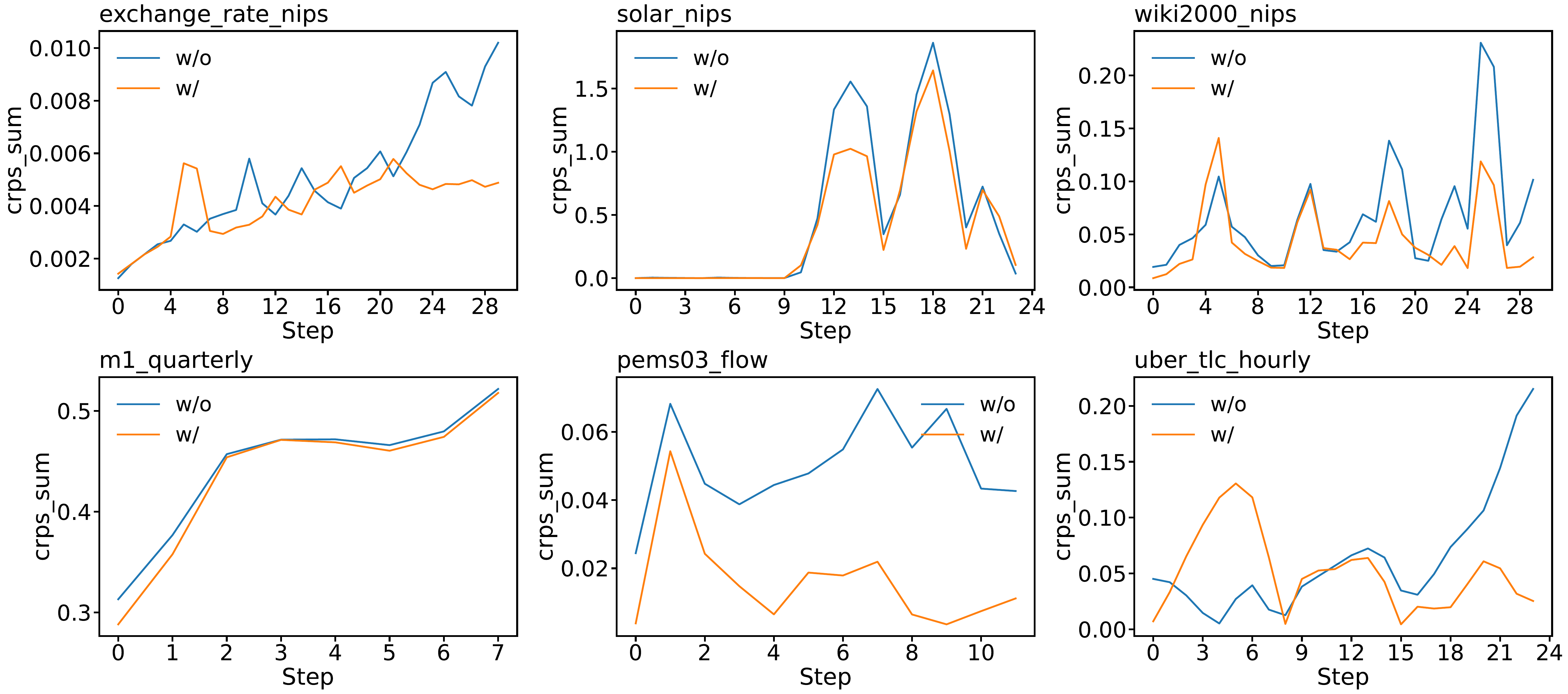}
  \caption{Step-wise $\operatorname{CRPS}_{\text{sum}}$ accuracy of Transformer. ``w/o'' denotes methods without time-dependent errors, while ``w/'' indicates our method. }
\label{fig:gpt_crps_sum_step}
\end{figure}

To investigate our performance gain at each forecast step, we calculate the \(\operatorname{CRPS}_{\text{sum}}\) for each forecast step. The results are shown in Fig.~\ref{fig:deepar_crps_sum_step} for GPVar and Fig.~\ref{fig:gpt_crps_sum_step} for the Transformer. Note that the \(\operatorname{CRPS}_{\text{sum}}\) reported in this section may have different scales compared to previous sections because they are not normalized. In multistep-ahead forecasting, since the predicted values are used as inputs for subsequent predictions within the prediction range, the residuals accumulate the effects of inaccuracies from previous steps. Therefore, the performance improvement depends not only on our modeling of error correlations but also on the properties of the residuals. These properties can be influenced by the absolute and relative time of the forecast and the seasonality of the data. For data without strong seasonality, residuals tend to be larger when predicting further ahead, making error accumulation more apparent. Conversely, for data with strong seasonality, the impact of error accumulation can vary. We observe that, in most scenarios, \(\operatorname{CRPS}_{\text{sum}}\) is reduced at the early forecasting stages. As predictions extend further into the future, some datasets (e.g., $\mathtt{traffic}$ in Fig.~\ref{fig:deepar_crps_sum_step}) show decreased improvement, likely due to seasonality effects. Conversely, other datasets (e.g., $\mathtt{wiki}$ in Fig.~\ref{fig:deepar_crps_sum_step}) exhibit larger improvements further into the future, possibly because the residuals accumulate over the steps.

\subsection{Alternative Parametrization of $\boldsymbol{C}_t$}\label{apx:ap}

\subsubsection{Learnable Lengthscales}\label{apx:ap_ll}
In this paper, the lengthscales are fixed when generating the correlation matrix \(\boldsymbol{C}_t\), and the flexibility of \(\boldsymbol{C}_t\) comes from dynamically generating the component weights of the kernel matrices. Making these lengthscales learnable parameters to find the optimal set of \(\{l_m\}_{m=1}^{M-1}\) is another approach we can explore to increase modeling flexibility. Based on the best model identified in Table~\ref{tab:crps_sum}, we experiment with treating the lengthscales as learnable parameters, jointly optimized with the base model. The results are shown in Table~\ref{tab:crps_sum_lr}. We do not observe significant improvement from learnable lengthscales.

\begin{table*}[htbp]
\scriptsize
\caption{Comparison of $\operatorname{CRPS}_{\text{sum}}$ accuracy. ``w/o'' denotes methods without time-dependent errors, while ``w/'' indicates our method. ``w/ (\(l\))'' indicates the lengthscales are learnable parameters. Boldface values indicate that models considering time-dependent errors have better performance. Mean and standard deviation are obtained from 10 runs of each model.}
\label{tab:crps_sum_lr}
\begin{center}
\begin{tabular}{lcccccc}
\toprule
                          & \multicolumn{3}{c}{GPVar}              & \multicolumn{3}{c}{Transformer} \\
\cmidrule(lr){2-7}
                          & w/o              & w/           & w/ (\(l\))           & w/o            & w/   & w/ (\(l\))  \\
\midrule
$\mathtt{exchange\_rate}$ & 0.0068±0.0004 & 0.0117±0.0004          & \textbf{0.0045±0.0001} & 0.0055±0.0002          & \textbf{0.0042±0.0002} & 0.0072±0.0002          \\
$\mathtt{solar}$          & 0.7103±0.0065 & \textbf{0.6929±0.0039} & 0.7727±0.0040          & 0.4960±0.0034          & \textbf{0.4132±0.0027} & 0.4138±0.0023          \\
$\mathtt{electricity}$    & 0.0430±0.0005 & 0.0403±0.0004          & \textbf{0.0351±0.0003} & \textbf{0.0494±0.0004} & 0.0638±0.0003          & 0.0858±0.0006          \\
$\mathtt{traffic}$        & 0.1095±0.0002 & \textbf{0.0649±0.0002} & 0.1297±0.0003          & \textbf{0.0717±0.0002} & 0.0981±0.0002          & 0.0950±0.0002          \\
$\mathtt{wiki}$           & 0.1745±0.0008 & \textbf{0.0743±0.0009} & 0.4839±0.0021 & 0.0841±0.0013          & 0.0500±0.0005          & \textbf{0.0472±0.0004} \\
$\mathtt{m4\_hourly}$     & 0.0613±0.0004 & \textbf{0.0358±0.0002} & 0.0527±0.0003          & 0.0651±0.0004          & 0.0616±0.0003          & \textbf{0.0355±0.0003} \\
$\mathtt{m1\_quarterly}$  & 0.3942±0.0030 & 0.3538±0.0017          & \textbf{0.3534±0.0017} & 0.4448±0.0027          & 0.4367±0.0028          & \textbf{0.3709±0.0120} \\
$\mathtt{pems03}$         & 0.0503±0.0001 & 0.0491±0.0002          & \textbf{0.0456±0.0001} & 0.0490±0.0001          & 0.0386±0.0001          & \textbf{0.0330±0.0001} \\
$\mathtt{uber\_hourly}$   & 0.0342±0.0006 & 0.0222±0.0004          & \textbf{0.0218±0.0003} & 0.0632±0.0003          & \textbf{0.0513±0.0005} & 0.0969±0.0005       \\
\bottomrule
\end{tabular}
\end{center}
\end{table*}

\subsubsection{Using Autocorrelations of an AR\((p)\) process}\label{apx:ap_ar}
One could parameterize $\boldsymbol{C}_t$ as fully learnable, positive definite symmetric Toeplitz matrices. For instance, an AR\((p)\) process has an autocorrelation matrix with a Toeplitz structure, allowing the modeling of negative correlations. This alternative approach may offer more flexibility in capturing complex correlation patterns in multivariate time series data. The autocorrelations of an AR\((p)\) process can be obtained by solving a set of equations known as the Yule-Walker equations \citep{guidolin2018essentials}. For example, if we consider an AR\((2)\) process and let \(\rho_k\) be the autocorrelation at lag \(k\):
\begin{equation}
    z_t=\phi_1 z_{t-1} + \phi_2 z_{t-2} + \epsilon_t.
\end{equation}
where \(\phi_1\) and \(\phi_2\) are the coefficients. We have \(\rho_0=1\) by definition and:
\begin{equation}
\begin{aligned}
    \rho_1 & = \phi_1 \rho_0 + \phi_2 \rho_1, \\
    & \ldots\\
    \rho_k & = \phi_1 \rho_{k-1} + \phi_2 \rho_{k-2}, k \geq 2.\\ 
\end{aligned}
\end{equation}

Since \(\rho_0=1\), we can solve for \(\rho_1\):
\begin{equation}
    \rho_1 = \frac{\phi_1}{1-\phi_2},
\end{equation}
and for any \(k \geq 2\), we can solve \(\rho_k\) iteratively by:
\begin{equation}
    \rho_k = \phi_1 \rho_{k-1} + \phi_2 \rho_{k-2}, k \geq 2.
\end{equation}

The collection \(\{\rho_0, \rho_1, \dots, \rho_k, \dots, \rho_{D-1}\}\) forms the first row or column of a Toeplitz matrix and can be used to parameterize \(\boldsymbol{C}_t\). We perform a hyperparameter search to find the best AR order \(p\) based on the validation loss. As shown in Table~\ref{tab:crps_sum_ar}, while the correlation matrix \(\boldsymbol{C}_t\) parameterized by an AR process shows promise in modeling both positive and negative correlations, it does not empirically provide an overall improvement compared to the kernel method used in this paper. This may be because cross-correlations in time series are predominantly positive. However, the AR method does show significant improvements on certain datasets where the kernel method does not perform well. For example, the AR method greatly improves GPVar on $\mathtt{exchange\_rate}$ and the Transformer on $\mathtt{electricity}$.

\begin{table*}[htbp]
\scriptsize
\caption{Comparison of $\operatorname{CRPS}_{\text{sum}}$ accuracy. ``w/o'' denotes methods without time-dependent errors, while ``w/'' indicates our method. ``w/ (AR)'' indicates \(\boldsymbol{C}_t\) is parameterized by an AR process. Boldface values indicate that models considering time-dependent errors have better performance. Mean and standard deviation are obtained from 10 runs of each model.}
\label{tab:crps_sum_ar}
\begin{center}
\begin{tabular}{lcccccc}
\toprule
                          & \multicolumn{3}{c}{GPVar}              & \multicolumn{3}{c}{Transformer} \\
\cmidrule(lr){2-7}
                          & w/o              & w/           & w/ (AR)           & w/o            & w/   & w/ (AR)  \\
\midrule
$\mathtt{exchange\_rate}$ & 0.0068±0.0004 & 0.0117±0.0004          & \textbf{0.0051±0.0002} & 0.0055±0.0002          & \textbf{0.0042±0.0002} & 0.0088±0.0004          \\
$\mathtt{solar}$          & 0.7103±0.0065 & 0.6929±0.0039          & \textbf{0.5923±0.0042} & 0.4960±0.0034          & 0.4132±0.0027          & \textbf{0.3362±0.0025} \\
$\mathtt{electricity}$    & 0.0430±0.0005 & \textbf{0.0403±0.0004} & 0.0433±0.0007          & 0.0494±0.0004          & 0.0638±0.0003          & \textbf{0.0252±0.0002} \\
$\mathtt{traffic}$        & 0.1095±0.0002 & \textbf{0.0649±0.0002} & 0.1095±0.0004          & \textbf{0.0717±0.0002} & 0.0981±0.0002          & 0.0878±0.0003          \\
$\mathtt{wiki}$           & 0.1745±0.0008 & \textbf{0.0743±0.0009} & 0.2375±0.0013          & 0.0841±0.0013          & \textbf{0.0500±0.0005} & 0.0512±0.0008          \\
$\mathtt{m4\_hourly}$     & 0.0613±0.0004 & 0.0358±0.0002          & \textbf{0.0298±0.0002} & 0.0651±0.0004          & \textbf{0.0616±0.0003} & 0.0680±0.0003          \\
$\mathtt{m1\_quarterly}$  & 0.3942±0.0030 & 0.3538±0.0017          & \textbf{0.1692±0.0029} & 0.4448±0.0027          & 0.4367±0.0028          & \textbf{0.4348±0.0028} \\
$\mathtt{pems03}$         & 0.0503±0.0001 & \textbf{0.0491±0.0002} & 0.0787±0.0002          & 0.0490±0.0001          & \textbf{0.0386±0.0001} & 0.0656±0.0001          \\
$\mathtt{uber\_hourly}$   & 0.0342±0.0006 & \textbf{0.0222±0.0004} & 0.0375±0.0004          & 0.0632±0.0003          & \textbf{0.0513±0.0005} & 0.0770±0.0007       \\
\bottomrule
\end{tabular}
\end{center}
\end{table*}

\subsection{Alternative Error Assumptions}\label{apx:ana}

A more suitable likelihood function can regularize the training process, potentially reducing residual correlations. For example, assuming the errors follow a multivariate \(t\)-distribution improves the robustness of the model to outliers. Additionally, a stronger base model can help produce residuals that are more independent. Based on these considerations, we designed our approach to adapt dynamically to varying levels of error correlation. The weighted correlation matrix assigns greater weight to the identity matrix when the errors exhibit lower correlation.

We also trained the baseline models using the likelihood of the multivariate \(t\)-distribution, and the results are shown in Table~\ref{tab:crps_sum_tdist}. While using an alternative distribution can lead to better performance on certain datasets when our method is not applied, we observed that our method effectively closes the performance gap in cases where the multivariate Gaussian assumption is outperformed by the \(t\)-distribution.

An important feature of our method is the ability to use a subset of time series in each training batch for model optimization, which enhances scalability. For the multivariate \(t\)-distribution, the distribution of these subsets of \(\mathbf{z}_t\) should have the same degrees of freedom as the full distribution of \(\mathbf{z}_t\). However, since the degrees of freedom are treated as an additional output of the model in each training batch, they are not guaranteed to be consistent across batches. While this is not problematic for deep learning, it violates the marginalization property of the \(t\)-distribution from a statistical standpoint.

We chose Gaussian noise for its beneficial properties, including its marginalization rule and well-defined conditional distribution, both essential for statistically consistent model training and reliable inference. To address model misspecification, a more effective approach could involve first transforming the original observations into Gaussian-distributed data using a Gaussian Copula \citep{salinas2019high}, and then applying our method.

\begin{table*}[!ht]
\scriptsize
\caption{$\operatorname{CRPS}_{\text{sum}}$ accuracy comparison. "w/o" denotes methods without time-dependent errors, while "w/" indicates our method. Bold values show models with time-dependent errors performing better. Mean and standard deviation are obtained from 10 runs of each model. "N/A" indicates that the model could not be properly fitted..}
\label{tab:crps_sum_tdist}
\begin{center}
\begin{tabular}{lcccccc}
\toprule
& \multicolumn{3}{c}{GPVar}  & \multicolumn{3}{c}{Transformer} \\
\cmidrule(lr){2-7}
                            & Gaussian (w/o)      & Gaussian (w/)           & $t$-distribution (w/o)  & Gaussian (w/o)         & Gaussian (w/)   & $t$-distribution (w/o) \\ 
\cmidrule(lr){1-7}
$\mathtt{exchange\_rate}$   & \textbf{0.0068±0.0004}   & 0.0117±0.0004      & 0.0159±0.0005     & 0.0055±0.0002      & \textbf{0.0042±0.0002}    & 0.0101±0.0003       \\
$\mathtt{solar}$            & 0.7103±0.0065       & \textbf{0.6929±0.0039}  & N/A               & 0.4960±0.0034      & \textbf{0.4132±0.0027}    & N/A                 \\
$\mathtt{electricity}$      & 0.0430±0.0005       & \textbf{0.0403±0.0004}  & 0.0467±0.0004     & 0.0494±0.0004      & 0.0638±0.0003        & \textbf{0.0466±0.0002}   \\
$\mathtt{traffic}$          & 0.1095±0.0002       & \textbf{0.0649±0.0002}  & 0.0679±0.0002     & \textbf{0.0717±0.0002}  & 0.0981±0.0002        & N/A                 \\
$\mathtt{wikipedia}$        & 0.1745±0.0008       & 0.0743±0.0009      & \textbf{0.0730±0.0004} & 0.0841±0.0013      & \textbf{0.0500±0.0005}    & 0.1979±0.0005       \\
$\mathtt{m4\_hourly}$       & 0.0613±0.0004       & \textbf{0.0358±0.0002}  & 0.0365±0.0003     & 0.0651±0.0004      & \textbf{0.0616±0.0003}    & 0.0665±0.0003       \\
$\mathtt{m1\_quarterly}$    & 0.3942±0.0030       & \textbf{0.3538±0.0017}  & 0.3550±0.0084     & 0.4448±0.0027      & \textbf{0.4367±0.0028}    & 0.4466±0.0044       \\
$\mathtt{pems03}$           & 0.0503±0.0001       & \textbf{0.0491±0.0002}  & 0.0679±0.0002     & 0.0490±0.0001      & \textbf{0.0386±0.0001}    & 0.0529±0.0002       \\
$\mathtt{uber\_hourly}$     & 0.0342±0.0006       & \textbf{0.0222±0.0004}  & 0.0666±0.0010     & 0.0632±0.0003      & 0.0513±0.0005        & \textbf{0.0340±0.0004}   \\
\bottomrule
\end{tabular}
\end{center}
\end{table*}

\subsection{Qualitative Results on Forecasting}\label{apx:mar_qualitative}
In this section, we provide qualitative analysis of the actual prediction performance by visualizing the predictions.

\begin{figure}[!ht]
  \centering\includegraphics[width=0.98\textwidth]{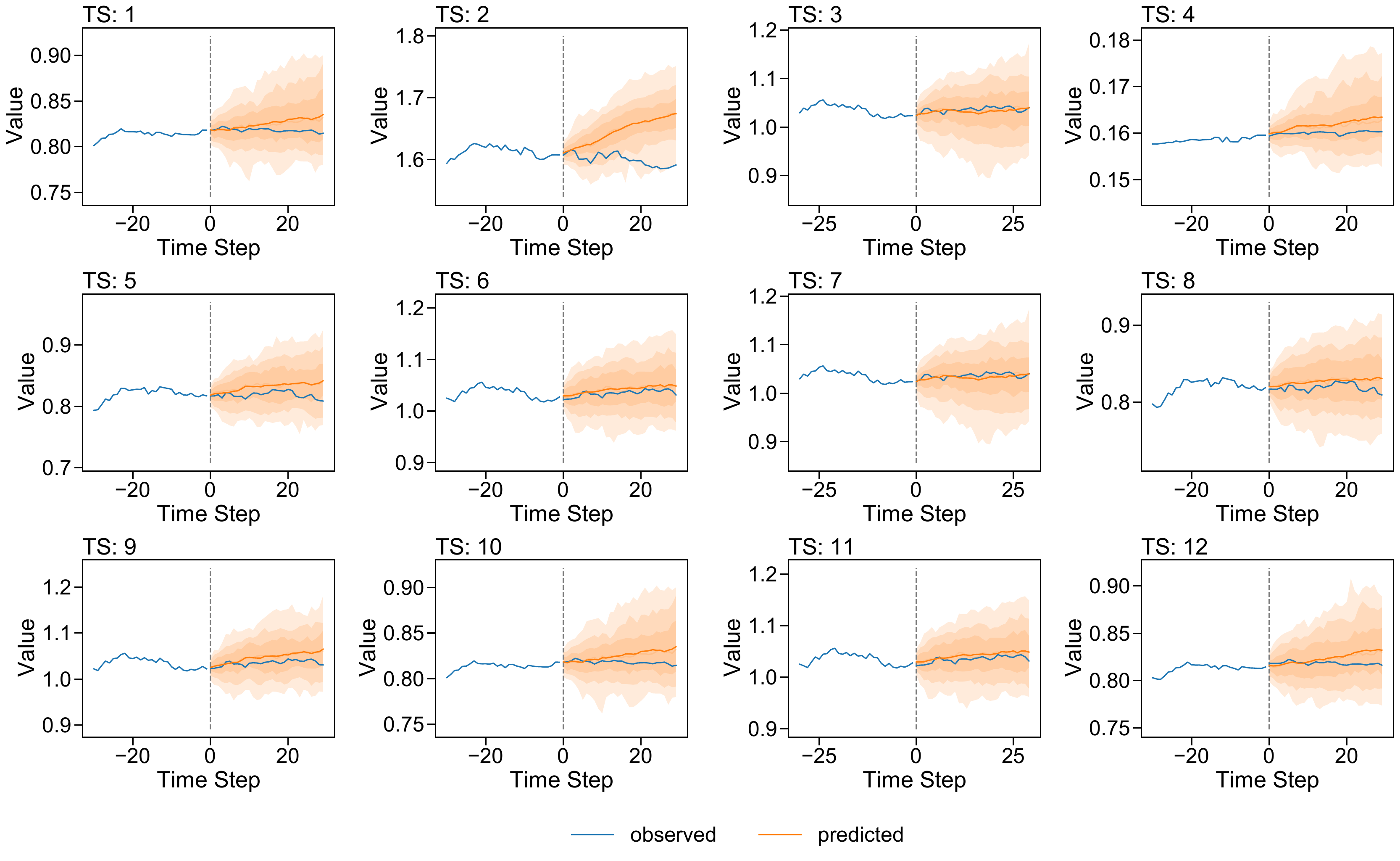}
  \caption{Visualization of forecasting results on \(\texttt{exchange\_rate}\) using GPVar with our method. }
\label{fig:deepar_exchange_rate_nips_vis}
\end{figure}

\begin{figure}[!ht]
  \centering\includegraphics[width=0.98\textwidth]{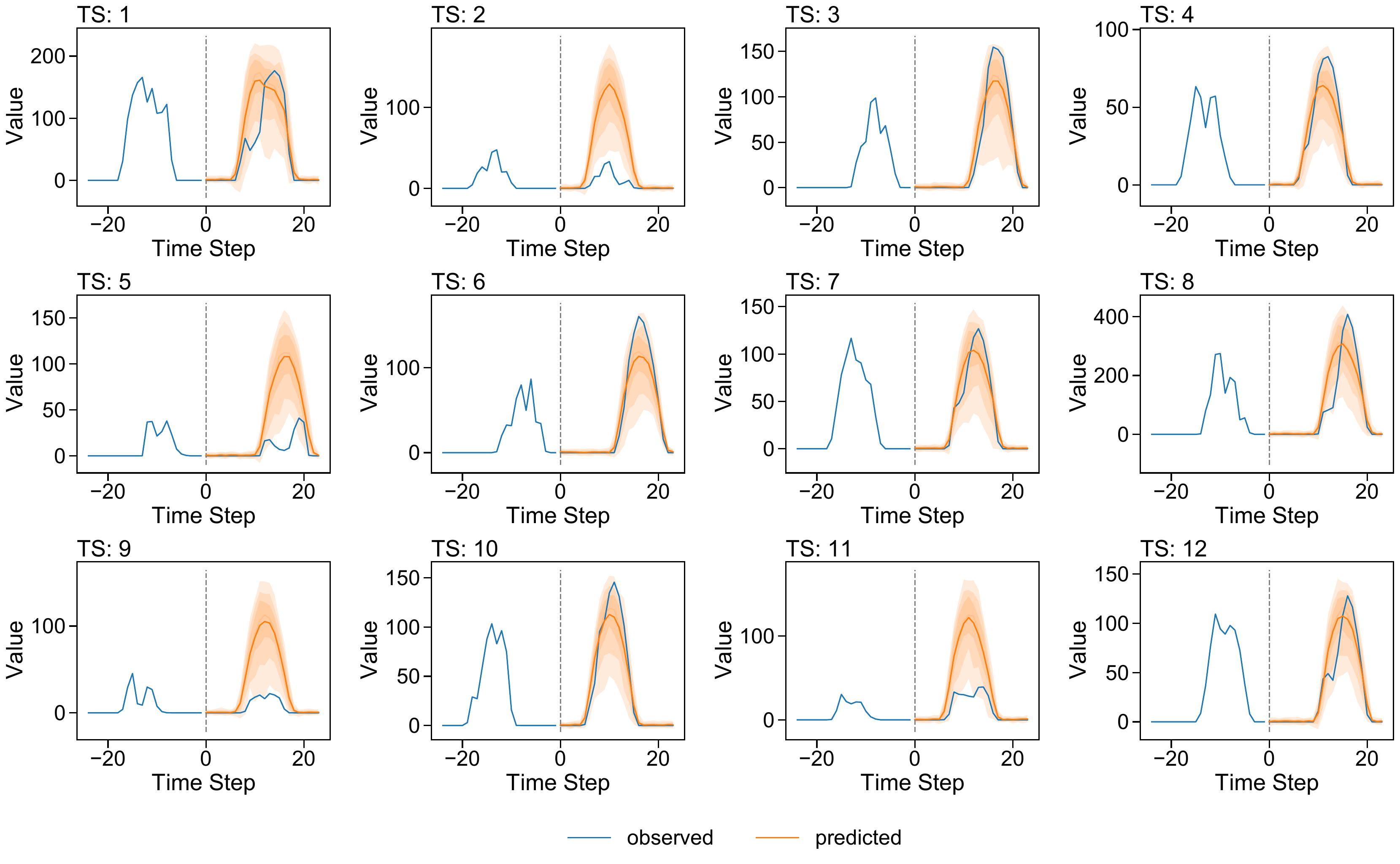}
  \caption{Visualization of forecasting results on \(\mathtt{solar}\) using GPVar with our method. }
\label{fig:deepar_solar_nips_vis}
\end{figure}

\begin{figure}[!ht]
  \centering\includegraphics[width=0.98\textwidth]{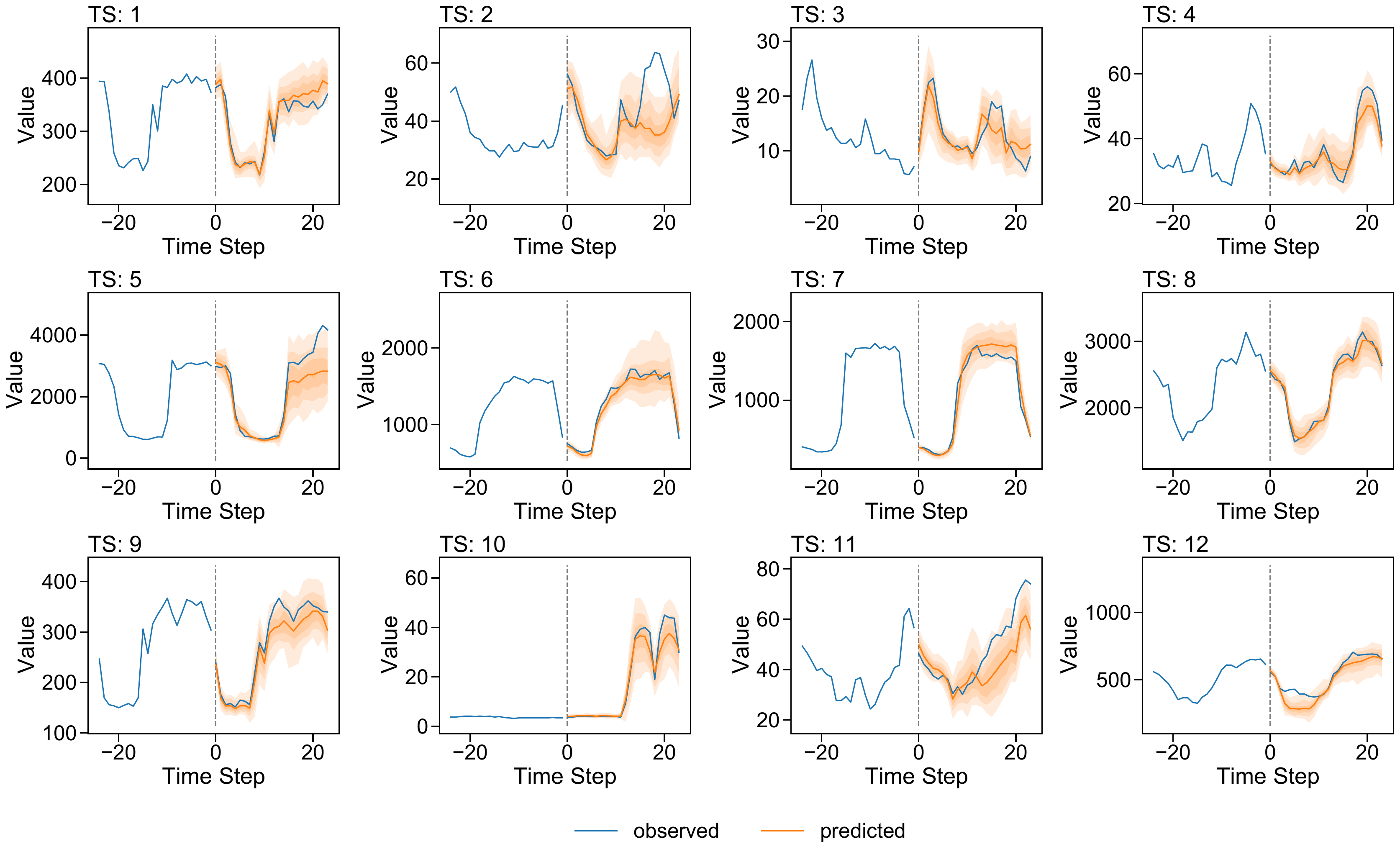}
  \caption{Visualization of forecasting results on \(\mathtt{electricity}\) using GPVar with our method. }
\label{fig:deepar_electricity_nips_vis}
\end{figure}

\begin{figure}[!ht]
  \centering\includegraphics[width=0.98\textwidth]{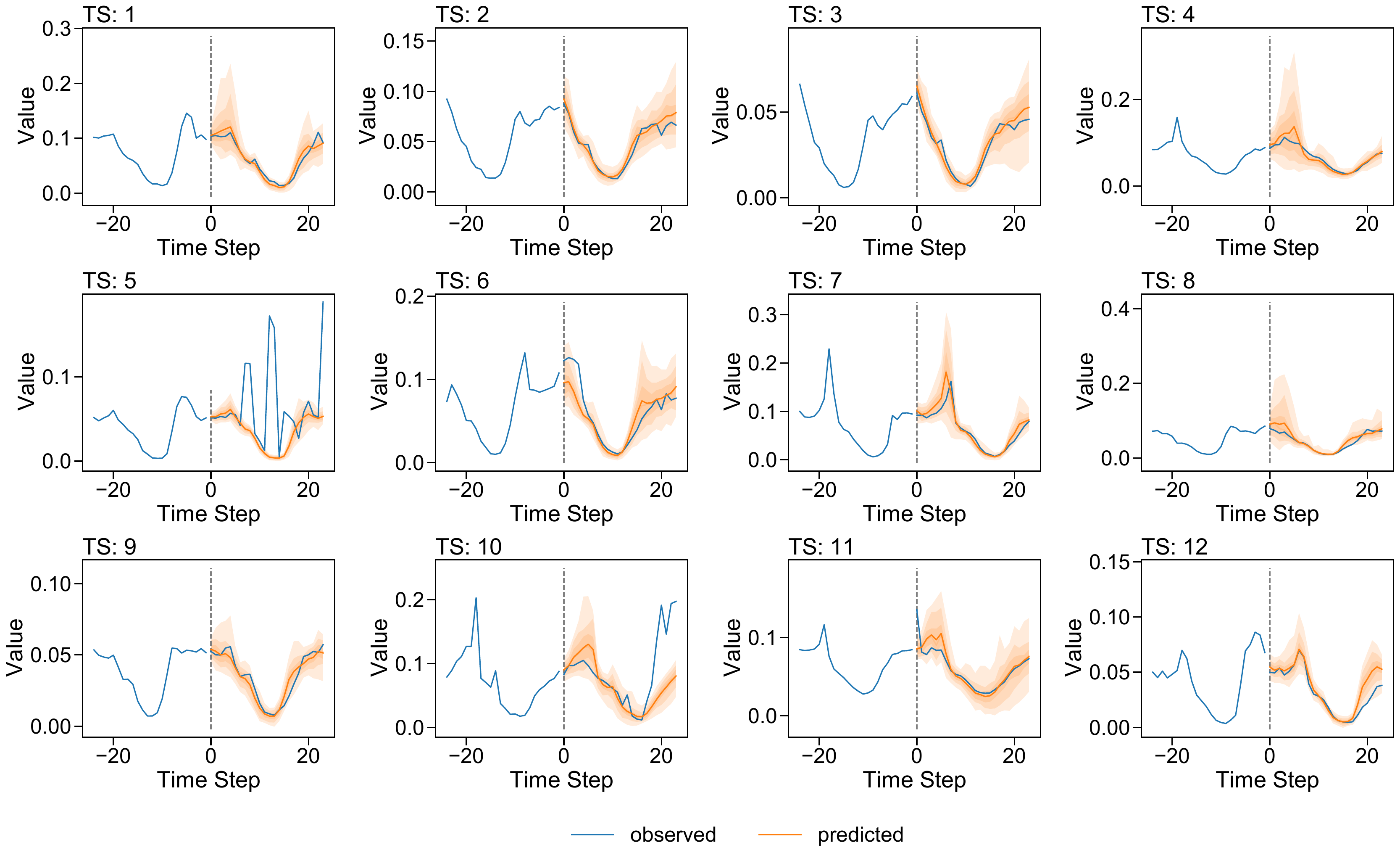}
  \caption{Visualization of forecasting results on \(\mathtt{traffic}\) using GPVar with our method. }
\label{fig:deepar_traffic_nips_vis}
\end{figure}

\begin{figure}[!ht]
  \centering\includegraphics[width=0.98\textwidth]{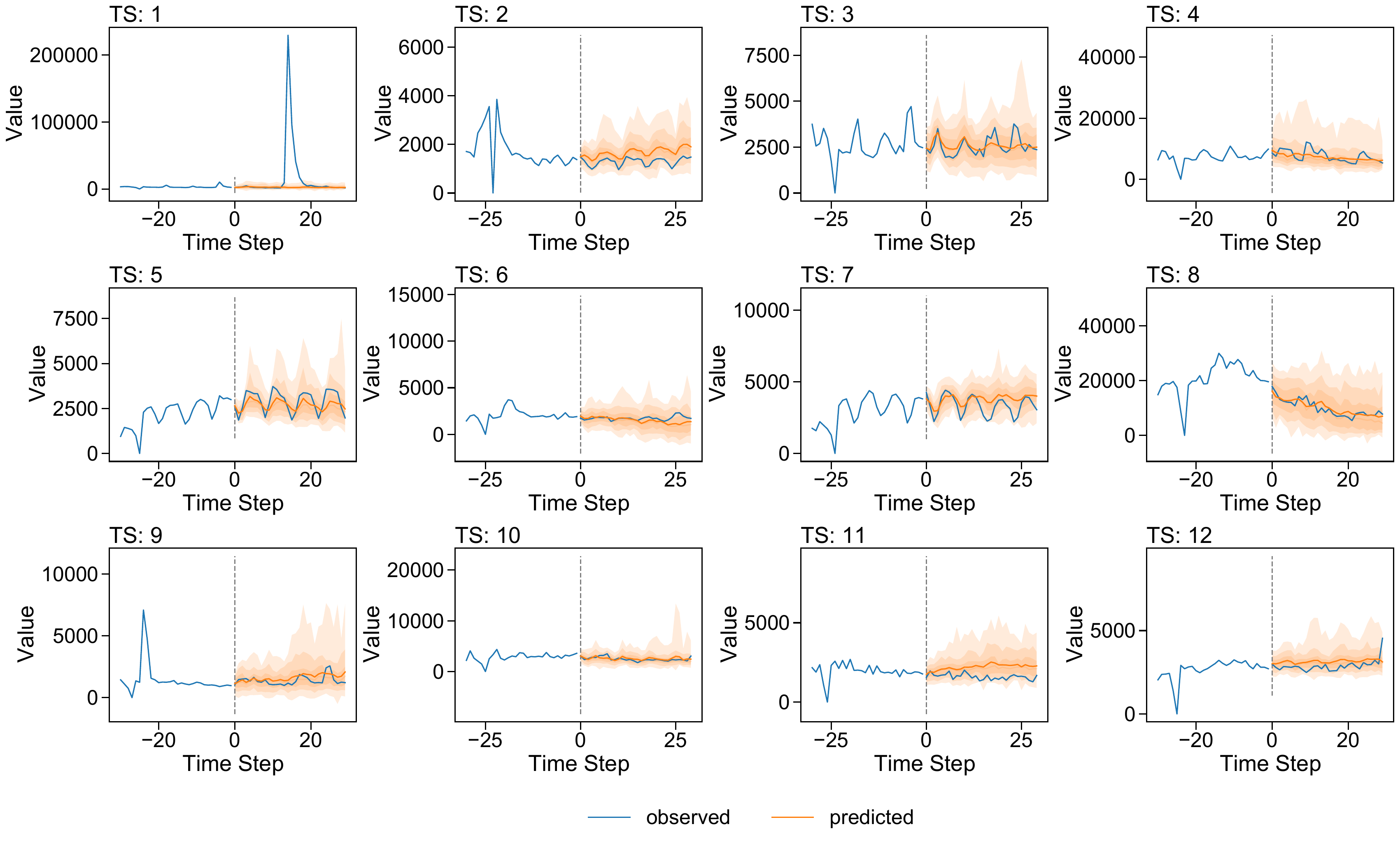}
  \caption{Visualization of forecasting results on \(\mathtt{wiki}\) using GPVar with our method. }
\label{fig:deepar_wiki2000_nips_vis}
\end{figure}

\begin{figure}[!ht]
  \centering\includegraphics[width=0.98\textwidth]{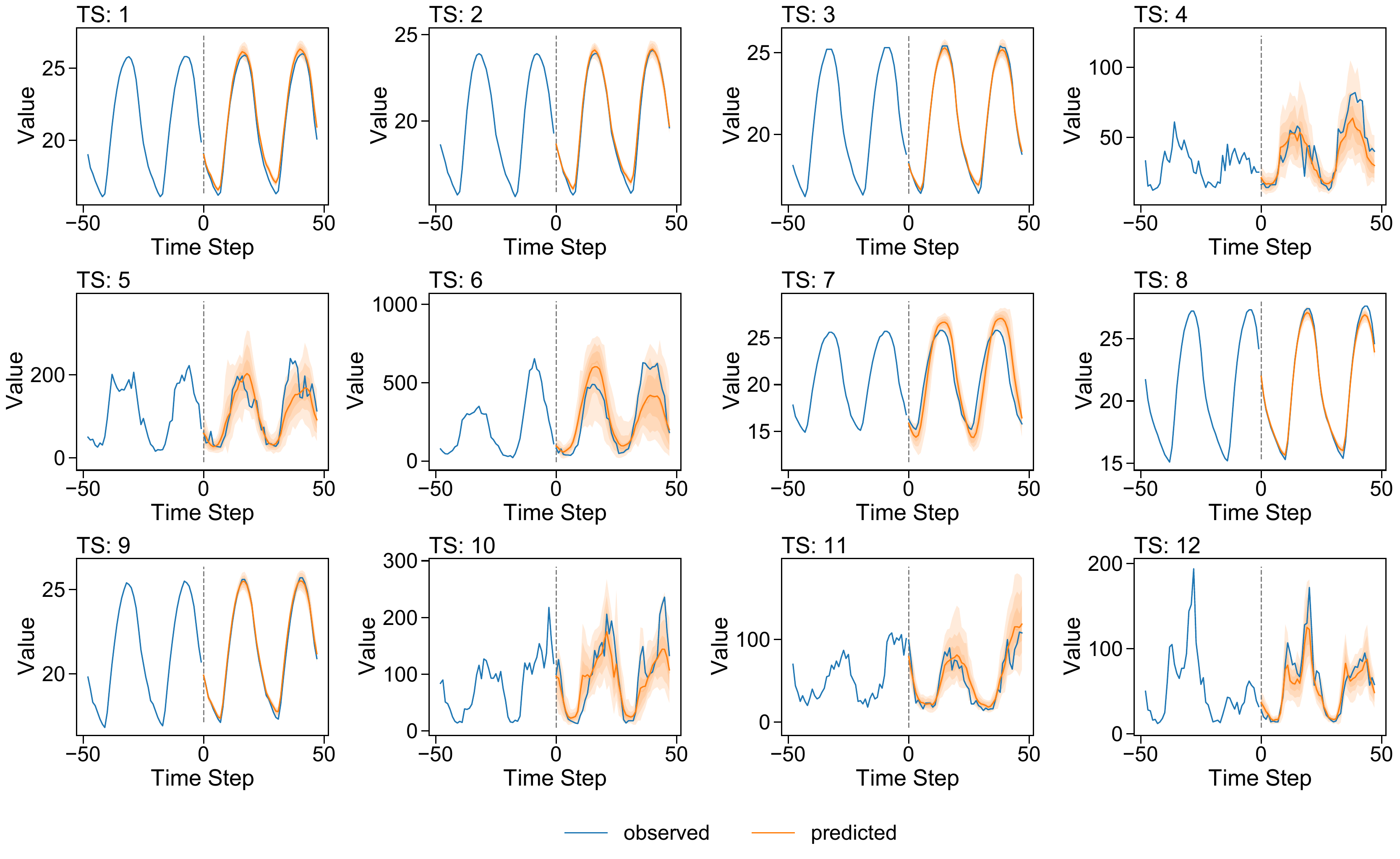}
  \caption{Visualization of forecasting results on \(\mathtt{m4\_hourly}\) using GPVar with our method. }
\label{fig:deepar_m4_hourly_vis}
\end{figure}

\begin{figure}[!ht]
  \centering\includegraphics[width=0.98\textwidth]{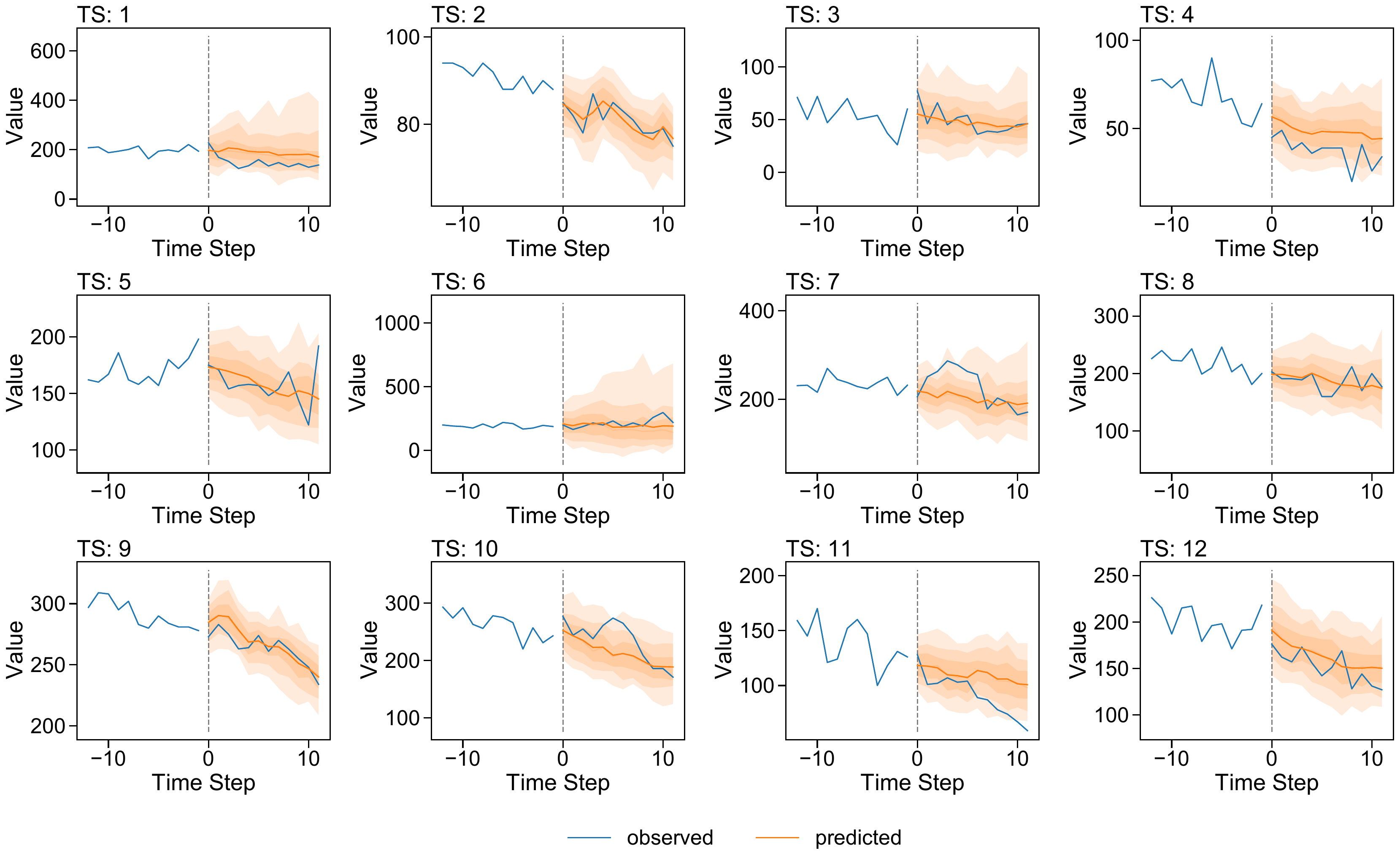}
  \caption{Visualization of forecasting results on \(\mathtt{pems03}\) using GPVar with our method. }
\label{fig:deepar_pems03_flow_vis}
\end{figure}

\begin{figure}[!ht]
  \centering\includegraphics[width=0.98\textwidth]{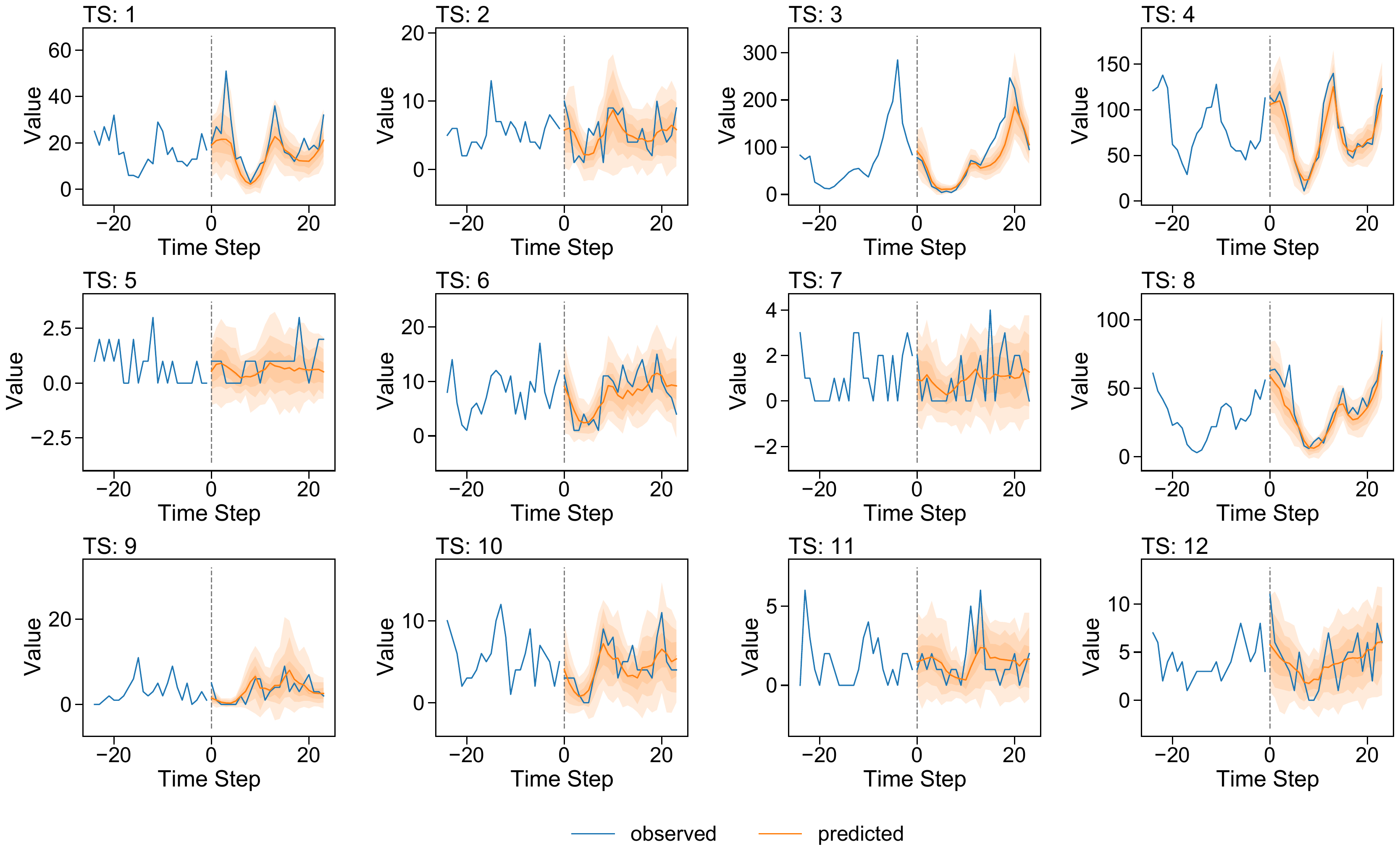}
  \caption{Visualization of forecasting results on \(\mathtt{uber\_hourly}\) using GPVar with our method. }
\label{fig:deepar_uber_tlc_hourly_vis}
\end{figure}


\clearpage
\section*{NeurIPS Paper Checklist}

\begin{enumerate}

\item {\bf Claims}
    \item[] Question: Do the main claims made in the abstract and introduction accurately reflect the paper's contributions and scope?
    \item[] Answer: \answerYes{} 
    \item[] Justification: The abstract and/or introduction have clearly stated the claims made, including the contributions made in the paper and important assumptions and limitations.

\item {\bf Limitations}
    \item[] Question: Does the paper discuss the limitations of the work performed by the authors?
    \item[] Answer: \answerYes{} 
    \item[] Justification: We have discussed the limitations of the work in the "Conclusion" section.

\item {\bf Theory Assumptions and Proofs}
    \item[] Question: For each theoretical result, does the paper provide the full set of assumptions and a complete (and correct) proof?
    \item[] Answer: \answerNA{} 
    \item[] Justification: We do not have theoretical result in this study.

    \item {\bf Experimental Result Reproducibility}
    \item[] Question: Does the paper fully disclose all the information needed to reproduce the main experimental results of the paper to the extent that it affects the main claims and/or conclusions of the paper (regardless of whether the code and data are provided or not)?
    \item[] Answer: \answerYes{} 
    \item[] Justification: We have fully disclosed all the information needed to reproduce the main experimental results of the paper in the Appendix.

\item {\bf Open access to data and code}
    \item[] Question: Does the paper provide open access to the data and code, with sufficient instructions to faithfully reproduce the main experimental results, as described in supplemental material?
    \item[] Answer: \answerNo{} 
    \item[] Justification: The code will be released after the paper is accepted. However, we have provided a sufficient amount of experimental details in the Appendix.

\item {\bf Experimental Setting/Details}
    \item[] Question: Does the paper specify all the training and test details (e.g., data splits, hyperparameters, how they were chosen, type of optimizer, etc.) necessary to understand the results?
    \item[] Answer: \answerYes{} 
    \item[] Justification: The paper specified all the training and test details necessary to understand the results.

\item {\bf Experiment Statistical Significance}
    \item[] Question: Does the paper report error bars suitably and correctly defined or other appropriate information about the statistical significance of the experiments?
    \item[] Answer: \answerYes{} 
    \item[] Justification: We ran all of our experiments for 10 times to calculate the standard deviation.

\item {\bf Experiments Compute Resources}
    \item[] Question: For each experiment, does the paper provide sufficient information on the computer resources (type of compute workers, memory, time of execution) needed to reproduce the experiments?
    \item[] Answer: \answerYes{} 
    \item[] Justification: The paper has indicated the type of compute workers CPU or GPU, internal cluster, or cloud provider, including relevant memory and storage.
    
\item {\bf Code Of Ethics}
    \item[] Question: Does the research conducted in the paper conform, in every respect, with the NeurIPS Code of Ethics \url{https://neurips.cc/public/EthicsGuidelines}?
    \item[] Answer: \answerYes{} 
    \item[] Justification: The research conducted in the paper conform, in every respect, with the NeurIPS Code of Ethics.

\item {\bf Broader Impacts}
    \item[] Question: Does the paper discuss both potential positive societal impacts and negative societal impacts of the work performed?
    \item[] Answer: \answerYes{} 
    \item[] Justification: We have discussed societal impacts in the last section of this paper.
    
\item {\bf Safeguards}
    \item[] Question: Does the paper describe safeguards that have been put in place for responsible release of data or models that have a high risk for misuse (e.g., pretrained language models, image generators, or scraped datasets)?
    \item[] Answer: \answerNA{} 
    \item[] Justification: The paper poses no such risks.

\item {\bf Licenses for existing assets}
    \item[] Question: Are the creators or original owners of assets (e.g., code, data, models), used in the paper, properly credited and are the license and terms of use explicitly mentioned and properly respected?
    \item[] Answer: \answerYes{} 
    \item[] Justification: The creators or original owners of assets (e.g., code, data, models), used in the paper, have been properly credited. The license and terms of use have been explicitly mentioned and properly respected.

\item {\bf New Assets}
    \item[] Question: Are new assets introduced in the paper well documented and is the documentation provided alongside the assets?
    \item[] Answer: \answerNA{} 
    \item[] Justification: The paper does not release new assets.

\item {\bf Crowdsourcing and Research with Human Subjects}
    \item[] Question: For crowdsourcing experiments and research with human subjects, does the paper include the full text of instructions given to participants and screenshots, if applicable, as well as details about compensation (if any)? 
    \item[] Answer: \answerNA{} 
    \item[] Justification: The paper does not involve crowdsourcing nor research with human subjects.

\item {\bf Institutional Review Board (IRB) Approvals or Equivalent for Research with Human Subjects}
    \item[] Question: Does the paper describe potential risks incurred by study participants, whether such risks were disclosed to the subjects, and whether Institutional Review Board (IRB) approvals (or an equivalent approval/review based on the requirements of your country or institution) were obtained?
    \item[] Answer: \answerNA{} 
    \item[] Justification: The paper does not involve crowdsourcing nor research with human subjects.

\end{enumerate}

\end{document}